\documentclass[10pt,twocolumn,letterpaper]{article}

\usepackage[pagenumbers]{cvpr} %

\usepackage{graphicx}
\usepackage{amsmath}
\usepackage{amssymb}
\usepackage{booktabs}
\usepackage{xcolor}

\usepackage[pagebackref,breaklinks,colorlinks]{hyperref}

\usepackage[capitalize]{cleveref}
\crefname{section}{Sec.}{Secs.}
\Crefname{section}{Section}{Sections}
\Crefname{table}{Table}{Tables}
\crefname{table}{Tab.}{Tabs.}

\definecolor{ggblue}{RGB}{68, 114, 196}
\definecolor{ggred}{RGB}{220, 20, 60}
\definecolor{gggreen}{RGB}{46, 139, 87}
\definecolor{ggorange}{RGB}{237, 125, 49}

\hypersetup{
    colorlinks=true,
    linkcolor=ggred,
    filecolor=ggblue,      
    urlcolor=ggblue,
    citecolor=gggreen,
}

\newcommand{\mypar}[1]{\vspace{1mm}\noindent\textbf{#1}}

\def\method{MCC\xspace}

\newcommand\crule[3][black]{\textcolor{#1}{\rule{#2}{#3}}}
\definecolor{unmasked}{HTML}{399C88}  
\definecolor{masked}{HTML}{333333}  

\definecolor{HighlightG}{HTML}{39b54a}  %
\definecolor{HighlightB}{HTML}{0071bc}  %

\definecolor{bluegreen}{HTML}{00829f}  %
\definecolor{maskcolor}{HTML}{43aa9a}  %

\newcommand{\figref}[1]{Fig.~\ref{fig:#1}}
\newcommand{\tabref}[1]{Table~\ref{tab:#1}}
\newcommand{\secref}[1]{\S\ref{sec:#1}}

\usepackage{tabulary}
\usepackage{multirow}

\newcolumntype{x}[1]{>{\centering\arraybackslash}p{#1pt}}
\newcolumntype{y}[1]{>{\raggedright\arraybackslash}p{#1pt}}
\newcolumntype{z}[1]{>{\raggedleft\arraybackslash}p{#1pt}}
\newlength\savewidth\newcommand\shline{\noalign{\global\savewidth\arrayrulewidth
    \global\arrayrulewidth 1pt}\hline\noalign{\global\arrayrulewidth\savewidth}}
\newcommand{\tablestyle}[2]{\setlength{\tabcolsep}{#1}\renewcommand{\arraystretch}{#2}\centering\footnotesize}

\usepackage{makecell}

\newcommand{\teaserin}[1]{%
\begin{minipage}{\linewidth}\begin{tikzpicture}
    \node (img) {\frame{\includegraphics[width=\linewidth]{#1}}};
\end{tikzpicture}\end{minipage}\vspace{-1mm}}

\newcommand{\teaserout}[1]{%
\begin{minipage}{\linewidth}\begin{tikzpicture}
    \node (img) {\frame{\includegraphics[width=\linewidth]{#1}}};
\end{tikzpicture}\end{minipage}\vspace{-1mm}}

\newcommand{\inputimg}[1]{%
\makecell{Input image\\ \frame{\includegraphics[width=\linewidth]{#1}}}}

\newcommand{\seenimg}[1]{%
\begin{minipage}{\linewidth}\begin{tikzpicture}
    \node (img) {\frame{\includegraphics[width=\linewidth]{#1}}};
    \node [below right=1mm, fill=white, opacity=0.7] at (img.north west){\color{black!0}{eeee}};
    \node [below right=1mm] at (img.north west){\color{black}{Seen}};
\end{tikzpicture}\end{minipage}\vspace{-1mm}}
\newcommand{\outimg}[1]{%
\begin{minipage}{\linewidth}\begin{tikzpicture}
    \node (img) {\frame{\includegraphics[width=\linewidth]{#1}}};
    \node [below right=1mm, fill=white, opacity=0.7] at (img.north west){\color{black!0}{eeeeee}};
    \node [below right=1mm] at (img.north west){\color{black}{Output}};
\end{tikzpicture}\end{minipage}\vspace{-1mm}}
\newcommand{\myimg}[1]{%
\begin{minipage}{\linewidth}\begin{tikzpicture}
    \node (img) {\frame{\includegraphics[width=\linewidth]{#1}}};
\end{tikzpicture}\end{minipage}\vspace{-1mm}}

\newcommand{\sceneimg}[2]{%
\begin{minipage}{\linewidth}\begin{tikzpicture}
    \node (img) {\frame{\includegraphics[height={#1}]{#2}}};
\end{tikzpicture}\end{minipage}\vspace{-1mm}}

\newcommand{\rbr}[1]{\left(#1\right)}
\newcommand{\sbr}[1]{\left[#1\right]}

\newcommand{\RR}{\mathbb{R}}

\newcommand{\app}{\raise.17ex\hbox{$\scriptstyle\sim$}}
\newcommand{\x}{\times}

\usepackage{nicefrac}

\makeatletter
\@namedef{ver@everyshi.sty}{}
\makeatother

\usepackage{tikz,pgfplots}
\usepackage{tango}

\pgfplotsset{compat = 1.3,
	legend style={font=\scriptsize},
	legend cell align={left},
	legend style={cells={align=left}, draw=black!20},
	grid=both,
	grid style={dotted},
	tick style={draw=none},
	enlarge x limits=false,
	enlarge y limits=false,
	axis line style={draw=black!100},
	axis lines=left,
}

\pgfplotscreateplotcyclelist{mycolormarklist}{%
	chameleon3,mark=diamond*,mark options={solid, fill=chameleon3}\\
	skyblue2,mark=pentagon*,mark options={solid}\\
	scarletred3,mark=triangle*,mark options={solid}\\
	plum1,mark=square*,mark options={solid}\\
	aluminium5,mark=otimes*,mark options={solid}\\
	orange1,mark=square*,mark options={solid}\\
	skyblue3\\
	chocolate2\\
	maroon\\
	butter3\\
}
\pgfplotscreateplotcyclelist{mystylelist}{%
	densely dashdotted\\
	solid\\
	dotted\\
	densely dashed\\
	dashed\\
	dashdotted\\
	densely dotted\\
	densely dash dot dot\\
	dotted\\
	dash dot dot\\
}

\def\x{$\times$}

\definecolor{xycolor}{HTML}{0071bc}
\newcommand{\xycolor}[1]{\textcolor{xycolor}{#1}}
\definecolor{wcolor}{RGB}{103, 78, 167}
\newcommand{\wcolor}[1]{\textcolor{wcolor}{#1}}
\definecolor{tcolor}{RGB}{80, 200, 180}

\definecolor{eicolor}{RGB}{153, 51, 102}
\newcommand{\eicolor}[1]{\textcolor{eicolor}{#1}}

\newcommand{\blockatta}[3]{\multirow{2}{*}{\(\left[\begin{array}{c}\text{\eicolor{MHA}(\wcolor{#1})}\\ \text{MLP(\wcolor{#2})}\end{array}\right]\)$\times$#3}
}
\newcommand{\blockattb}[3]{\multirow{3}{*}{\(\left[\begin{array}{c}\text{\eicolor{MHA}(\wcolor{#1})}\\ \text{MLP(\wcolor{#2})}\\ \text{\texttt{[cls]} readout}\end{array}\right]\)$\times$#3}
}

\newcommand\blfootnote[1]{%
  \begingroup
  \renewcommand\thefootnote{}\footnote{#1}%
  \addtocounter{footnote}{-1}%
  \endgroup
}

\def\cvprPaperID{****}
\def\confName{CVPR}
\def\confYear{2023}

\begin{document}

\title{Multiview Compressive Coding for 3D Reconstruction}

\newcommand{\authorskip}{\hspace{2.5mm}}
\author{
Chao-Yuan Wu \authorskip Justin Johnson \authorskip Jitendra Malik \authorskip Christoph Feichtenhofer \authorskip Georgia Gkioxari \\[2mm]
FAIR, Meta AI\vspace{1mm}
}

\maketitle

\begin{abstract}
A central goal of visual recognition is to understand objects and scenes from a single image.
2D recognition has witnessed tremendous progress thanks to large-scale learning and general-purpose representations.
Comparatively, 3D poses new challenges stemming from occlusions not depicted in the image.
Prior works try to overcome these by inferring from multiple views or rely on scarce CAD models and category-specific priors which hinder scaling to novel settings.
In this work, we explore single-view 3D reconstruction by learning generalizable representations inspired by advances in self-supervised learning. 
We introduce a simple framework that operates on 3D points of single objects or whole scenes coupled with category-agnostic large-scale training from diverse RGB-D videos. 
Our model, Multiview Compressive Coding (\method), learns to compress the input appearance and geometry to predict the 3D structure by querying a 3D-aware decoder.
\method's generality and efficiency allow it to learn from large-scale and diverse data sources with strong generalization to novel objects imagined by DALL$\cdot$E 2 or captured in-the-wild with an iPhone. 
\blfootnote{Project page: \href{https://mcc3d.github.io}{https://mcc3d.github.io}}
\end{abstract}
\vspace{-6mm}
\section{Introduction}
\label{sec:intro}
\begin{figure}[t]
\centering
\hspace{-4mm}
\resizebox{1.02\linewidth}{!}{
\subfloat[\textbf{MCC Overview}\label{fig:teaser:a}]{
\includegraphics[width=0.78\linewidth]{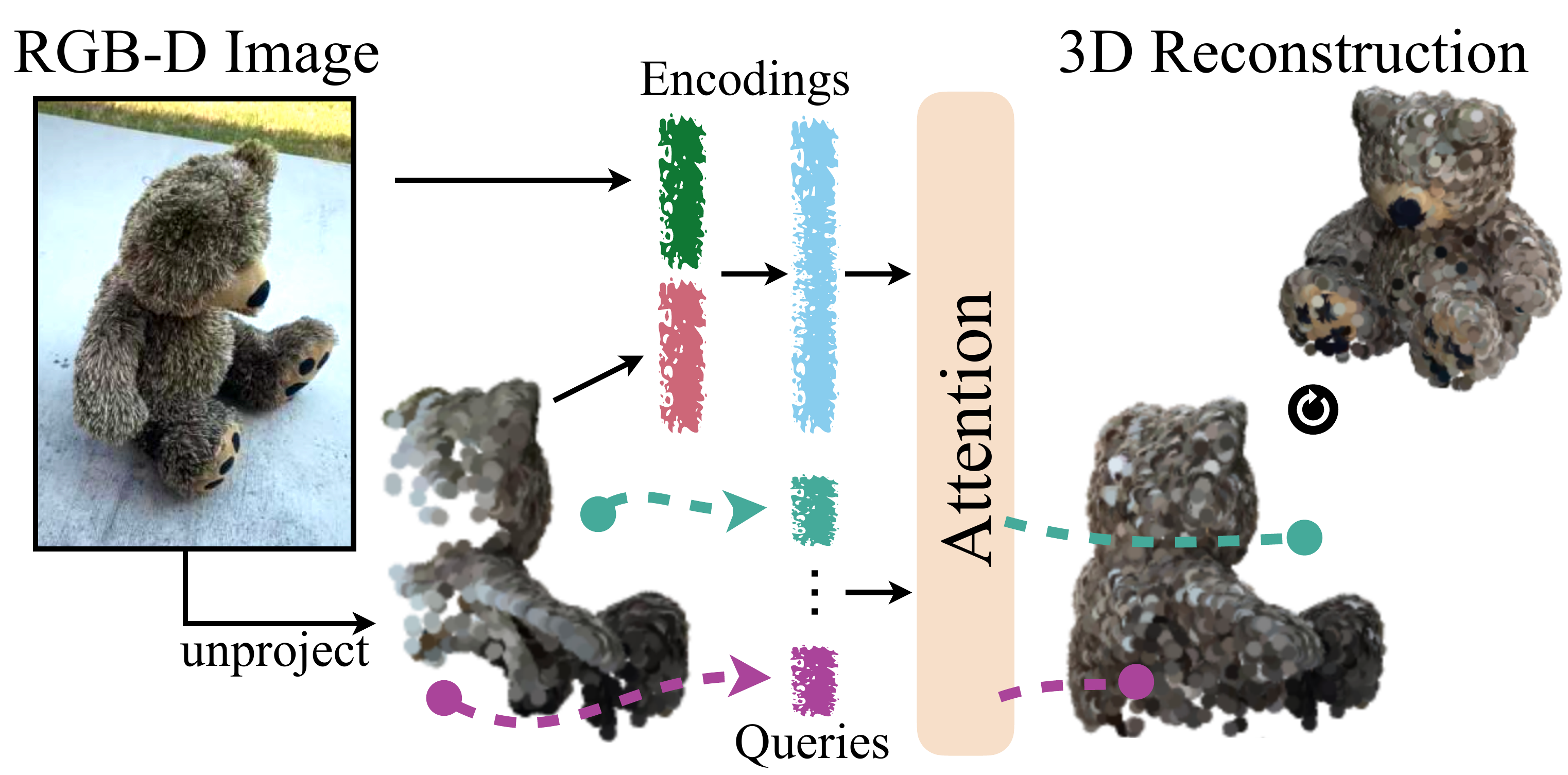}
}}

\hspace{-4mm}
\resizebox{1.02\linewidth}{!}{
\subfloat[\textbf{3D Reconstructions by MCC}\label{fig:teaser:b}]{%
\resizebox{0.78\linewidth}{!}{
\tablestyle{1.0pt}{1.05}
\begin{tabular}{@{}x{35}x{35}x{35}x{35}x{35}x{35}@{}}
\teaserin{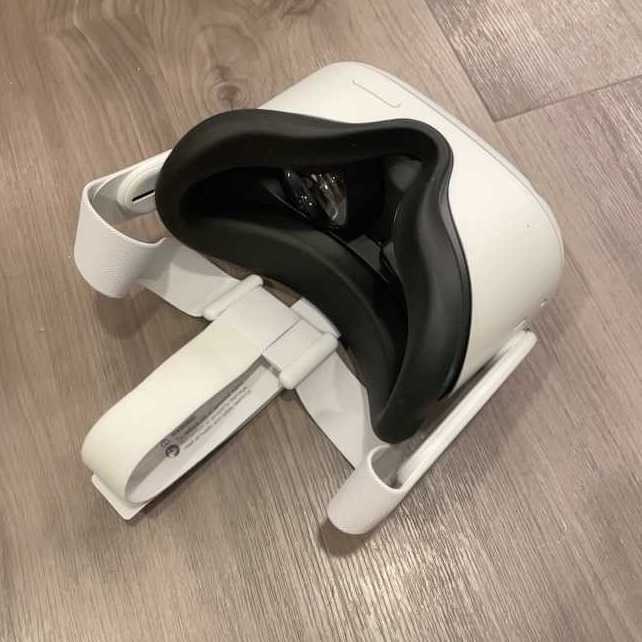}&
\teaserout{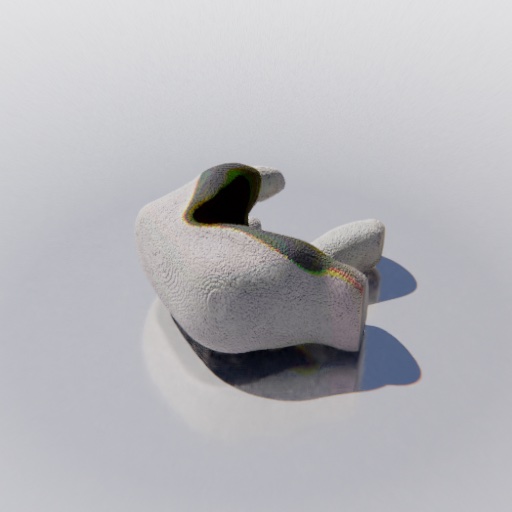}&
\teaserin{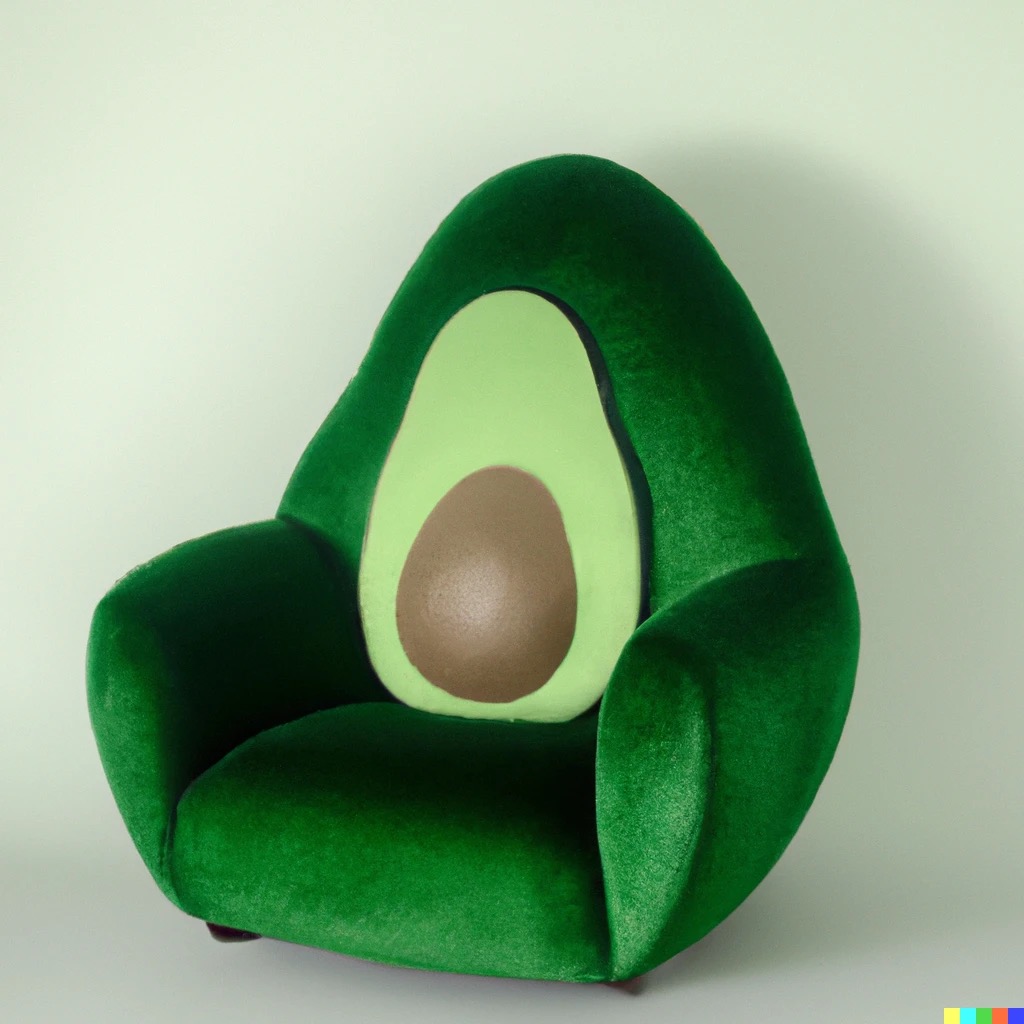}&
\teaserout{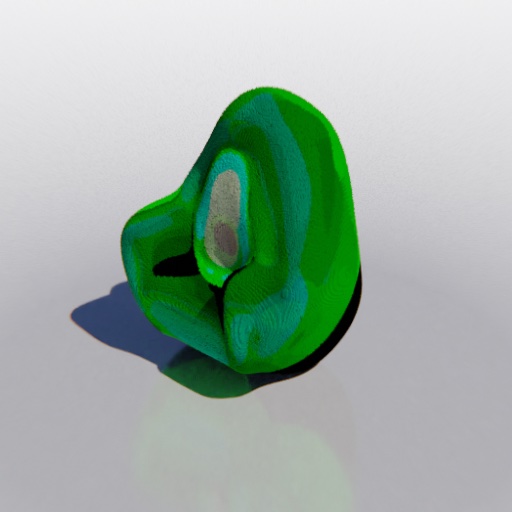}&
\teaserin{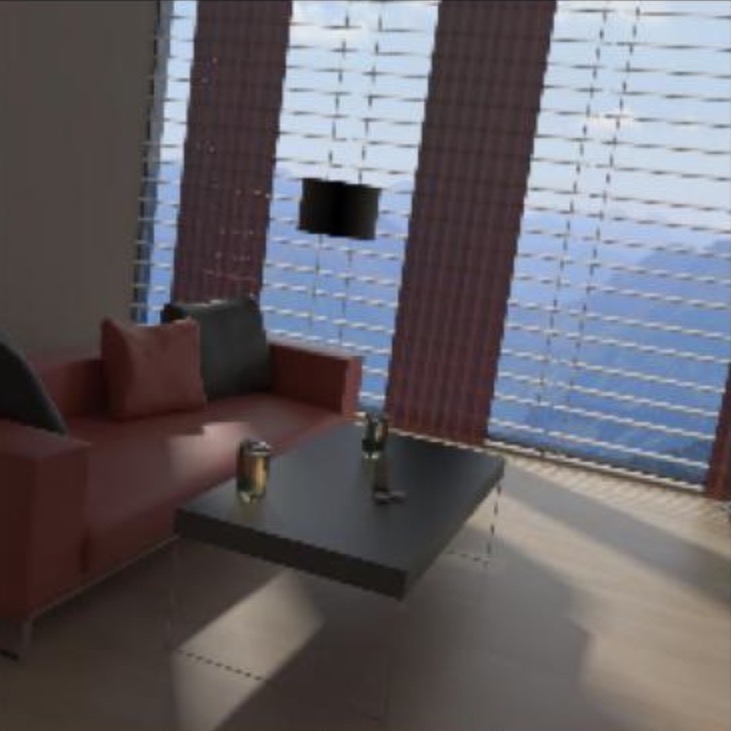}&
\teaserout{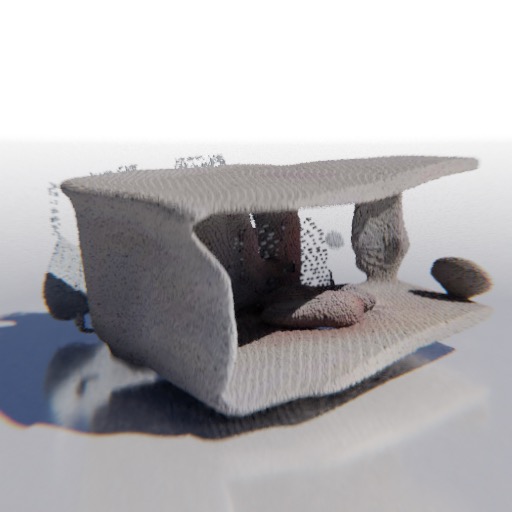}\\
Input & Output & Input & Output & Input & Output
\end{tabular}
}}}
\vspace{-2mm}
\caption{\textbf{Multiview Compressive Coding (\method).} (a): \method encodes an input RGB-D image and  uses an attention-based model to predict the occupancy and color of query points to form the final 3D reconstruction. (b): \method generalizes to novel objects captured with iPhones (left) or imagined by DALL$\cdot$E 2~\cite{ramesh2022hierarchical} (middle). It is also general -- it works not only on objects but also scenes (right).}
\label{fig:teaser}
\vspace{-4mm}
\end{figure}

Images depict objects and scenes in diverse settings.
Popular 2D visual tasks, such as object classification~\cite{deng2009imagenet} and segmentation~\cite{lin2014microsoft,zhou2017scene}, aim to recognize them on the image plane.
But image planes do not capture scenes in their entirety.
Consider \figref{teaser:a}. 
The toy's left arm is not visible in the image.
This is framed by the task of 3D reconstruction: given an image, \emph{fully} reconstruct the scene in 3D.

3D reconstruction is a longstanding problem in AI with applications in robotics and AR/VR. 
Structure from Motion~\cite{hartley2003multiple,szeliski2022computer} lifts images to 3D by triangulation.
Recently, NeRF~\cite{nerf} optimizes radiance fields to synthesize novel views. 
These approaches require many views of the same scene during inference and do not generalize to novel scenes from a single image. 
Others~\cite{wang2018pixel2mesh,gkioxari2019mesh} predict 3D from a single image but rely on expensive CAD supervision~\cite{chang2015shapenet,pix3d}.
Reminiscent of generalized cylinders~\cite{nevatia1977description}, some introduce object-specific priors via category-specific 3D templates~\cite{3DRCNN_CVPR18,kanazawa2018learning,kulkarni2020acsm}, pose~\cite{novotny2019c3dpo} or symmetries~\cite{wu2020unsupervised}.
While impressive, these methods cannot scale as they rely on onerous 3D annotations and category-specific priors which are not generally true.
Alas large-scale learning, which has shown promising generalization results for images~\cite{radford2021learning} and language~\cite{brown2020language}, is largely underexplored for 3D reconstruction.

Image-based recognition is entering a new era thanks to domain-agnostic architectures, like transformers~\cite{transformer,vit}, and large-scale category-agnostic learning~\cite{he2022masked} .
Motivated by these advances, we present a scalable, general-purpose model for 3D reconstruction from a single image. 
We introduce a simple, yet effective, framework that operates directly on 3D points.
3D points are general as they can capture any objects or scenes and are more versatile and efficient than meshes and voxels.
Their generality and efficiency enables large-scale category-agnostic training.
In turn, large-scale training makes our 3D model effective.

Central to our approach is an input encoding and a queriable 3D-aware decoder.
The input to our model is a single RGB-D image, which returns the visible (\emph{seen}) 3D points via unprojection.
Image and points are encoded with transformers.
A new 3D point, sampled from 3D space, queries a transformer decoder conditioned on the input
to predict its occupancy and its color.
The decoder reconstructs the full, \emph{seen} and \emph{unseen}, 3D geometry, as shown in \figref{teaser:a}.
Our occupancy-based formulation, introduced in~\cite{mescheder2019occupancy}, frames 3D reconstruction as a binary classification problem and removes constraints pertinent to specialized representations (\eg, deformations of a 3D template) or a fixed resolution.
Being tasked with predicting the \emph{unseen} 3D geometry of diverse objects or scenes, our decoder learns a strong 3D representation. 
This finding directly connects to recent advances in image-based self-supervised learning and masked autoencoders (MAE)~\cite{he2022masked} which learn powerful image representations by predicting masked (unseen) image patches.

Our model inputs single RGB-D images, which are ubiquitous thanks to advances in hardware.
Nowadays, depth sensors are found in iPhone's front and back cameras.
We show results from iPhone captures in~\secref{exp:obj} and \figref{teaser:b}.
Our decoder predicts point cloud occupancies. 
Supervision is sourced from multiple RGB-D views, \eg, video frames, with relative camera poses, \eg, from COLMAP~\cite{schoenberger2016sfm,schoenberger2016mvs}.
The posed views produce 3D point clouds which serve as proxy ground truth.
These point clouds are far from ``perfect'' as they are amenable to sensor and camera pose noise.
However, we show that when used at scale they are sufficient for our model.
This suggests that 3D annotations, which are expensive to acquire, can be replaced with many RGB-D video captures, which are much easier to collect.

We call our approach Multiview Compressive Coding (\method), as it learns from many views, compresses appearance and geometry and learns a 3D-aware decoder.
We demonstrate the generality of \method by experimenting on six diverse data sources: CO3D~\cite{reizenstein21co3d}, Hypersim~\cite{roberts2021hypersim}, Taskonomy~\cite{zamir2018taskonomy}, ImageNet~\cite{deng2009imagenet}, in-the-wild iPhone captures and DALL$\cdot$E 2~\cite{ramesh2022hierarchical} generations.
These datasets range from large-scale captures of more than 50 common object types, to holistic scenes, such as warehouses, auditoriums, lofts, restaurants, and imaginary objects.
We compare to state-of-the-art methods, tailored for single objects~\cite{yu2021pointr,reizenstein21co3d,henzler2021unsupervised} and scene reconstruction~\cite{kulkarni2021drdf} and show our model's superiority in both settings with a unified architecture.
Enabled by \method's general purpose design, we show the impact of large-scale learning in terms of reconstruction quality and zero-shot generalization on novel object and scene types.

\section{Related Work}
\label{sec:related}

\mypar{Multiview 3D reconstruction} is a longstanding problem in computer vision. 
Traditional techniques include binocular stereopsis~\cite{wheatstone1838xviii}, SfM~\cite{hartley2003multiple,szeliski2022computer,scharstein2002taxonomy,torresani2008nonrigid,tomasi1992shape}, and SLAM~\cite{smith1990estimating,castellanos1999spmap}.
Reconstruction by analysis~\cite{esteban2004silhouette} or synthesis via volume rendering~\cite{kajiya1984ray} of implicit~\cite{nerf,zhang2020nerf++} and explicit~\cite{sitzmann2019deepvoxels,liu2020neural} representations have shown to produce strong results.
Supervised approaches predict voxels~\cite{xie2019pix2vox,wang2021multi} or meshes~\cite{worchel2022multi,wen2019pixel2mesh++} by training deep nets.
These techniques produce high-quality outputs, but rely on multiple views at test time.
In this work, we assume a single RGB-D image during inference.

\mypar{Single-view 3D reconstruction} is challenging.
One line of work trains models that predict 3D geometry via CAD~\cite{wang2018pixel2mesh,gkioxari2019mesh}, meshes~\cite{xu2019disn,kulkarni2021drdf}, voxels~\cite{girdhar2016learning,wu2017marrnet} or point clouds~\cite{fan2017point,mescheder2019occupancy} supervision. 
Results are commonly demonstrated on synthetic simplistic benchmarks, such as ShapeNet~\cite{chang2015shapenet}, or for a small set of object categories, as in Pix3D~\cite{pix3d}.
Weakly supervised approaches use category-specific priors via 3D shape templates~\cite{kanazawa2018learning,kulkarni2020acsm,ucmrGoel20} and pose~\cite{novotny2019c3dpo} or learn via 2D silhouettes and re-projection on posed views~\cite{kato2018neural,liu2019soft,chen2019learning,ravi2020pytorch3d}. 
While impressive, these approaches are limited to specific objects from a closed-world vocabulary.
Some~\cite{yan2016perspective,wallace2019few} explore category-agnostic models, but focus on synthetic datasets.
In this work, we learn a general-purpose 3D representation from RGB-D views from a diverse and large set of data sources of real-world objects and scenes.

\mypar{Shape completion} methods complete the 3D geometry of partial reconstructions. 
For objects, methods directly output full point clouds~\cite{huang2020pf,yu2021pointr,yuan2018pcn} or deploy generative models~\cite{yan2022shapeformer,zhou20213d}, but are typically tied to a fixed resolution.
For scenes, techniques include plane fitting~\cite{monszpart2015rapter}, 3D model fitting and retrieval~\cite{geiger2015joint,nan2012search} or leverage symmetries~\cite{kim2012acquiring} and predict 3D semantics~\cite{song2017semantic,cao2022monoscene,firman2016structured}.
Our model tackles both objects and scenes with a unified architecture and outputs any-resolution 3D geometry with a 3D-aware decoder.
We compare to recent shape completion techniques.

\mypar{Implicit 3D representations}
such as SDFs~\cite{saito2019pifu,park2019deepsdf} and occupancy nets (OccNets)~\cite{mescheder2019occupancy} have proven effective 3D representations.
NeRF~\cite{nerf} optimizes per-scene neural fields for view synthesis. 
NeRF extensions target scene generalization by encoding input views with deep nets~\cite{yu2021pixelnerf,henzler2021unsupervised,raj2021pixel} or improve reconstruction quality by supervising with depth~\cite{deng2022depth}.
\method adopts an occupancy-based representation, similar to OccNets~\cite{mescheder2019occupancy}, with an attention mechanism on encoded appearance and geometric cues which allows it to predict in any 3D region, even outside the camera frustum, efficiently.
We show that this strategy outperforms the global-feature strategy from OccNets~\cite{mescheder2019occupancy} or single-location features used in NeRF-based methods~\cite{henzler2021unsupervised,reizenstein21co3d}.

\mypar{Self-supervised learning} has advanced image~\cite{bao2021beit,radford2021learning,he2022masked} and language~\cite{devlin2018bert,brown2020language} understanding.
For images, masked autoencoders~\cite{he2022masked} paired with transformers and large-scale category-agnostic training learn general representations for 2D recognition. 
We draw from these findings and extend the architecture and learning for the task of 3D reconstruction.

\begin{figure*}[t]
\centering
\includegraphics[width=1.01\linewidth]{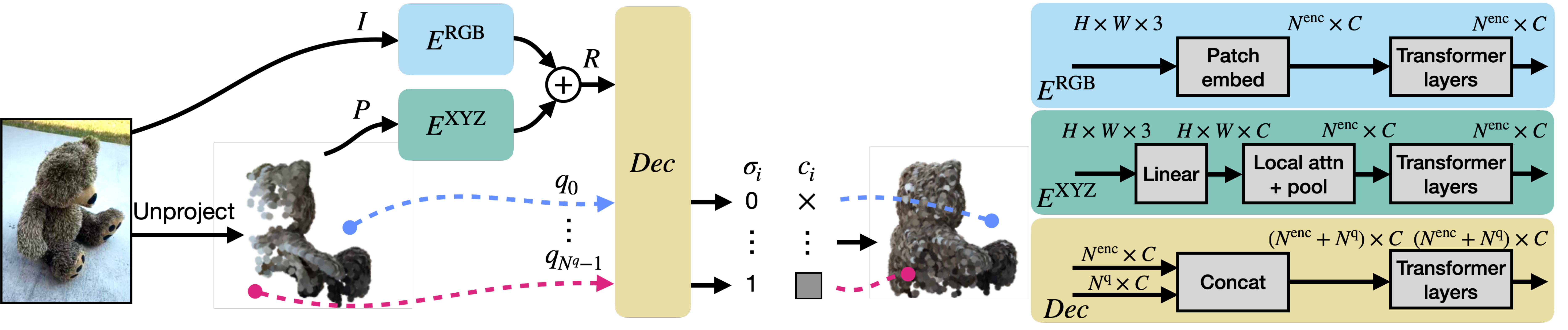}
\vspace{-7mm}
\caption{\textbf{Model Overview.} Given an RGB-D image, MCC unprojects the pixels of the input RGB image $I$ to the corresponding 3D points $P$. 
An image encoder $E^\mathrm{RGB}$ and a geometry encoder $E^\mathrm{XYZ}$ encode $I$ and $P$ into a 3D-aware representation $R$.
A decoder predicts the occupancy $\sigma_i$ and color $c_i$ of query $q_i$, $i=0, \ldots, N_q-1$, conditioned on $R$.
The predicted colored points form the final 3D reconstruction.
}
\label{fig:model}
\vspace{-3mm}
\end{figure*}

\section{Multiview Compressive Coding (\method)}
\label{sec:method}

\method adopts an encoder-decoder architecture.
The input RGB-D image is fed to the encoder to produce encoding $R$.
The decoder inputs a \emph{query} 3D point $q_i \in \RR^3$, along with $R$, to predict its occupancy probability $\sigma_i \in \sbr{0,1}$, as in \cite{mescheder2019occupancy}, and RGB color $c_i \in \sbr{0,1}^3$.
\figref{model} illustrates our model.

During training, we supervise \method with ``true" points derived from posed RGB-D views.
These point clouds serve as ground truth: $q_i$ is labeled as positive if it is close to the ground truth and negative otherwise.
Intuitively, the other views guide the model to reason about what parts of the \emph{unseen} space belong to the object or scene.
As a result, the input encoding $R$ learns a representation of the \emph{full} 3D geometry and guides the decoder to make the right prediction.

During inference, the model predicts occupancy and color for a grid of points at any desired resolution.
The set of occupied colored points forms the final reconstruction.

\method requires only \emph{points} for supervision, extracted from posed RGB-D views, \eg, video frames.
Note that the derived point clouds, which serve as ground truth, are far from perfect due to noise in the captures and pose estimation.
However, when used at scale they are sufficient.
This deviates from OccNets~\cite{mescheder2019occupancy} and other distance-based works~\cite{park2019deepsdf,saito2019pifu} which rely on clean CAD models or 3D meshes.
This is an important finding as it suggests that expensive CAD supervision can be replaced with cheap RGB-D video captures.
This property of \method allows us to train on a wide range of diverse data.
In \secref{exp:obj}, we show that large-scale training is crucial for high-quality reconstruction.

\subsection{\method Encoder}
The input to our model is a single RGB-D image.
Let $I \in \RR^{H\times W\times 3}$ be the RGB image and $\Delta \in \RR^{H\times W}$ the associated depth. 
We use $\Delta$ to unproject the pixels into their positions $P \in \RR^{H\times W\times 3}$ in 3D.
$I$ and $P$ are encoded into a single representation $R$ via
\begin{align}
R := f\rbr{E^\mathrm{RGB}(I), E^\mathrm{XYZ}(P)} \in \RR^{N^\mathrm{enc} \times C}
\label{eq:enc}
\end{align}
$E^\mathrm{RGB}$ and $E^\mathrm{XYZ}$ are two transformers~\cite{transformer}.
$E^\mathrm{RGB}$ follows a ViT architecture~\cite{vit} to encode the input image $I$.\linebreak
$E^\mathrm{XYZ}$ processes the input points $P$ similar to a ViT, but encodes 3D coordinates instead of RGB color channels.
We explain in detail how to adapt a ViT to encode the input points $P$ in \secref{method:design}.
$f$ concatenates the two outputs from the transformers along the channel dimension followed by a linear projection to $C$-dimensions.
$N^\mathrm{enc}$ is the number of tokens used in the transformers.
\figref{model} shows an illustration.

The proposed two-tower design is general and performant.
Alternative designs are ablated in \secref{exp:obj}.

\subsection{\method Decoder}
The decoder takes as input the output of the encoder, $R$, and $N^q$ 3D point queries $q_i$, $i=0,\ldots,N_q-1$, to predict occupancy and colors for each point,
\begin{align}
(\sigma_0, c_0), (\sigma_1, c_1), \ldots := Dec(R, q_0, q_1, \ldots)
\label{eq:dec}
\end{align}
The decoder $Dec$ linearly projects each query $q_i$ to $C$-dimensions (the same as $R$), concatenates them with $R$ in the token dimension, and then uses a transformer to model the interactions between $R$ and queries.
We draw inspiration from MAE~\cite{he2022masked} for this design.
The output feature of each query token is passed through a binary classification head that predicts its occupancy $\sigma_i$, and a 256-way classification head that predicts its RGB color $c_i$~\cite{van2016conditional}.

\begin{figure}[t]
\centering
\begin{minipage}[c]{0.51\linewidth}\vspace{-7mm}
\includegraphics[width=\linewidth]{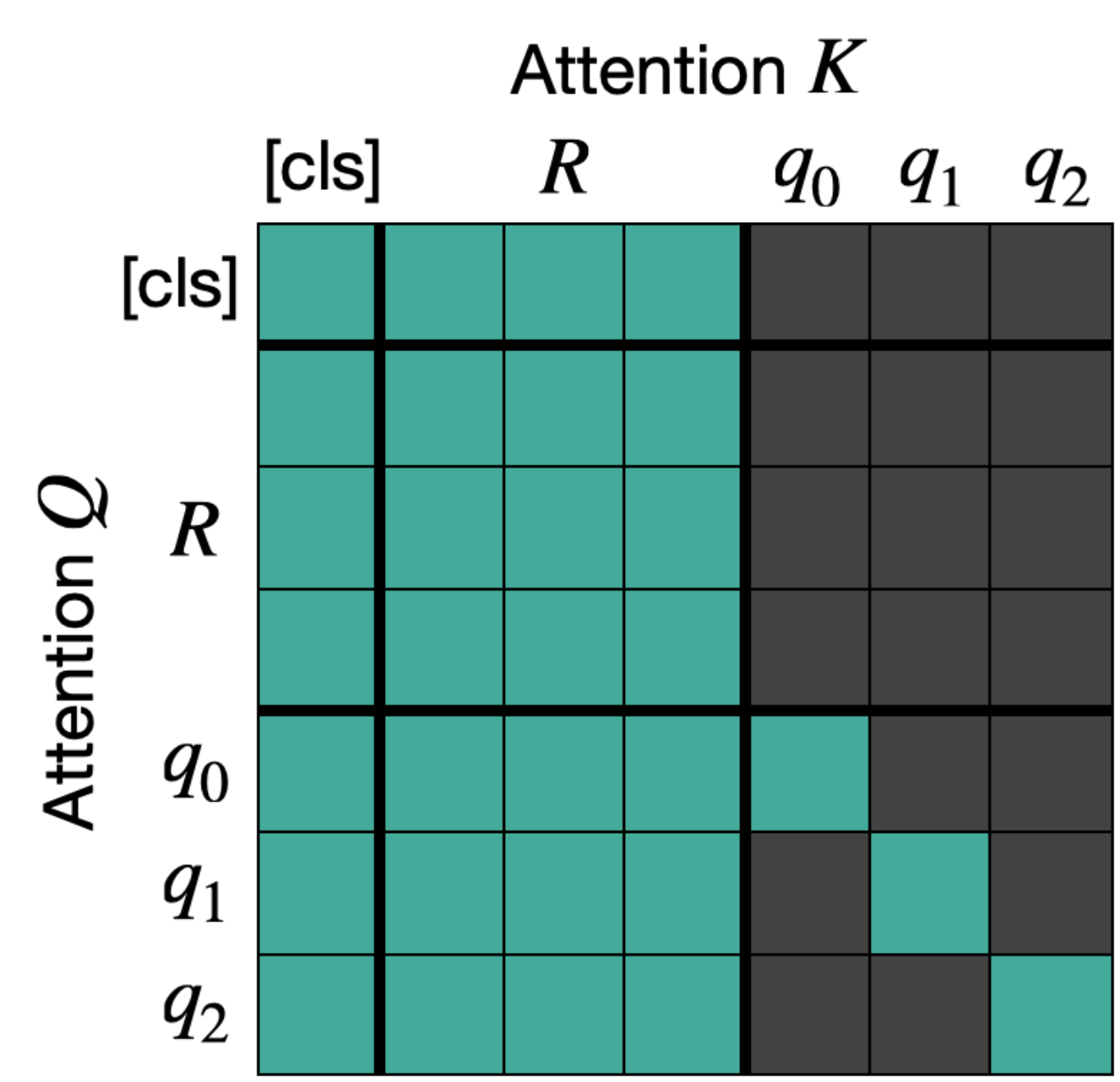}
\end{minipage}\hfill
\begin{minipage}[c]{0.45\linewidth}\vspace{1mm}
\caption{\textbf{Attention Masking Pattern in \method's Decoder.}
The masking in \method's decoder ensures a query cannot depend on another, apart from itself. 
\texttt{cls} is a learnable global summary token, following~\cite{vit,devlin2018bert}. \\
\crule[unmasked]{0.3cm}{0.3cm} Unmasked \\
\crule[masked]{0.3cm}{0.3cm} Masked
}
\label{fig:mask}
\end{minipage}
\vspace{-7mm}
\end{figure}

As described in Eq.~\ref{eq:dec}, we feed multiple queries to the decoder for efficiency via parallelization, which significantly speeds up training and inference.
However, since all tokens attend to all tokens in a standard transformer, this creates undesirable dependencies among queries.
To break the unwanted dependencies, we mask out the attention weights such that tokens cannot attend to the other queries (except for self).
This masking pattern is illustrated in \figref{mask}.

\method's attention architecture differentiates it from prior 3D reconstruction approaches.
In~\cite{mescheder2019occupancy,niemeyer2020differentiable}, points condition on a globally pooled image feature; in~\cite{yu2021pixelnerf,henzler2021unsupervised,raj2021pixel} they condition on the projected locations of the image feature map.
In \secref{exp:obj} we show that \method's design performs better.

The computation of the decoder grows with the number of queries, while the encoder embeds the input image once regardless of the final output resolution.
By using a relatively lightweight decoder, our inference is made efficient even at high resolutions, and the encoder cost is amortized.
This allows us to dynamically change output resolutions and does not require re-computing the input encoding $R$.

\subsection{Query Sampling}
\mypar{Training.}
\method samples $N^q=550$ queries from the 3D world space uniformly and per training example.
We ablate sampling strategies in \secref{exp:obj}.
A query is considered ``occupied" (positive) if it is located within radius $\tau = 0.1$ to a ground truth point, and ``unoccupied" (negative) otherwise.
The ground truth is defined as the union of all unprojected points from all RGB-D views of the scene. 

\mypar{Inference.}
We uniformly sample a grid of points covering the 3D space.
Queries with occupancy score greater than a threshold of $0.1$ and their color predictions form the final reconstruction.
Techniques such as Octree~\cite{meagher1982geometric} could be easily integrated to further speed up test-time sampling.

\subsection{Implementation Details}\label{sec:method:design}

\mypar{$E^\mathrm{XYZ}$ Patch Embeddings.}
Note that the depth values, and consequently the 3D locations in $P$, might be unknown for some points (\eg, due to sensor uncertainty).
Thus, the convolution-based patch embedding design in a ViT~\cite{vit} is not directly applicable.
We use a self-attention-based design instead.
First, the 3D coordinates are transformed.
For pixels with unknown depth, we learn a special $C$-dimensional embedding.
For pixels with valid depth, their 3D points are linearly transformed to a $C$-dimensional vector.
This results in a 16$\times$16$\times$$C$ representation for each 16$\times$16 patch.
A transformer, shared across patches, converts each patch to a $C$-dimensional vector via a learned patch token which summarizes the patch~\cite{devlin2018bert}.
This results in $\nicefrac{W}{16}\times \nicefrac{H}{16}$ tokens
(and thus $N^\mathrm{enc} = \nicefrac{W}{16}\times \nicefrac{H}{16} + 1$ with the additional global token used in a ViT~\cite{vit}).

\mypar{$E^\mathrm{RGB}$ Patch Embeddings.}
For RGB, we follow standard ViTs~\cite{vit} and embed each 16$\times$16 patch with a convolution.

\mypar{Architecture.}
The $E^\mathrm{RGB}$ and $E^\mathrm{XYZ}$ encoder use a 12-layer 768-dimensional ``ViT-Base" architecture~\cite{transformer,vit}.
The input image size is 224$\times$224.
Our decoder is a lighter-weight 8-layer 512-dimensional transformer, following MAE~\cite{he2022masked}.
Detailed specifications can be found in the Appendix.

\begin{figure*}
\tablestyle{1.0pt}{1.05}
\subfloat{%
\begin{tabular}{@{}x{39}x{39}x{39}x{39}@{}}%
\multirow{2}{*}[0mm]{\inputimg{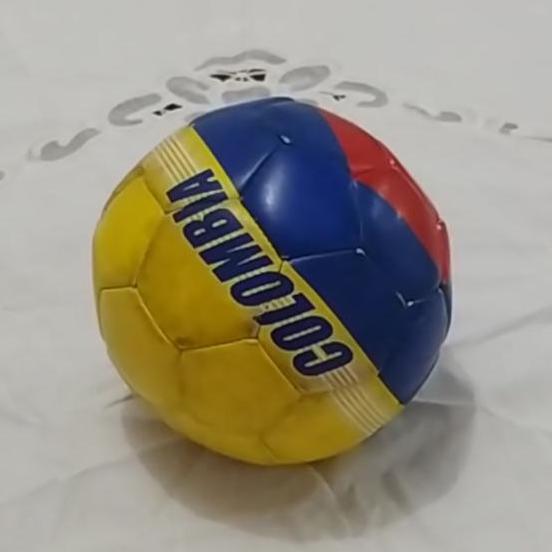}\hspace{-1.9mm}}&
\seenimg{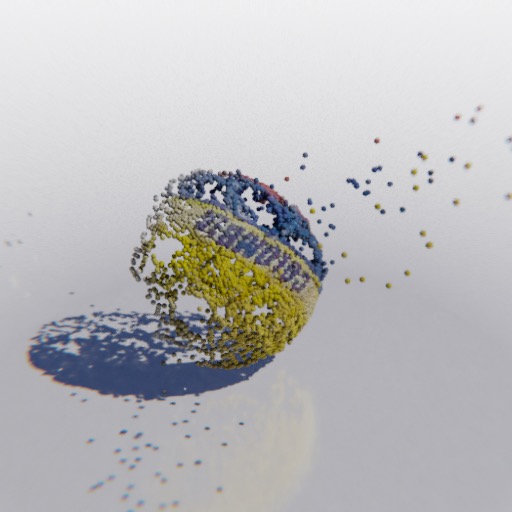}&
\myimg{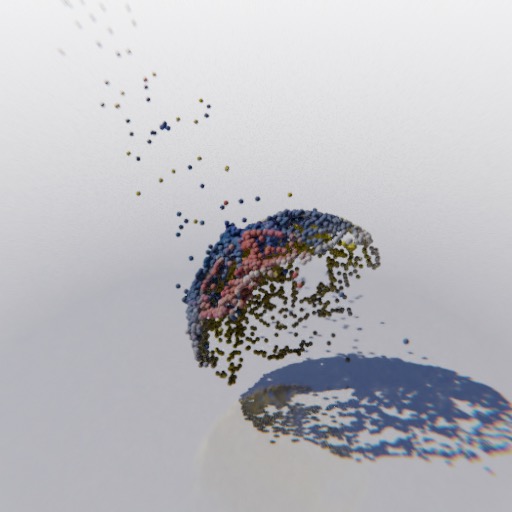}&
\myimg{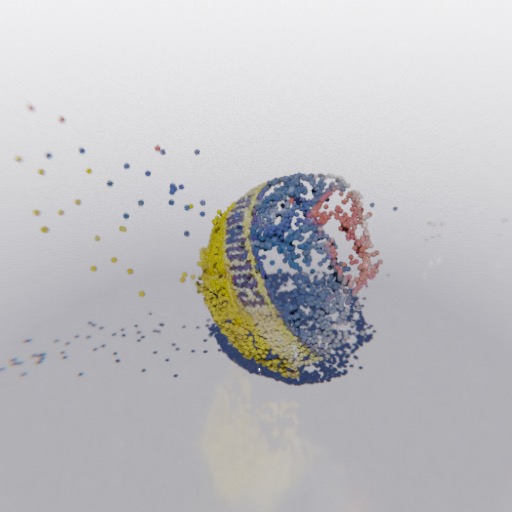}\\
&
\outimg{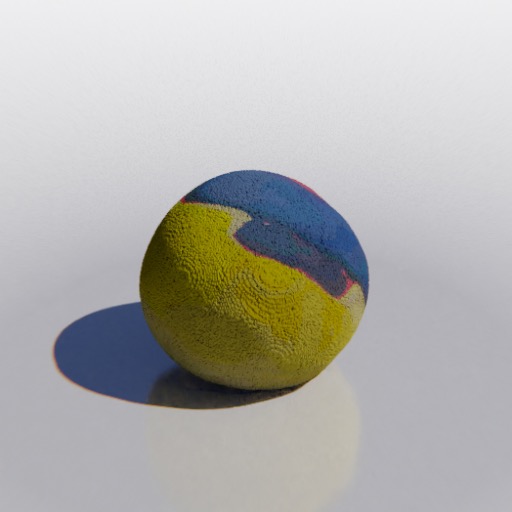}&
\myimg{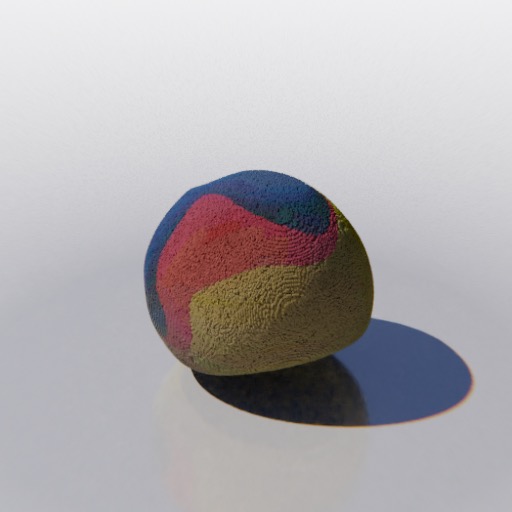}&
\myimg{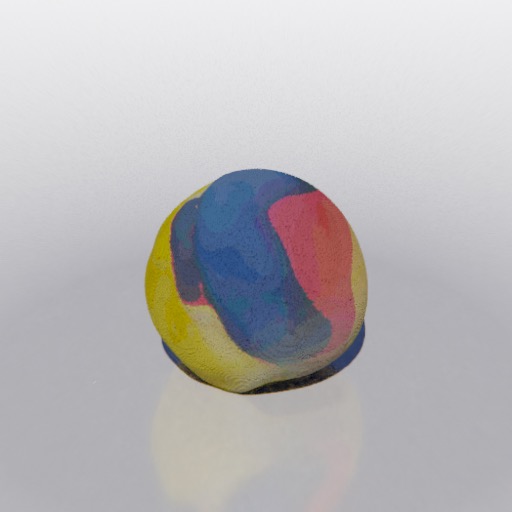}\\
\end{tabular}
}\hfill
\subfloat{%
\begin{tabular}{@{}x{39}x{39}x{39}x{39}@{}}%
\multirow{2}{*}[0mm]{\inputimg{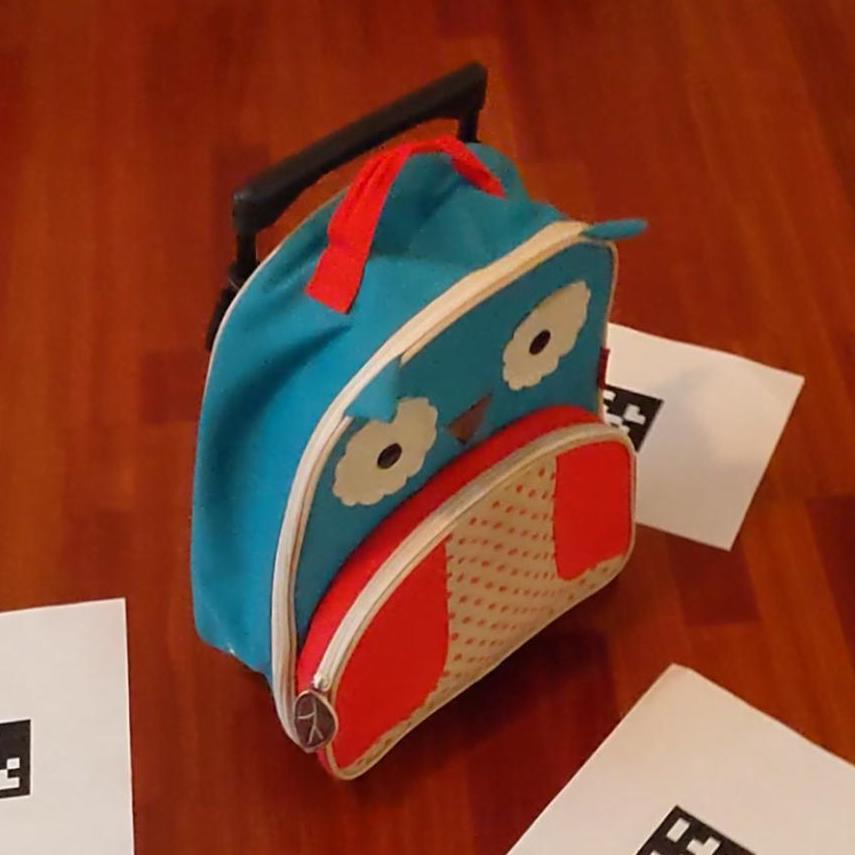}\hspace{-1.9mm}}&
\seenimg{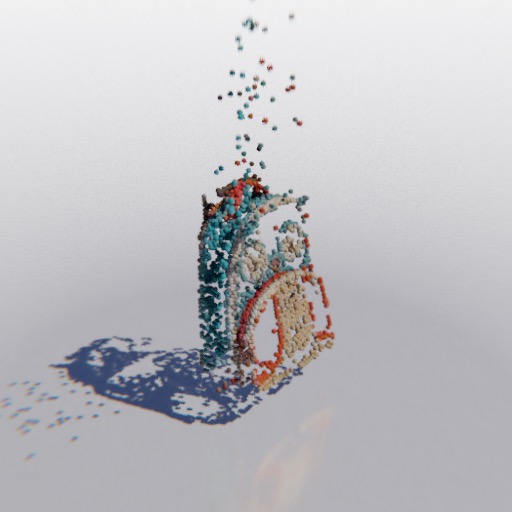}&
\myimg{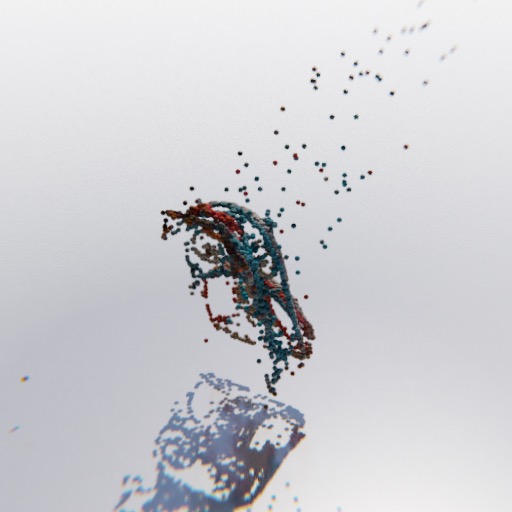}&
\myimg{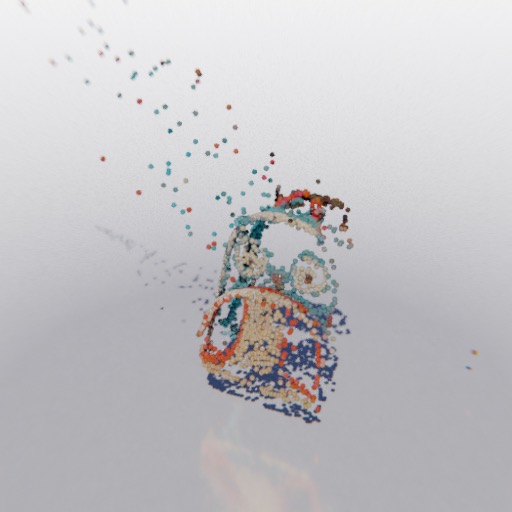}\\
&
\outimg{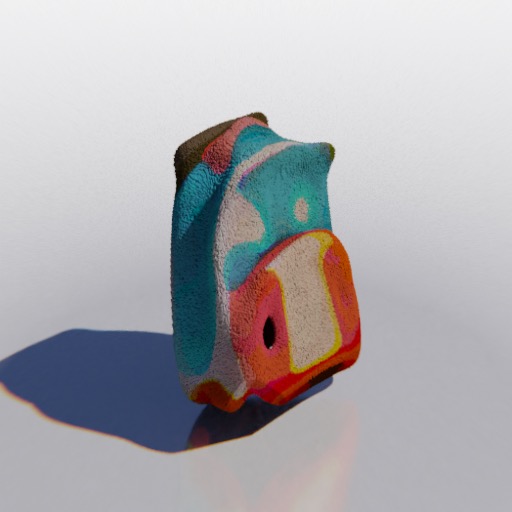}&
\myimg{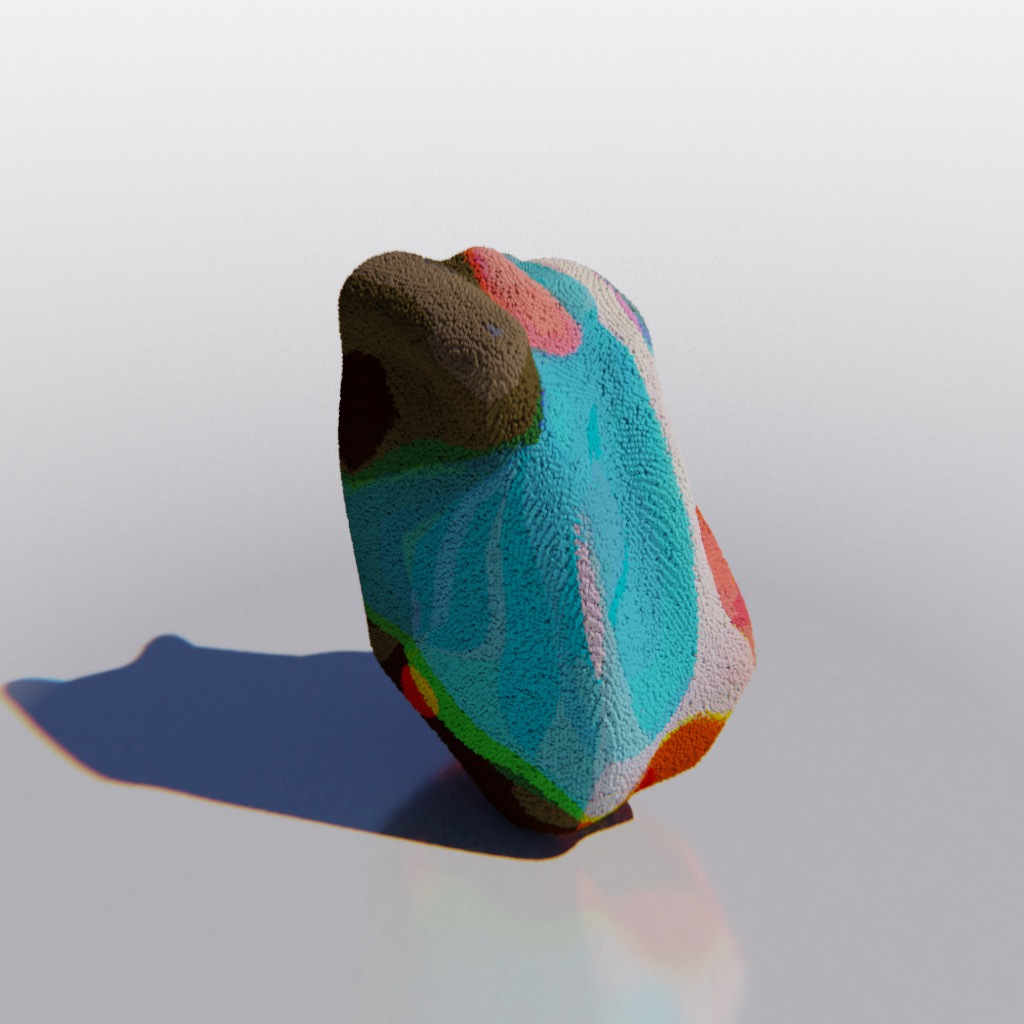}&
\myimg{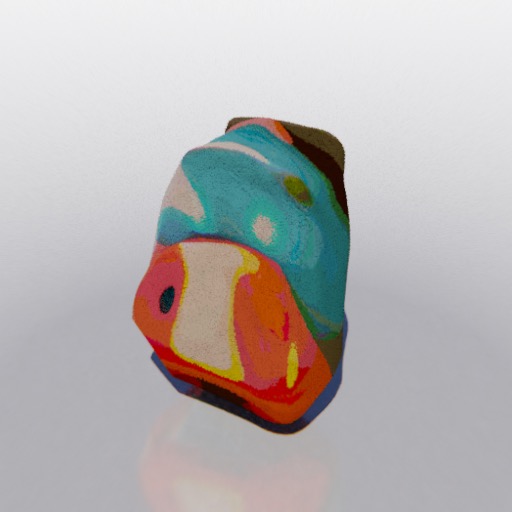}\\
\end{tabular}
}\hfill
\subfloat{%
\begin{tabular}{@{}x{39}x{39}x{39}x{39}@{}}%
\multirow{2}{*}[0mm]{\inputimg{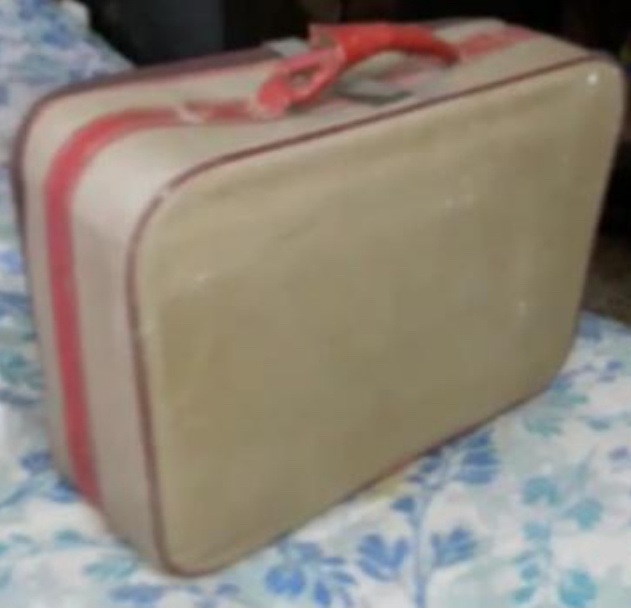}\hspace{-1.9mm}}&
\seenimg{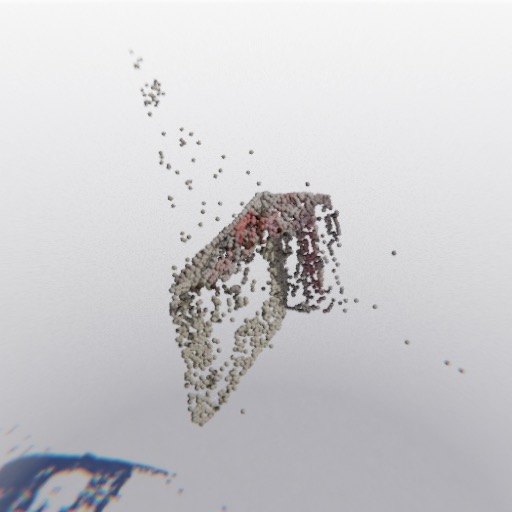}&
\myimg{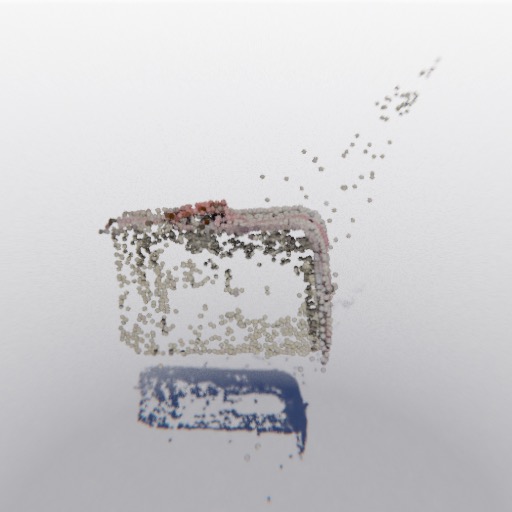}&
\myimg{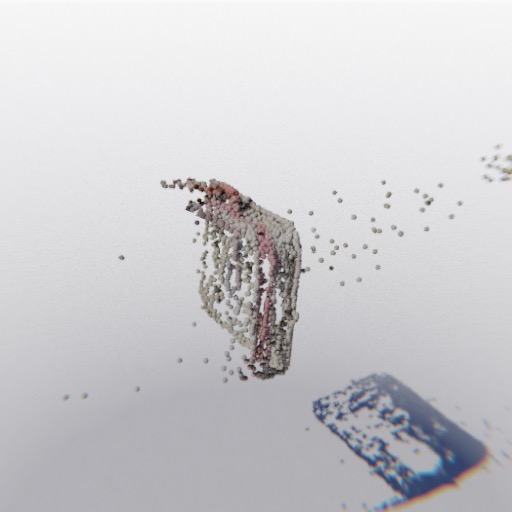}\\
&
\outimg{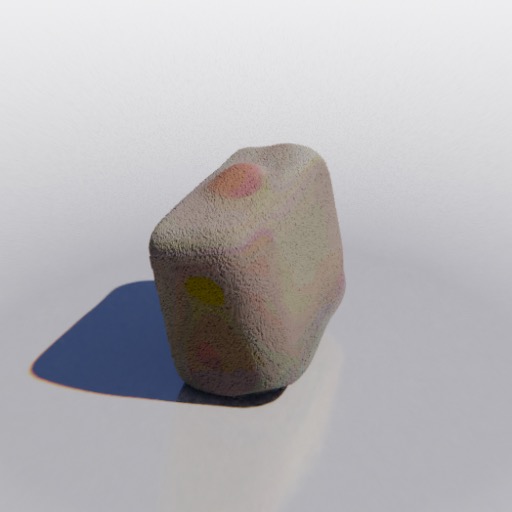}&
\myimg{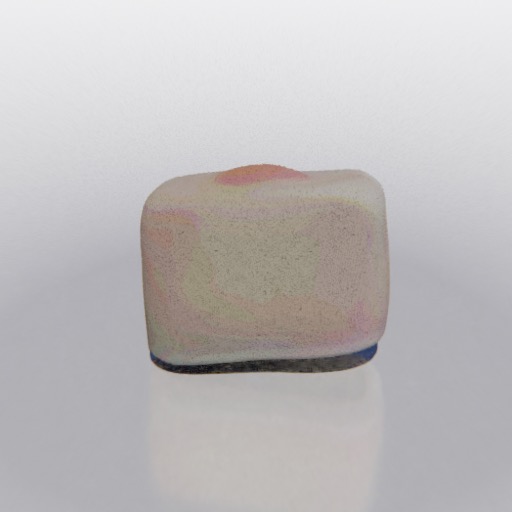}&
\myimg{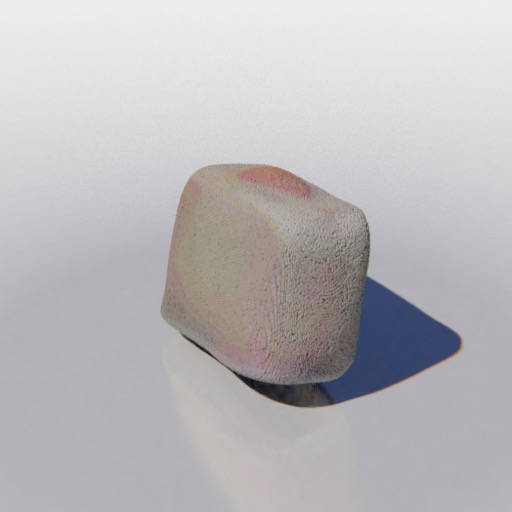}\\
\end{tabular}
}

\subfloat{%
\begin{tabular}{@{}x{39}x{39}x{39}x{39}@{}}%
\multirow{2}{*}[0mm]{\inputimg{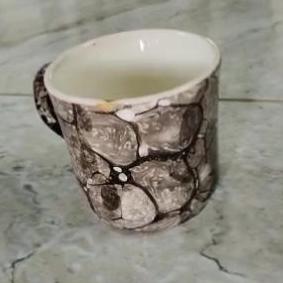}\hspace{-1.9mm}}&
\seenimg{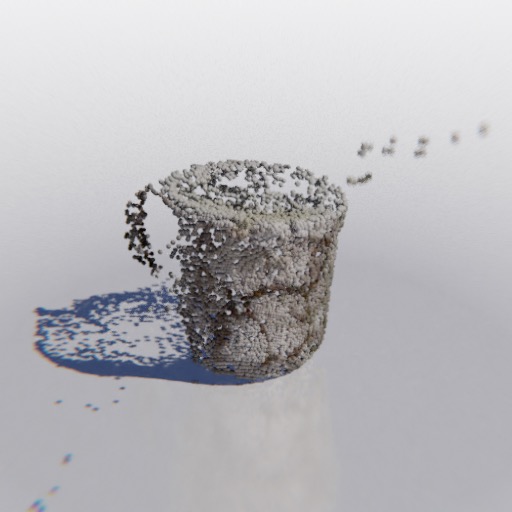}&
\myimg{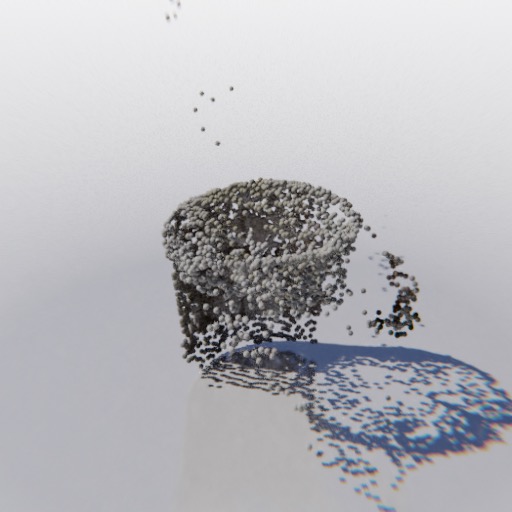}&
\myimg{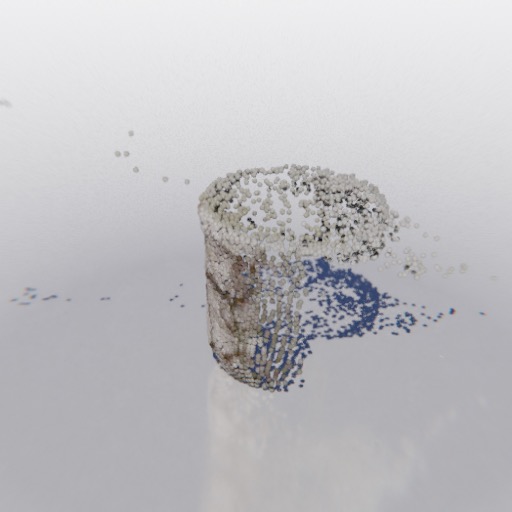}\\
&
\outimg{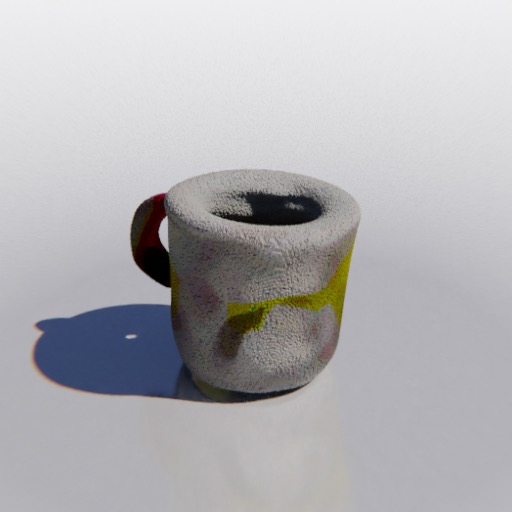}&
\myimg{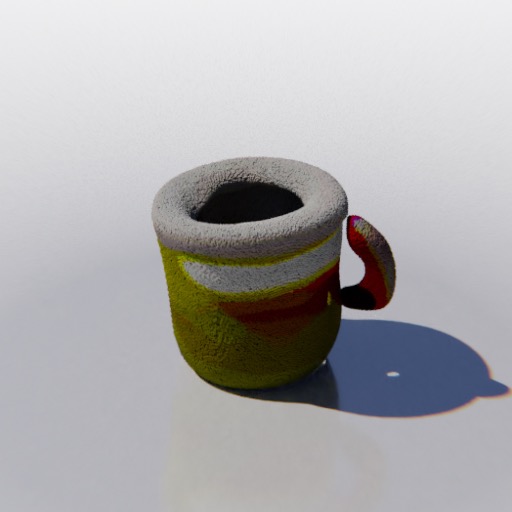}&
\myimg{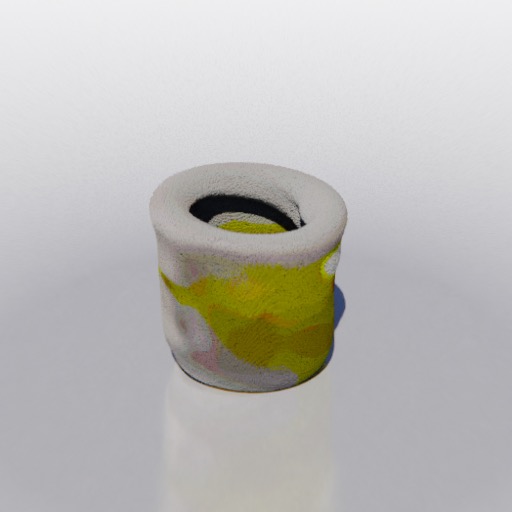}\\
\end{tabular}
}\hfill
\subfloat{%
\begin{tabular}{@{}x{39}x{39}x{39}x{39}@{}}%
\multirow{2}{*}[0mm]{\inputimg{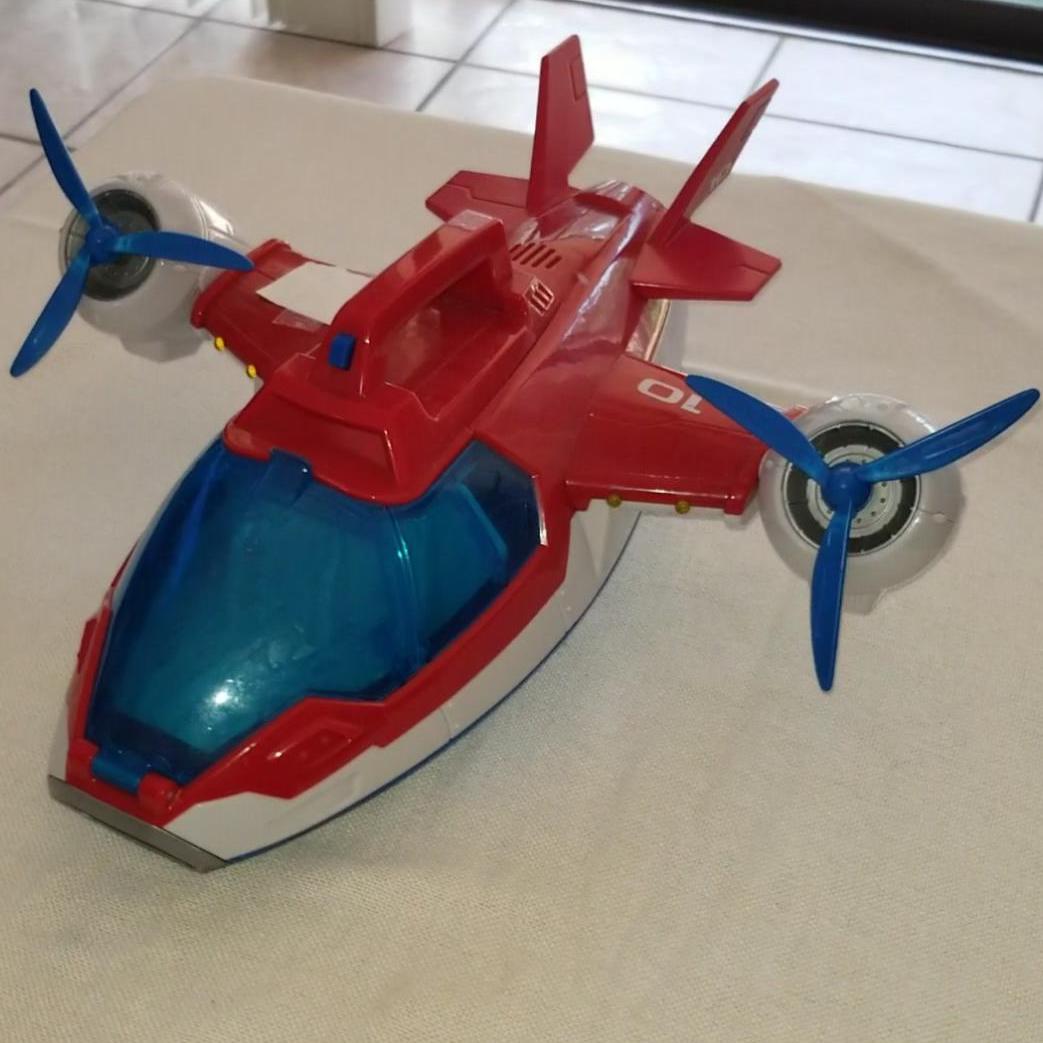}\hspace{-1.9mm}}&
\seenimg{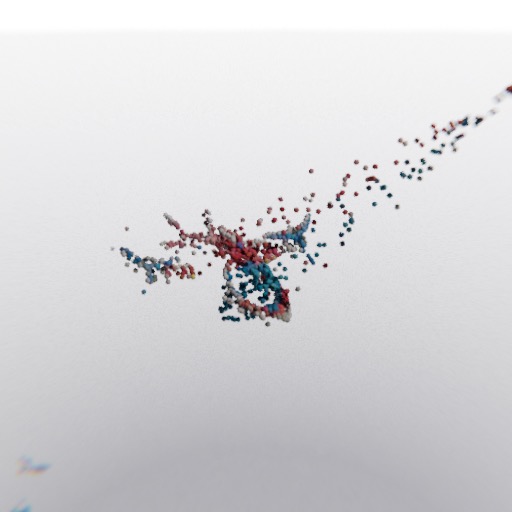}&
\myimg{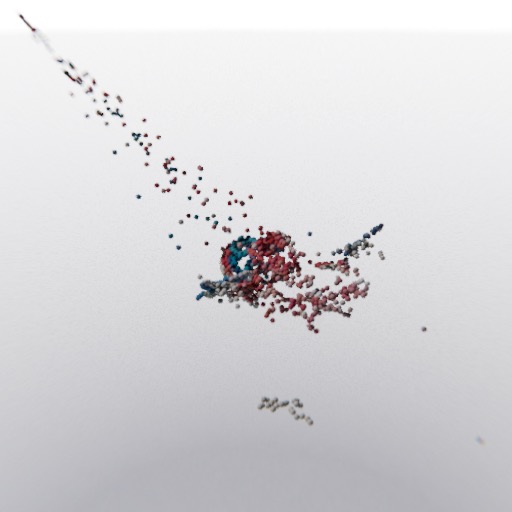}&
\myimg{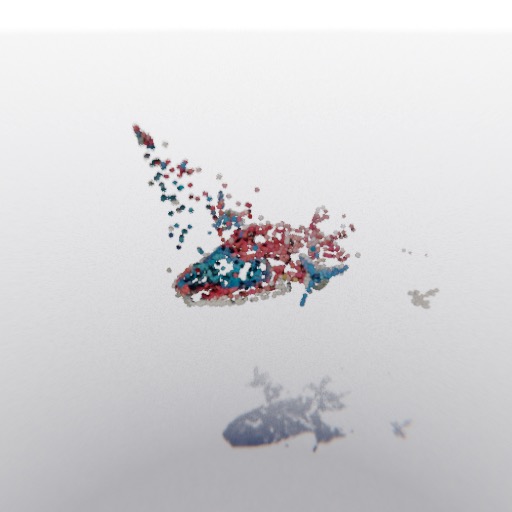}\\
&
\outimg{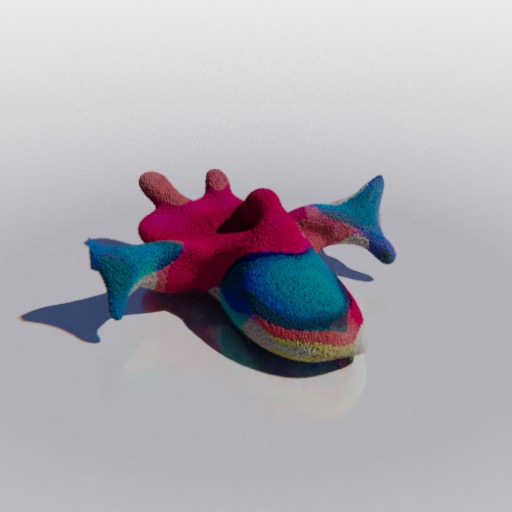}&
\myimg{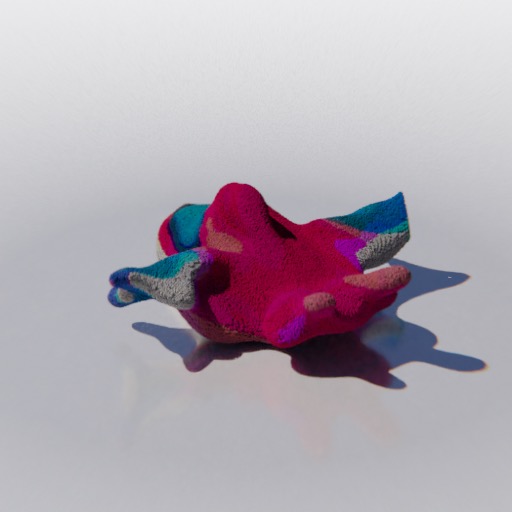}&
\myimg{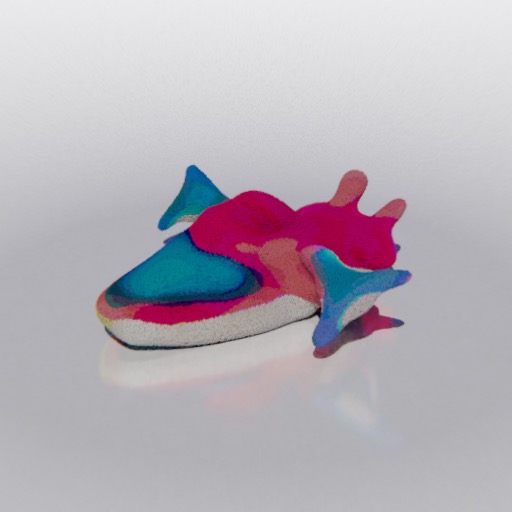}\\
\end{tabular}
}\hfill
\subfloat{%
\begin{tabular}{@{}x{39}x{39}x{39}x{39}@{}}%
\multirow{2}{*}[0mm]{\inputimg{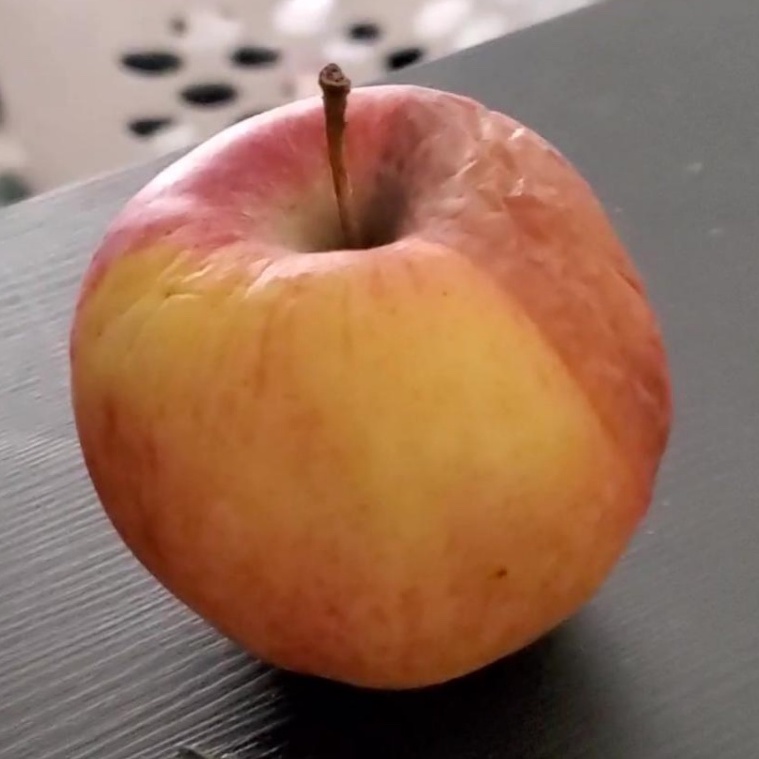}\hspace{-1.9mm}}&
\seenimg{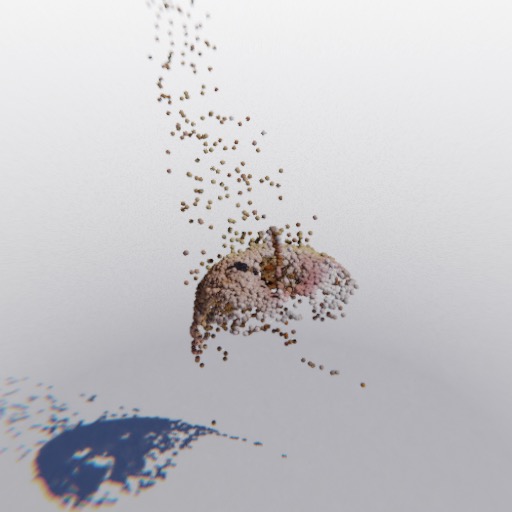}&
\myimg{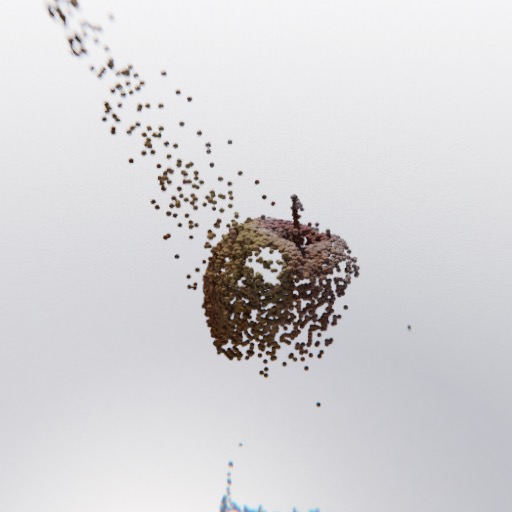}&
\myimg{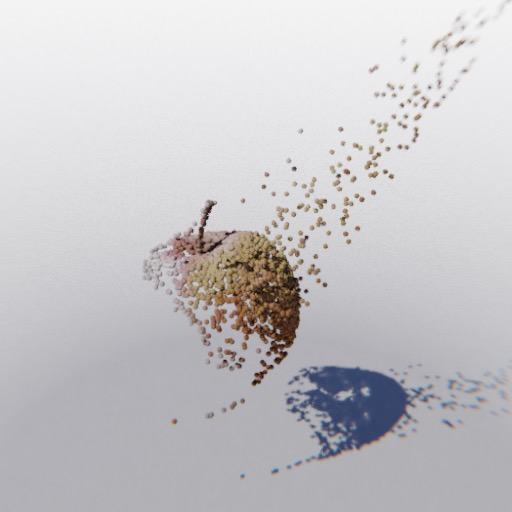}\\
&
\outimg{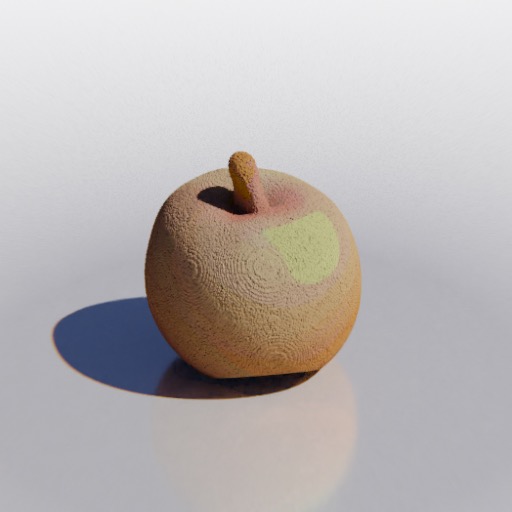}&
\myimg{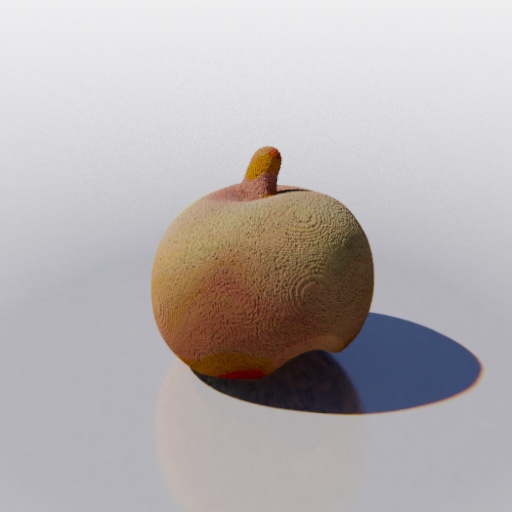}&
\myimg{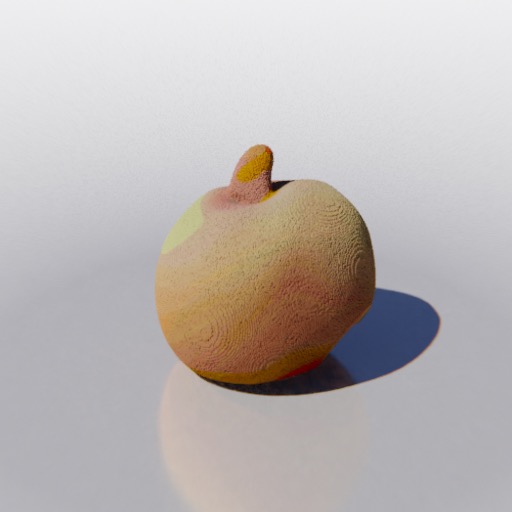}\\
\end{tabular}
}
\vspace{-3mm}
\caption{\textbf{Predictions on CO3D-v2 \emph{Novel} Categories.}
For each example, we show the input image (left), the unprojected seen 3D points (top), and our reconstruction (bottom).
We show results for a variety of object types, shapes, textures and occlusion patterns.
We emphasize that we do \emph{not} use any shape priors such as symmetries, canonical views, or mean shapes. 
See \href{https://mcc3d.github.io/}{project page} for animations. 
}\label{fig:co3dqual}
\vspace{-3mm}
\end{figure*}

\begin{table*}[t]
\subfloat[\textbf{Encoder Structure}\label{tab:abl:enc}]{%
\tablestyle{1.0pt}{1.05}
\begin{tabular}{@{}lx{28}x{28}x{28}@{}}
&Acc  & Cmp  & F1 \\
\shline
Shared & 42.6 & {77.0} & 52.5\\
\textbf{Decoupled} (ours) & 47.5 & 76.0 & \textbf{56.7}\\
\\
\end{tabular}
}\hfill%
\subfloat[\textbf{$E^\mathrm{XYZ}$ Design}\label{tab:abl:xyz}]{%
\tablestyle{1.0pt}{1.05}
\begin{tabular}{@{}lx{28}x{28}x{28}@{}}
&Acc  & Cmp  & F1 \\
\shline
MLP & 43.4 & {79.8} & 54.5\\
PointNet~\cite{qi2017pointnet} & 45.6 & 80.3 & \underline{56.6}\\
\textbf{Transformer} (ours) & 47.5 & 76.0 & \textbf{56.7}
\end{tabular}
}\hfill
\subfloat[\textbf{Training Query Sampling}\label{tab:abl:sampling}]{%
\tablestyle{1.0pt}{1.05}
\begin{tabular}{@{}lx{28}x{28}x{28}@{}}
&Acc  & Cmp  & F1 \\
\shline
Contrastive& 45.0 & 78.7 & {55.6}\\
\textbf{Uniform} (ours) & 47.5 & 76.0 & \textbf{56.7}\\
\\
\end{tabular}
}%
\vspace{-2mm}

\center
\subfloat[\textbf{Feature Conditioning}\label{tab:abl:feat}]{%
\tablestyle{1.0pt}{1.05}
\begin{tabular}{@{}lx{27}x{27}x{27}@{}}
&Acc  & Cmp  & F1 \\
\shline
Loc-pooled & 49.2 & 22.6 & 28.2\\
Global & 44.7 & 77.1 & 54.5\\
\textbf{Detailed} (ours) & 47.5 & 76.0 & \textbf{56.7}\\
\end{tabular}
}%
\hfill
\subfloat[\textbf{Decoder Design}\label{tab:abl:dec}]{%
\tablestyle{1.0pt}{1.05}
\begin{tabular}{@{}lx{27}x{27}x{27}@{}}
&Acc  & Cmp  & F1 \\
\shline
Loc+MLP & 49.2 & 22.6 & 28.2\\
Cross-attn & 42.3 & 49.5 & 43.7\\
\textbf{Concat+attn} (ours) & 47.5 & 76.0 & \textbf{56.7}\\
\end{tabular}
}
\hfill
\subfloat[\textbf{Comparison to Prior Work with Explicit Design}\label{tab:abl:explicit}]{%
\tablestyle{1.0pt}{1.05}
\begin{tabular}{@{}lx{27}x{27}x{27}x{27}@{}}
&Acc  & Cmp  & F1 & CD ($\downarrow$)\\
\shline
PoinTr~\cite{yu2021pointr} & 79.6 & 27.1 & 39.7 & 0.065\\
{MCC (w/o RGB)} & 46.5 & 70.8 & 53.9 & 0.047\\
\textbf{MCC} & 47.5 & 76.0 & \textbf{56.7} & \textbf{0.040}\\
\end{tabular}
}
\vspace{-3mm}
\caption{\textbf{Ablations on CO3D-v2}, which validate \method's design choices. We highlight ablation (e) which shows that an attention-based decoder outperforms an MLP, and (f) where we find that \method's queriable decoder performs better than an explicit design~\cite{yu2021pointr}. Higher is better for Accuracy (Acc), Completeness (Cmp), and F1. Lower is better for Chamfer distance (CD).}
\label{tab:abl}
\vspace{-4mm}
\end{table*}

\section{Object Reconstruction Experiments}
\label{sec:exp:obj}

\method works naturally for both objects and scenes.
In \secref{exp:obj}, we show results and compare to competing methods for single object reconstruction.
In \secref{exp:scene}, we show results on scenes.

\mypar{Dataset.}
We use CO3D-v2~\cite{reizenstein21co3d} as our main dataset for single object reconstruction.
It consists of $\app$37k short videos of 51 object categories; the largest dataset of 3D objects in the wild.
To show generalization to new objects, we hold out 10 randomly selected categories for evaluation and train on the remaining 41.
The list of held-out categories is available in the Appendix.
Since CO3D is object-centric, we focus on foreground objects specified by segmentation masks provided in CO3D.
Full 3D annotations, such as 3D meshes, are not available.
CO3D extracts point clouds from the videos via COLMAP~\cite{schoenberger2016sfm,schoenberger2016mvs}, which are inevitably noisy and are used to train our model.
Despite imperfect supervision, we show that \method learns to reconstruct 3D shapes and texture and even corrects the noisy depth inputs.

\mypar{Metrics.}
Following Kulkarni \etal~\cite{kulkarni2021drdf}, we report: accuracy (acc), the percentage of predicted points within $\rho$ to a ground truth point, completeness (cmp), the percentage of ground truth points within $\rho$ from a predicted point, and their F-score (F1) which drives our comparisons. $\rho$ is $0.1$.

\mypar{Training Details.}
We train with Adam~\cite{kingma2014adam} for 150k iterations with an effective batch size of 512 using 32 GPUs, a base learning rate of 10$^{-4}$ with a cosine schedule and a linear warm-up for the first 5\% of iterations.
Training takes $\app$2.5 days.
We randomly scale augment images by $s \in \sbr{0.8, 1.2}$. 
We also perform 3D augmentations by randomly rotating 3D points along each axis by $\theta \in \sbr{-180^o, 180^o}$.
Rotation is applied to the \emph{seen} points $P$, the queries and the ground truth. 
Image $I$ and points $P$ are aligned through the concatenation of their encodings (Eq.~\ref{eq:enc}). 
Points $P$ and queries are consistent as well, as both are rotated. 
Essentially, our 3D augmentations build in rotation equivariance.

\mypar{Coordinate System.}
We adopt the original CO3D coordinate system from~\cite{reizenstein21co3d}, where objects are normalized to have zero-mean and unit-variance.
Training and testing points are sampled from $\sbr{-3, 3}$ along each axis.
Evaluation points are sampled with a granularity of 0.1.

\subsection{Qualitative Results on Novel Categories}
\label{sec:co3dv2qual}
\figref{co3dqual} shows qualitative results on the CO3D test set of novel categories.
We show reconstructions for a variety of shapes and object types.
\method tackles heavy self-occlusions, \eg, the mug handle is barely visible in the input image, and complex shapes, \eg, the toy airplane.
In addition to shape, \method predicts texture which is difficult especially for unseen regions.
For instance, the left and back side of the kids backpack is completely invisible, but \method predicts to propagate the color from the right side.
We also note that \method is robust to noisy depth from COLMAP, present at varying degrees and depicted in the \emph{seen} points of each example (top row).
\method corrects and completes the geometry in spite of the noise in depth inputs.
We emphasize that we do not make geometric assumptions nor use any priors such as symmetry or mean templates when reconstructing objects. 
MCC learns only from data.

\subsection{Ablation Study}
\mypar{Encoder Structure.}
In \tabref{abl:enc}, we ablate our encoder design which models $I$ and $P$ with two separate transformers (decoupled) and compare to a shared transformer which models the fused (sum) patch embeddings of $I$ and $P$ (shared).
Our decoupled design performs slightly better.

\mypar{$E^\mathrm{XYZ}$ Design.}
\tabref{abl:xyz} compares our transformer to an MLP and PointNet~\cite{qi2017pointnet} for the $E^\mathrm{XYZ}$ encoder.
PointNet and our transformer, which model point interactions, work slightly better than an MLP, though not critically.

\mypar{Training Query Sampling.}
In \tabref{abl:sampling}, we compare our uniform sampling strategy with a contrastive-style sampling, 
where each example samples a fixed number of positives and negatives.
Both work similarly. We choose uniform sampling because of its simplicity.

\mypar{Feature Conditioning.}
Our input encoding $R$ uses all $N^\mathrm{enc}$ tokens from the appearance $I$ and geometry $P$ encodings.
We call this detailed conditioning and compare it with two popular choices: one where a globally average-pooled vector is used, as in \cite{mescheder2019occupancy,niemeyer2020differentiable}, and one where the feature vector is bilinearly interpolated at the projected location in the feature map, as in~\cite{yu2021pixelnerf,henzler2021unsupervised,raj2021pixel}.
\tabref{abl:feat} validates our choice.

\mypar{Decoder Design.}
As described in \secref{method}, \method's decoder concatenates queries to the input encoding $R$ in the token dimension, and a transformer models their interactions (concat+attn).
We compare this design with two popular ones.
Recent works on image-conditioned NeRF~\cite{yu2021pixelnerf,henzler2021unsupervised,raj2021pixel} condition points on their projected location in the feature map followed by an MLP (loc+MLP) -- this comparison was also presented in the context of feature conditioning strategies.
Another approach is cross-attention (cross-attn), where the encoded input $R$ only serves as \emph{keys}/\emph{values} but not as \emph{queries} to a transformer, \eg, in Perceiver models~\cite{jaegle2021perceiver,jaegle2021perceiverio}.
\tabref{abl:dec} shows that our decoder is critical for performance.

\begin{figure}[t]
\centering
\tablestyle{1.0pt}{1.05}
\subfloat{%
\resizebox{\linewidth}{!}{
\begin{tabular}{@{}cx{50}x{50}x{50}x{50}@{}}
Input image & Seen & PoinTr~\cite{yu2021pointr} & \method w/o RGB & \method\\
\frame{\includegraphics[height=17.5mm]{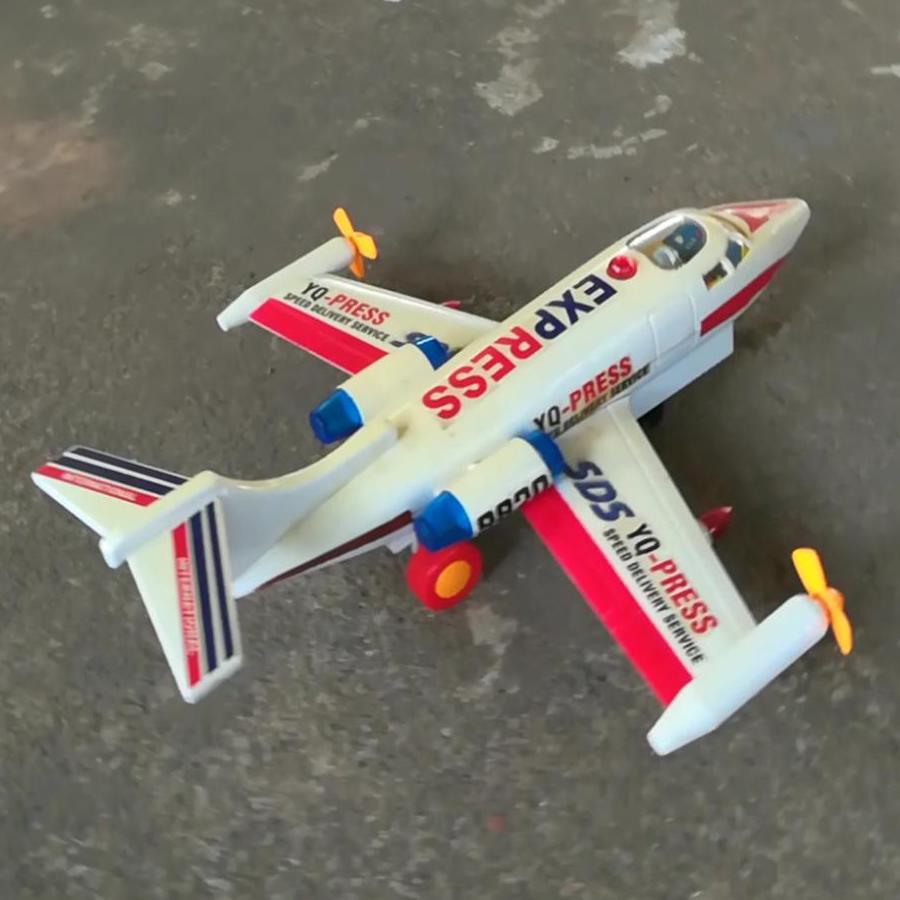}}&
\frame{\includegraphics[height=17.5mm]{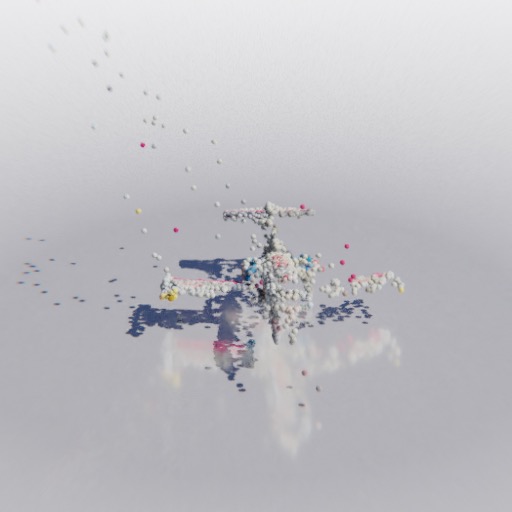}}&
\frame{\includegraphics[height=17.5mm]{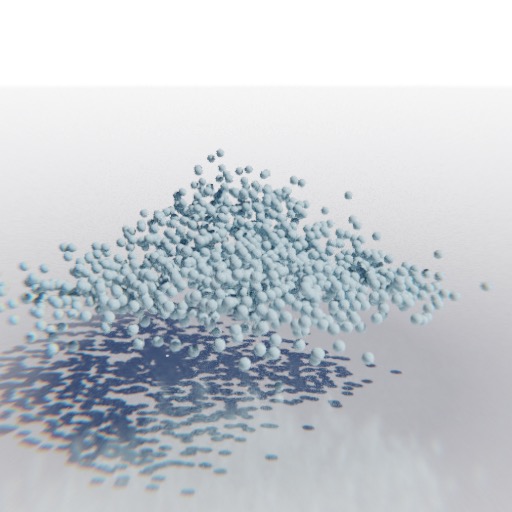}}&
\frame{\includegraphics[height=17.5mm]{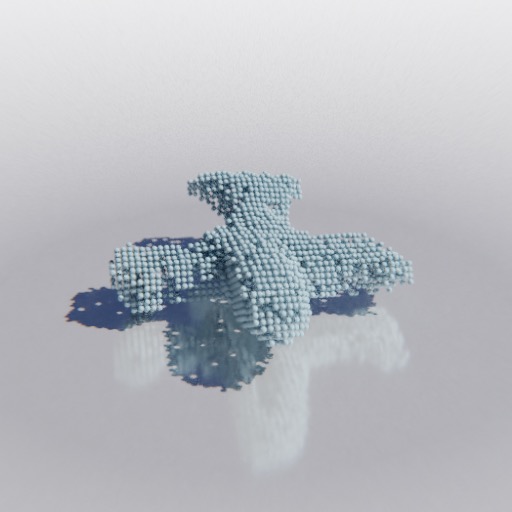}}&
\frame{\includegraphics[height=17.5mm]{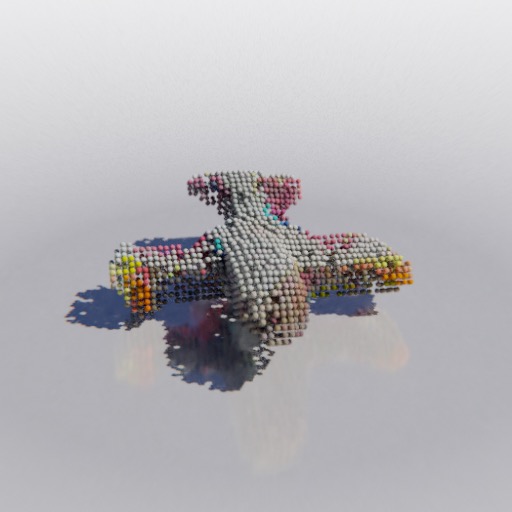}}\\
\frame{\includegraphics[height=17.5mm]{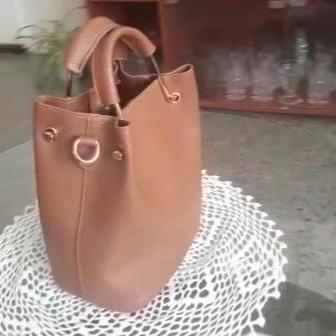}}&
\frame{\includegraphics[height=17.5mm]{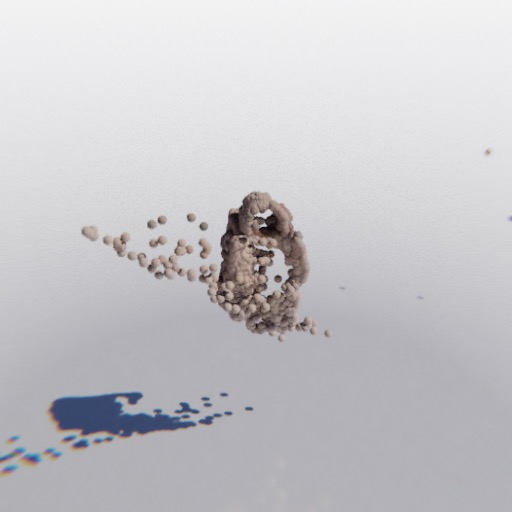}}&
\frame{\includegraphics[height=17.5mm]{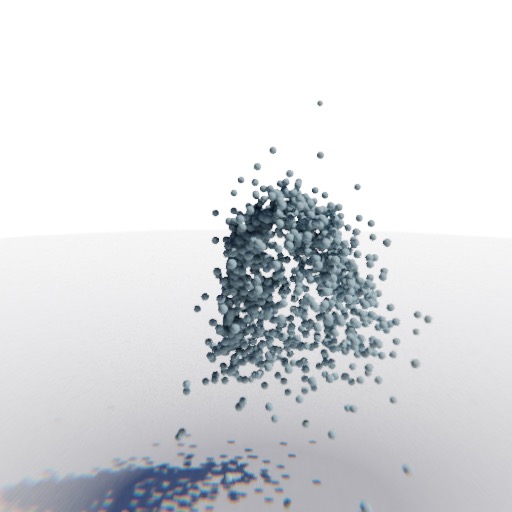}}&
\frame{\includegraphics[height=17.5mm]{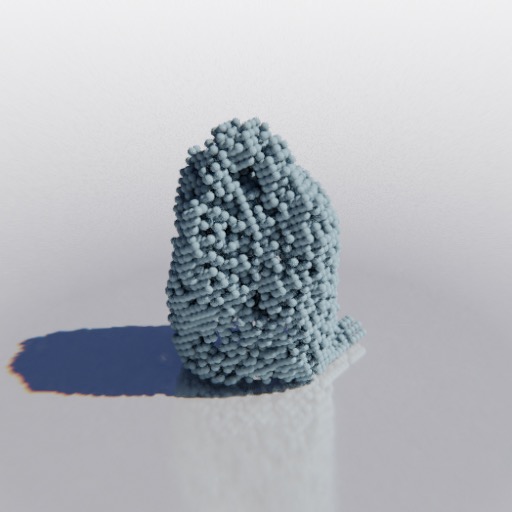}}&
\frame{\includegraphics[height=17.5mm]{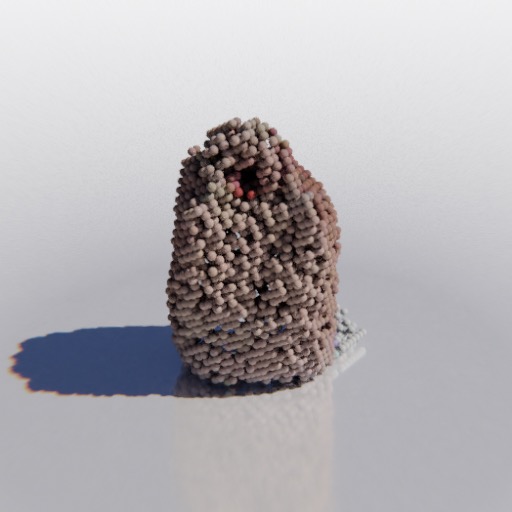}}\\
\end{tabular}
}}
\vspace{-4mm}
\caption{\textbf{Qualitative Comparison to PoinTr~\cite{yu2021pointr}.} 
\method predicts shape details while PoinTr tends to place points roughly around the object.
For a fair comparison, \method predicts the same number of points as PoinTr.
Unlike PoinTr, \method also predicts color.}
\label{fig:pointr}
\vspace{-4mm}
\end{figure}

\mypar{Comparison to Prior Work with an Explicit Design.}
Finally, we compare \method and its queriable 3D decoder with a state-of-the-art 3D point completion method PoinTr~\cite{yu2021pointr}.
PoinTr inputs an incomplete point cloud and predicts a fixed-resolution output using a transformer which models explicit geometric point relations (via nearest neighbors).
We train PoinTr on CO3D which inputs the set of \emph{seen} points $P$.
For a fair comparison, we implement PoinTr with the same 12-layer architecture as ours, which is stronger than their 6-layer one.
Since PoinTr does not use RGB, we compare with a \method variant that ignores texture by encoding $P$ but not $I$.
We additionally report chamfer distance (CD), as in~\cite{yu2021pointr}, and use the same number of points for a fair comparison.
\tabref{abl:explicit} shows that \method outperforms PoinTr by a large margin.
\figref{pointr} presents a qualitative comparison.
In \secref{nerf}, we also compare to NeRF-based methods.

\begin{figure}[t]
\small
\resizebox{0.498\linewidth}{!}{
\subfloat[\textbf{Scaling \# examples (same categories)}\label{fig:scaling:example}]{%
	\begin{tikzpicture}
	\begin{axis}[
	clip=false,
	width=0.51\linewidth,
	height=0.495\linewidth,
	xlabel={\# examples (\%)},
	ylabel={F1 on held-out cat. (\%)},
	xtick = {25, 50, 75, 100},
	xticklabels = {25, 50, 75, 100},
	every axis plot/.append style={very thick,mark options={scale=0.5, solid}},
	ymin=49,
	ymax=56,
	xmax=120,
	cycle multiindex* list={%
		mycolormarklist
	}
	]
	\addplot+[skyblue2,mark=diamond*,mark options={solid}]
	table[row sep=crcr] {
		x y\\
		25 51.6\\
		50 53.2\\
		75 53.6\\
		100 54.8\\
	};
	\addplot[mark=none, dashed, line width=0.5pt] coordinates {(25,51.6) (110,51.6)};
    \draw[->, line width=0.7pt](axis cs:100,51.6)--(axis cs:100,54.6);
    \node[anchor=west] at (axis cs:100,53.6) {+3.2\%};
	\end{axis}
	\end{tikzpicture}
}}\hfill
\resizebox{0.498\linewidth}{!}{
\subfloat[\textbf{Scaling \# categories}\label{fig:scaling:cat}]{%
	\begin{tikzpicture}
	\begin{axis}[
	clip=false,
	width=0.51\linewidth,
	height=0.495\linewidth,
	xlabel={\# categories (\%)},
	ylabel={F1 on held-out cat. (\%)},
	xtick = {25, 50, 75, 100},
	xticklabels = {25, 50, 75, 100},
	every axis plot/.append style={very thick,mark options={scale=0.5, solid}},
	ymin=49,
	ymax=56,
	xmax=120,
	cycle multiindex* list={%
		mycolormarklist
	}
	]
	\addplot+[skyblue2,mark=diamond*,mark options={solid}]
	table[row sep=crcr] {
		x y\\
		25 49.8\\
		50 51.3\\
		75 54.2\\
		100 54.8\\
	};
	\addplot[mark=none, dashed, line width=0.5pt] coordinates {(25,49.8) (110,49.8)};
    \draw[->, line width=0.7pt](axis cs:100,49.8)--(axis cs:100,54.6);
    \node[anchor=west] at (axis cs:100,53.6) {+5.0\%};
	\end{axis}
	\end{tikzpicture}
}}
\vspace{-2mm}
\caption{\textbf{Scaling Behavior Analysis.}
We train \method on (a) a varying number of examples uniformly sampled from all training categories
and (b) all examples from a varying number of training categories.
All models are evaluated on the same held-out set of novel categories.
We see clear performance gains from scaling training data, especially when expanding the number of categories.
This supports that category-agnostic models and large-scale training are promising for 3D reconstruction.}
\label{fig:scaling}
\vspace{-3mm}
\end{figure}
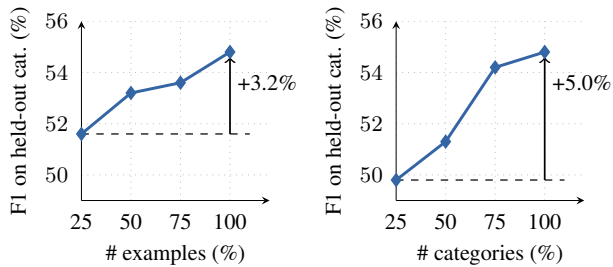

\begin{figure*}[t]
\vspace{-1mm}
\tablestyle{1.0pt}{1.05}
\hspace{-3mm}
\subfloat[iPhone\label{fig:iphone}]{%
\resizebox{0.333\linewidth}{!}{
\begin{tabular}{@{}x{45}x{45}x{45}x{45}@{}}
\multirow{2}{*}{\inputimg{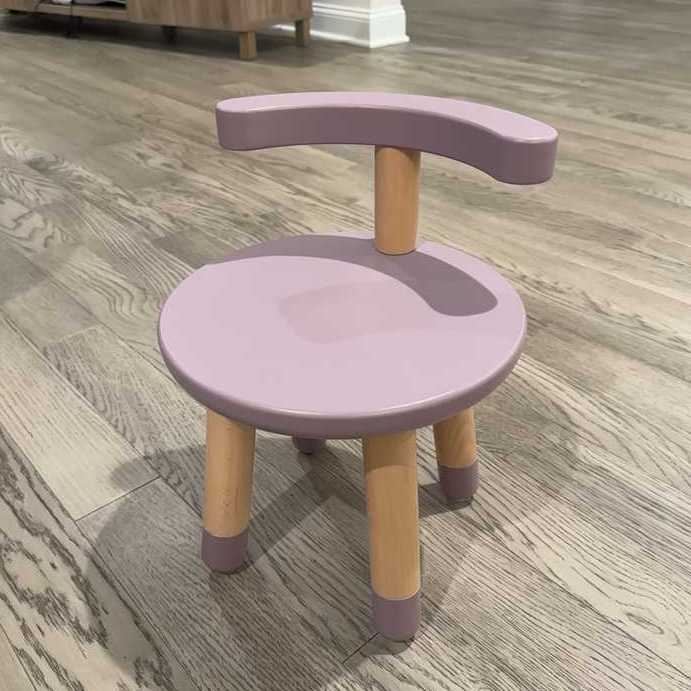}\hspace{-1.9mm}}&
\seenimg{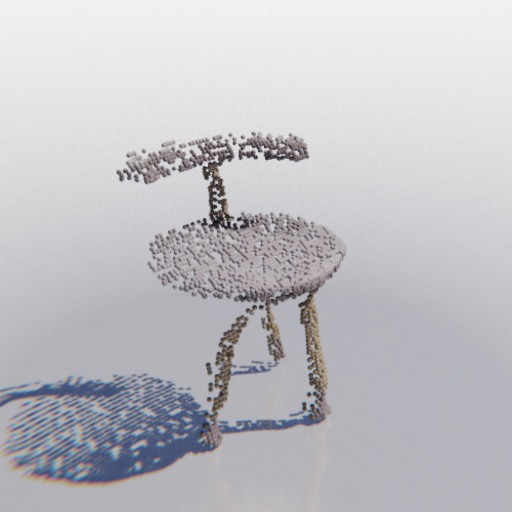}&
\myimg{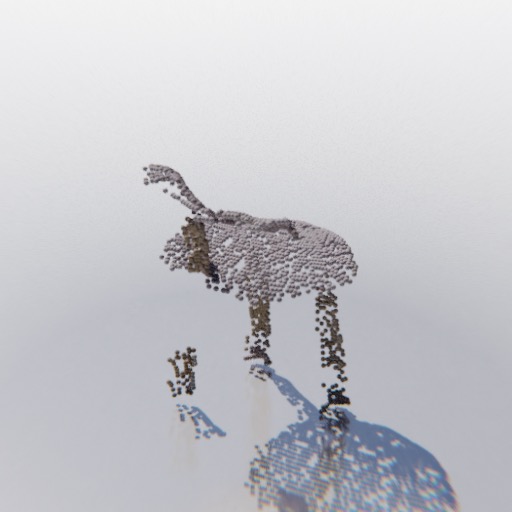}&
\myimg{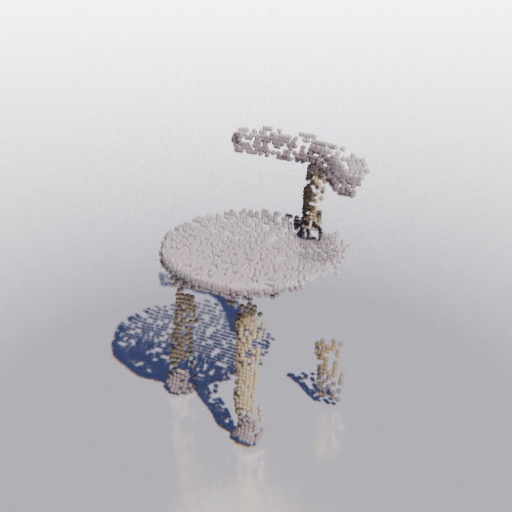}\\
&
\outimg{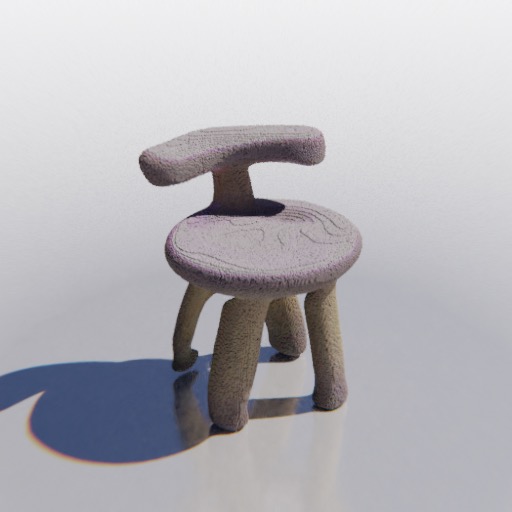}&
\myimg{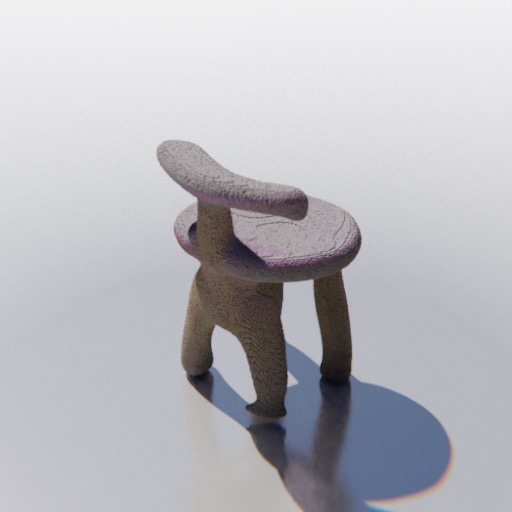}&
\myimg{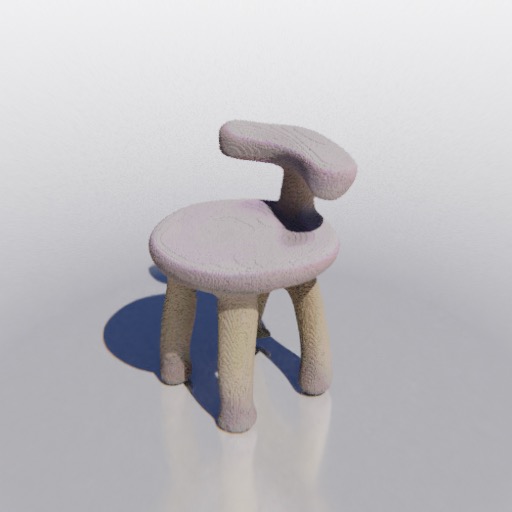}\\
\multirow{2}{*}{\inputimg{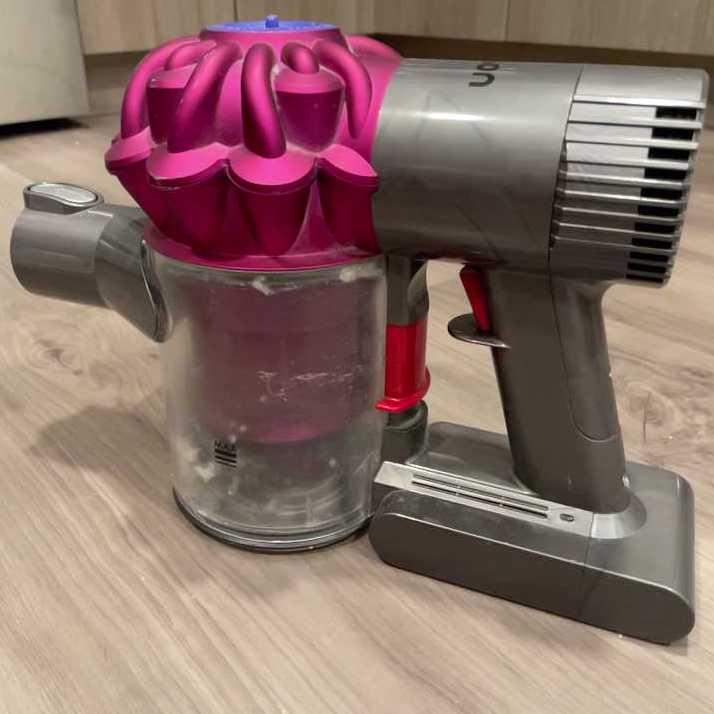}\hspace{-1.9mm}}&
\seenimg{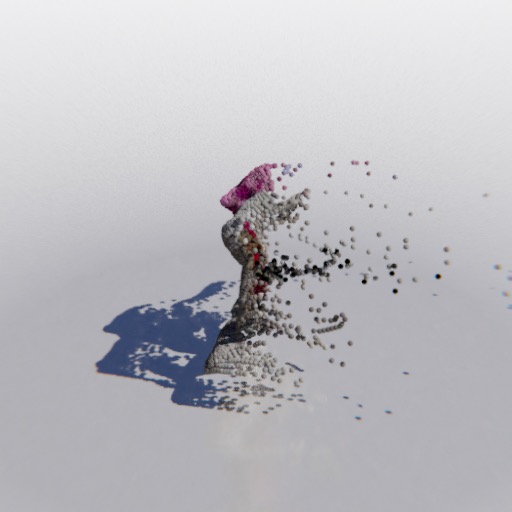}&
\myimg{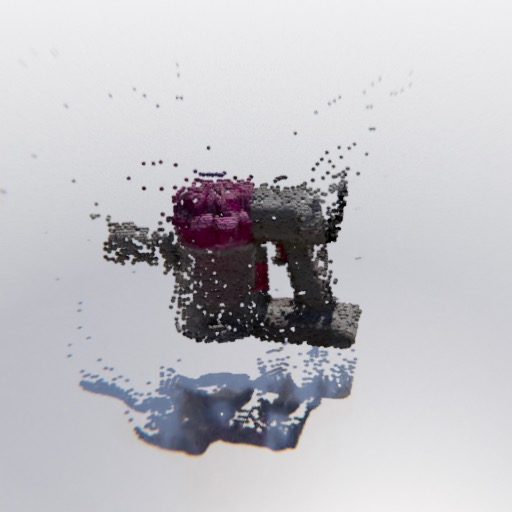}&
\myimg{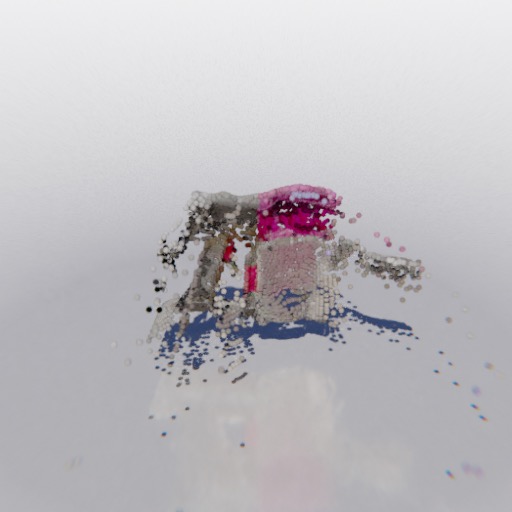}\\
&
\outimg{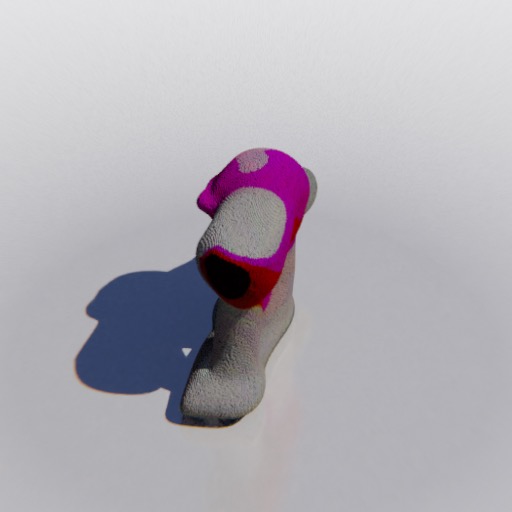}&
\myimg{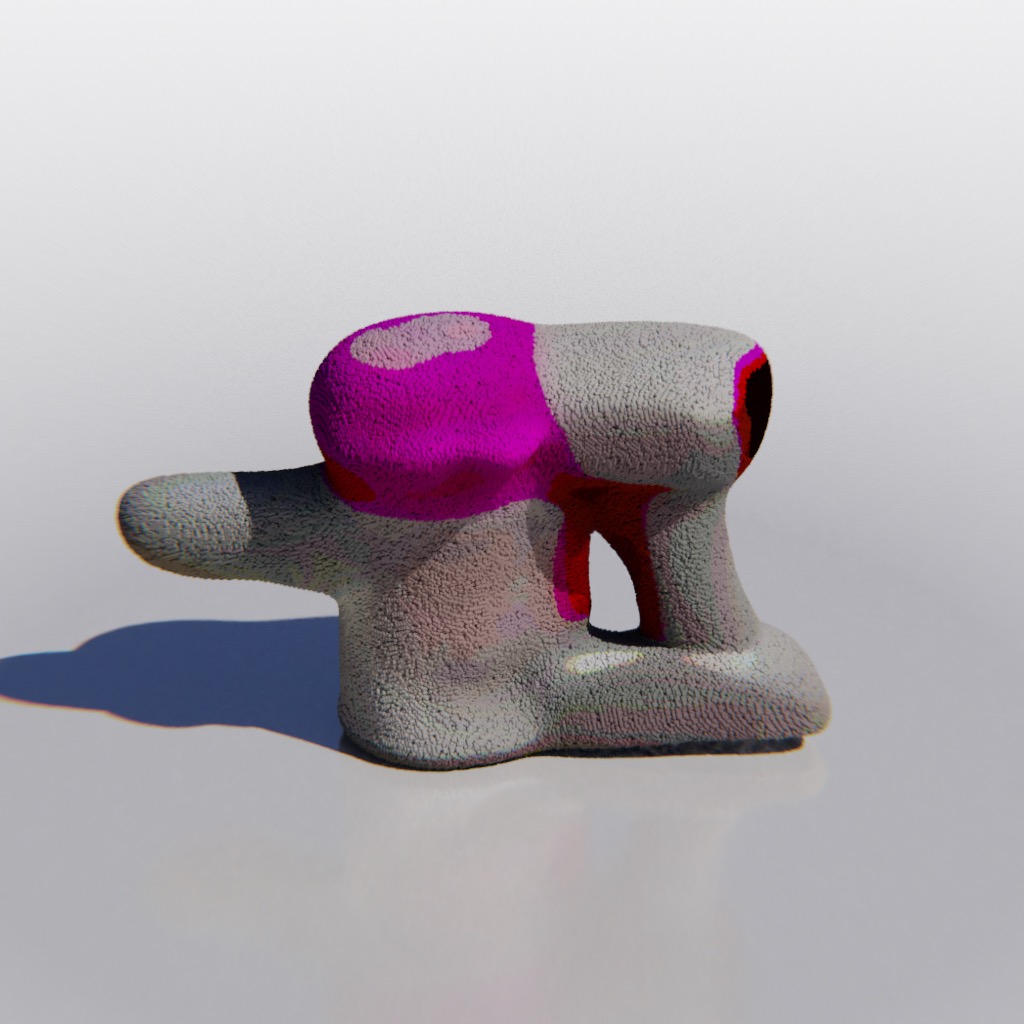}&
\myimg{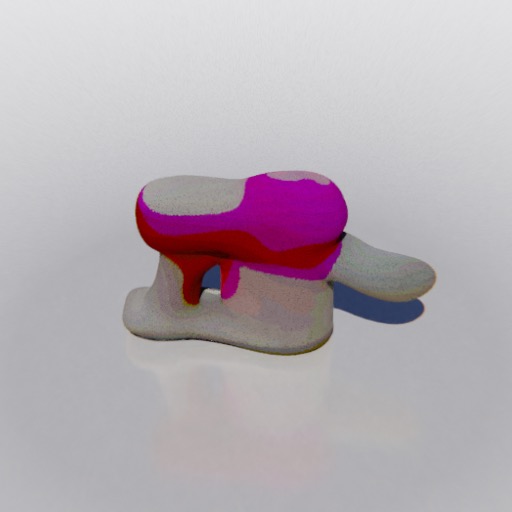}\\
\multirow{2}{*}{\inputimg{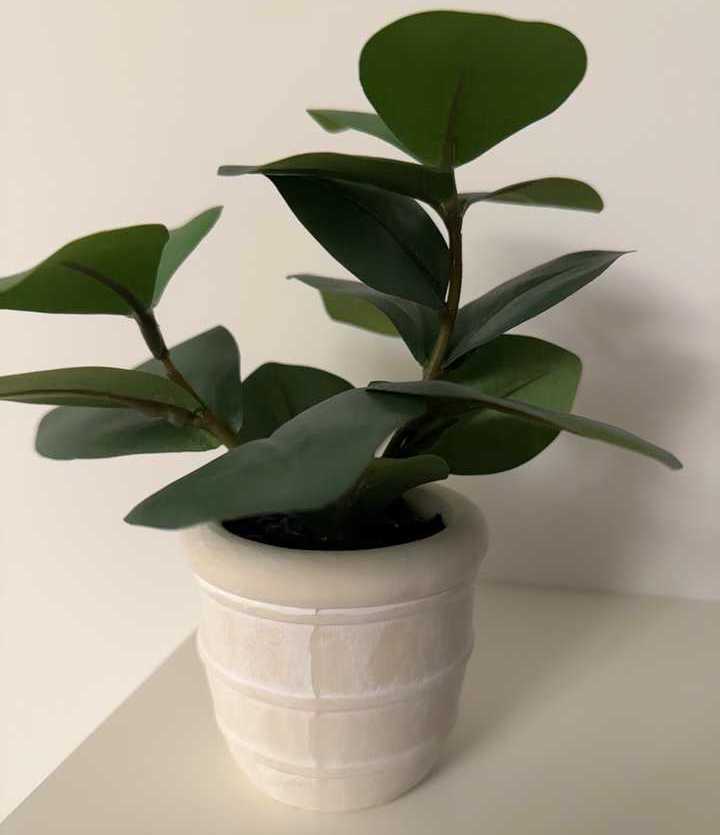}\hspace{-1.9mm}}&
\seenimg{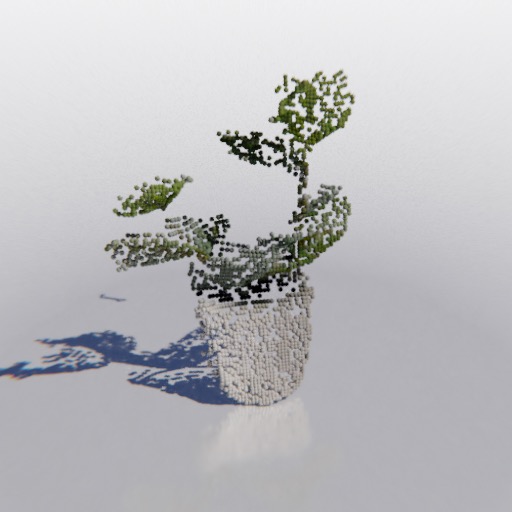}&
\myimg{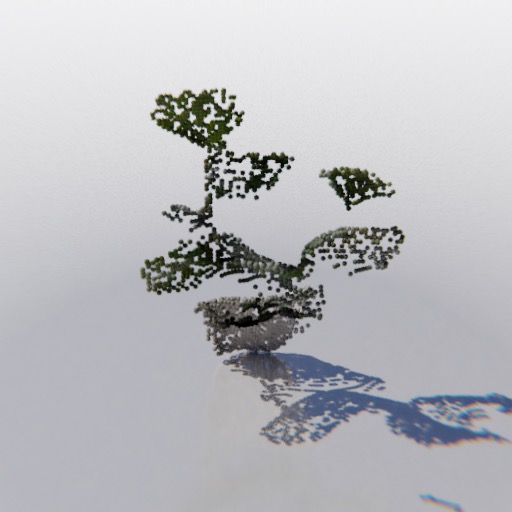}&
\myimg{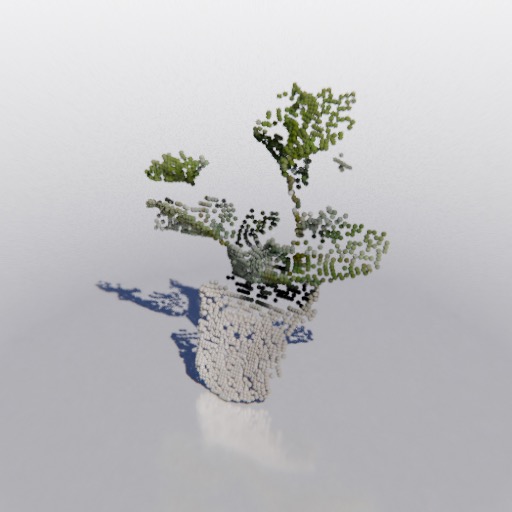}\\
&
\outimg{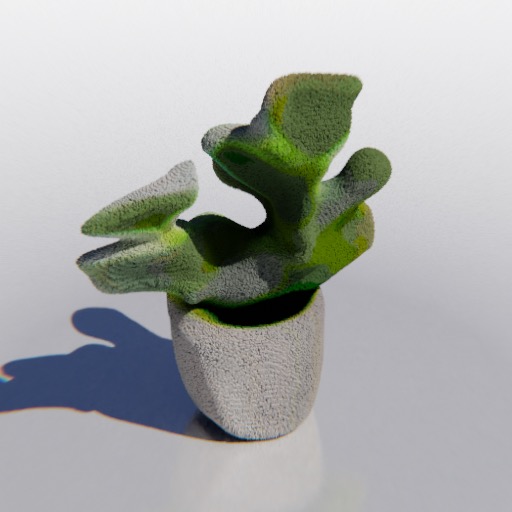}&
\myimg{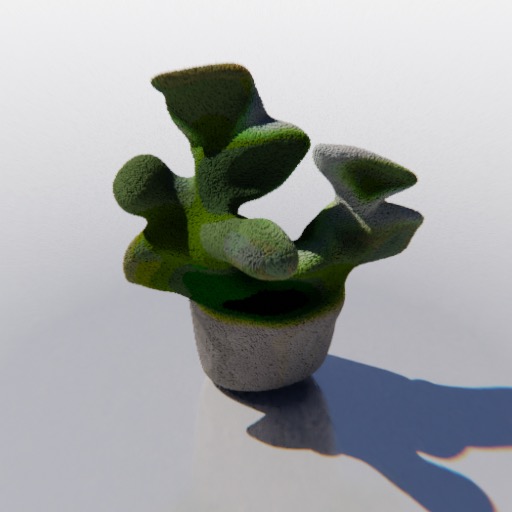}&
\myimg{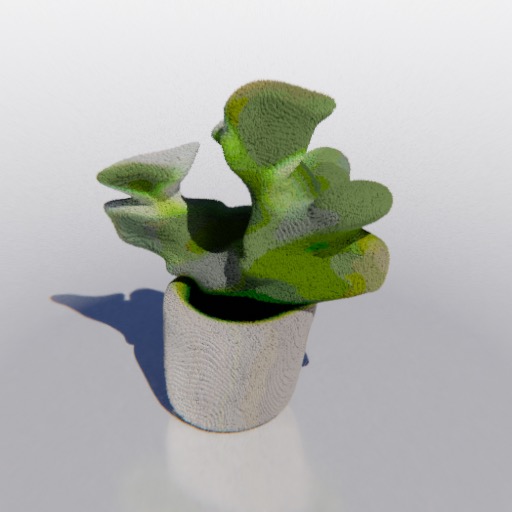}\\
\multirow{2}{*}{\inputimg{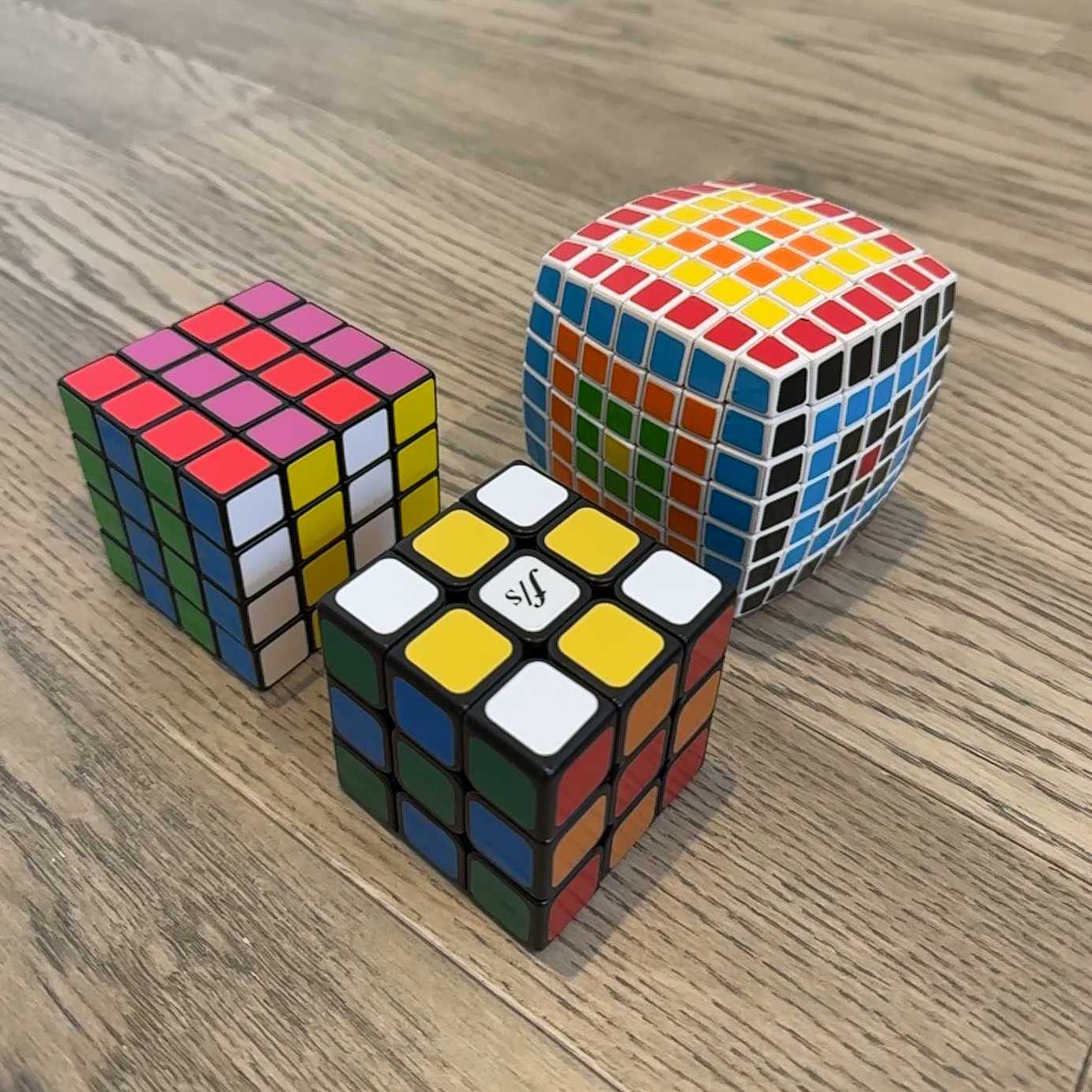}\hspace{-1.9mm}}&
\seenimg{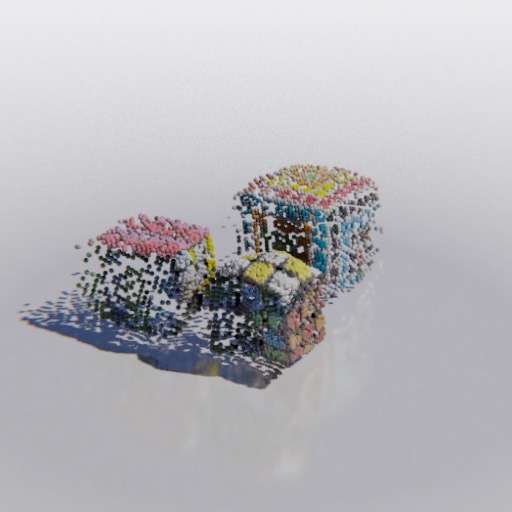}&
\myimg{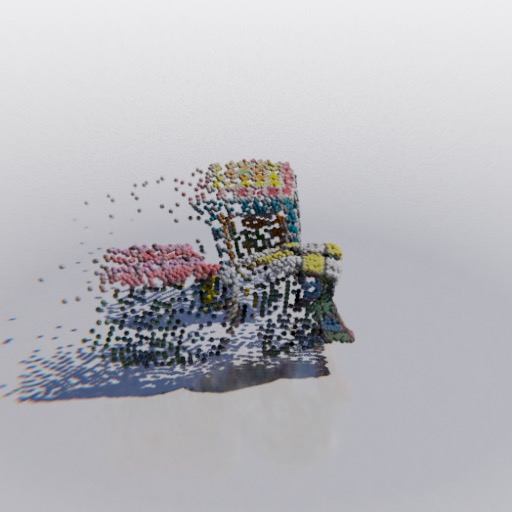}&
\myimg{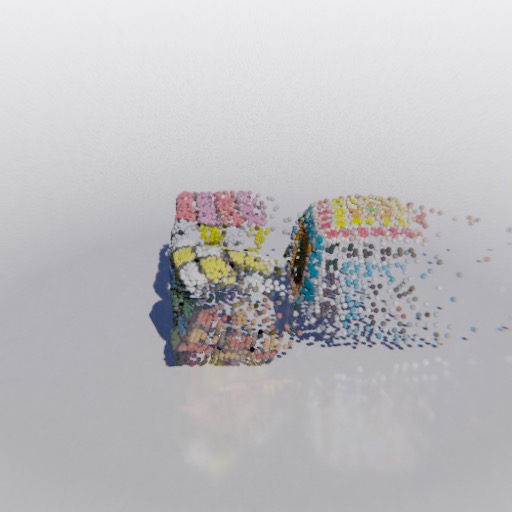}\\
&
\outimg{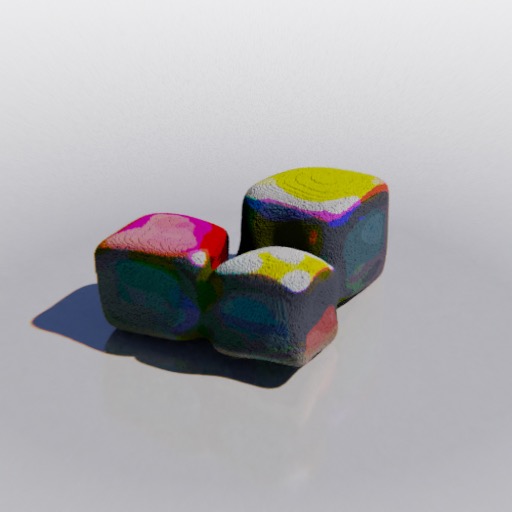}&
\myimg{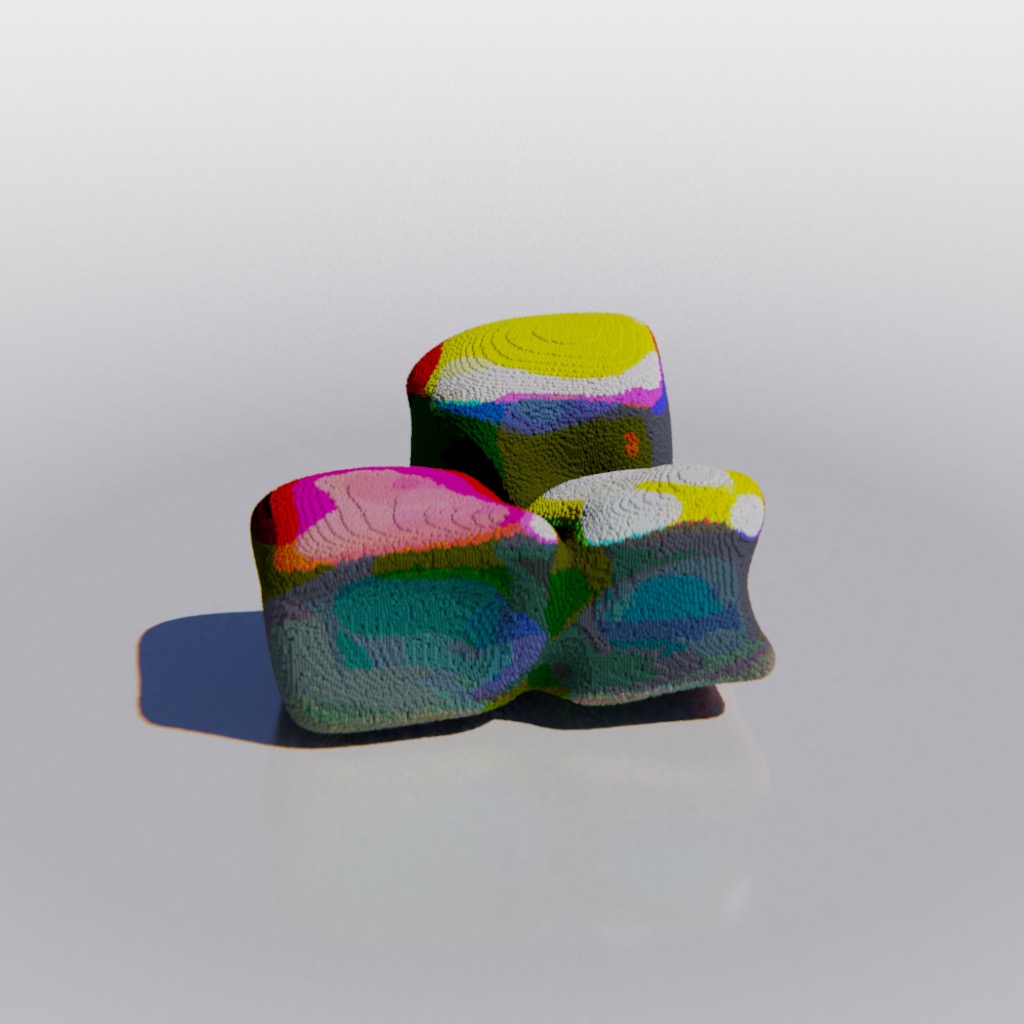}&
\myimg{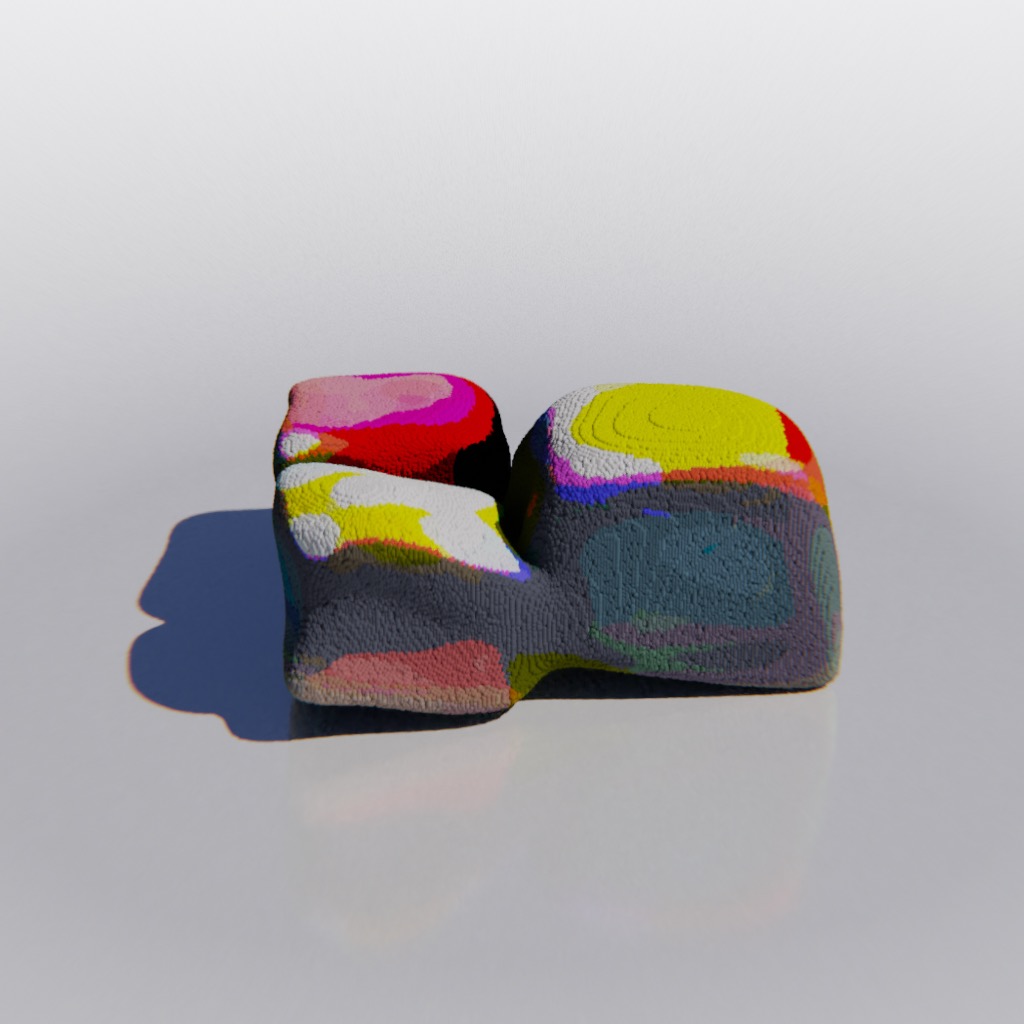}
\end{tabular}
}}\hspace{0mm}
\subfloat[ImageNet\label{fig:imagenet}]{%
\resizebox{0.333\linewidth}{!}{
\begin{tabular}{@{}x{45}x{45}x{45}x{45}@{}}
\multirow{2}{*}[-2mm]{\inputimg{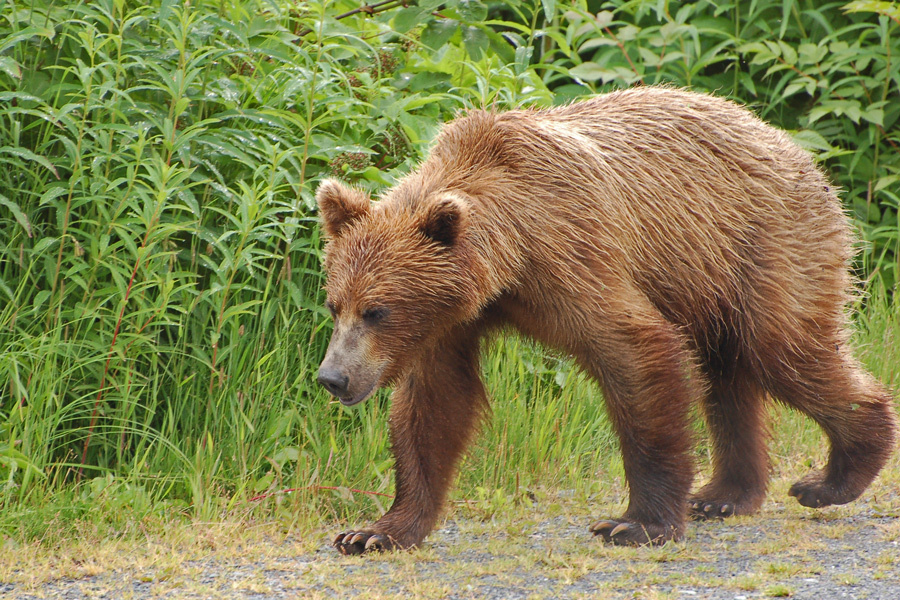}\hspace{-1.9mm}}&
\seenimg{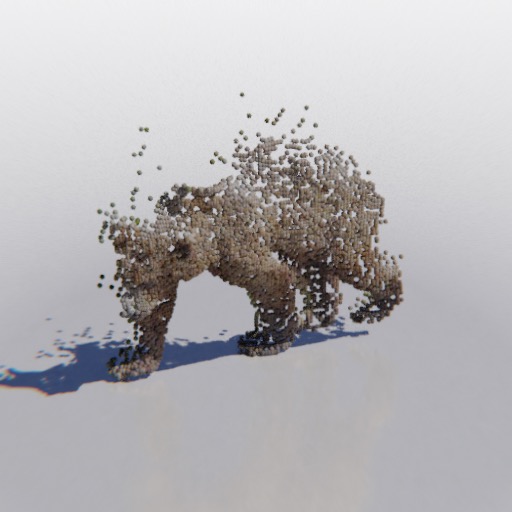}&
\myimg{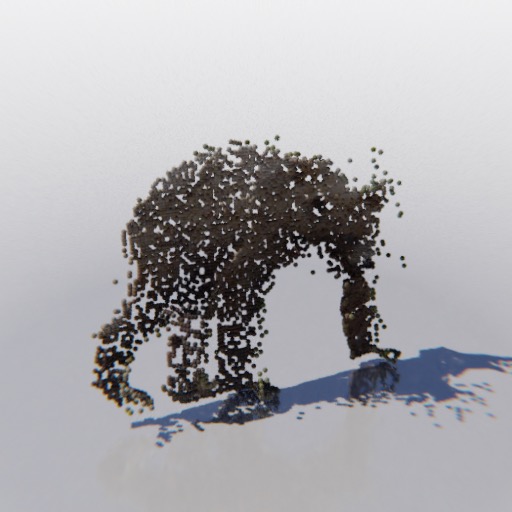}&
\myimg{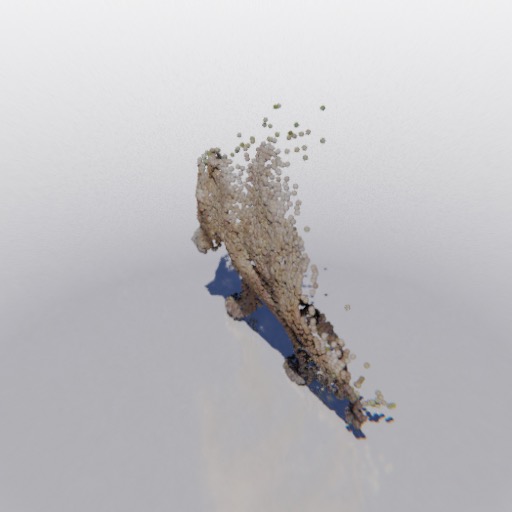}\\
&
\outimg{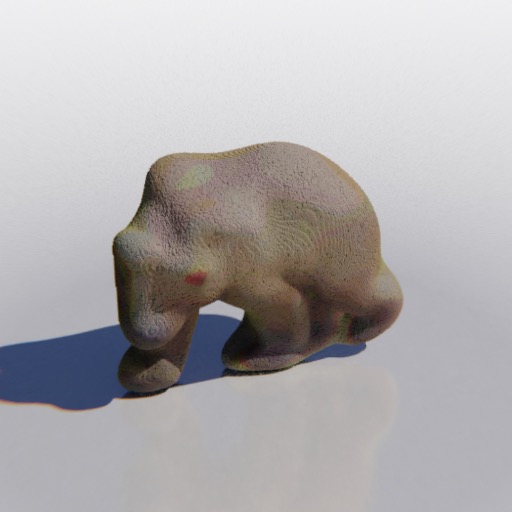}&
\myimg{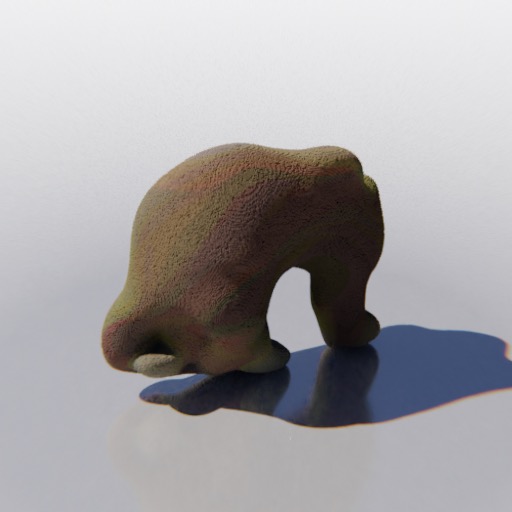}&
\myimg{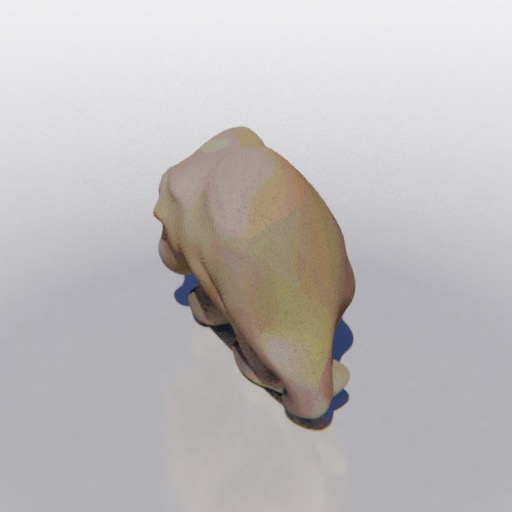}\\
\multirow{2}{*}[0mm]{\inputimg{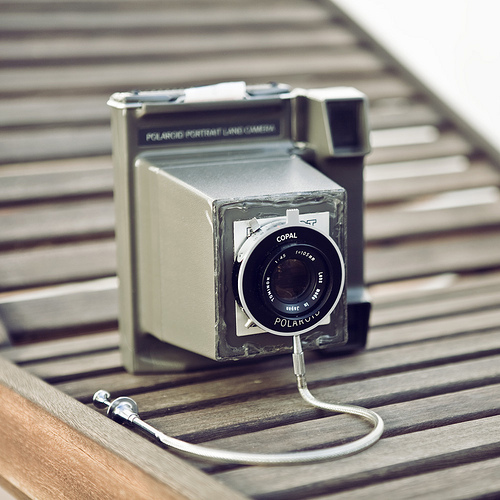}\hspace{-1.9mm}}&
\seenimg{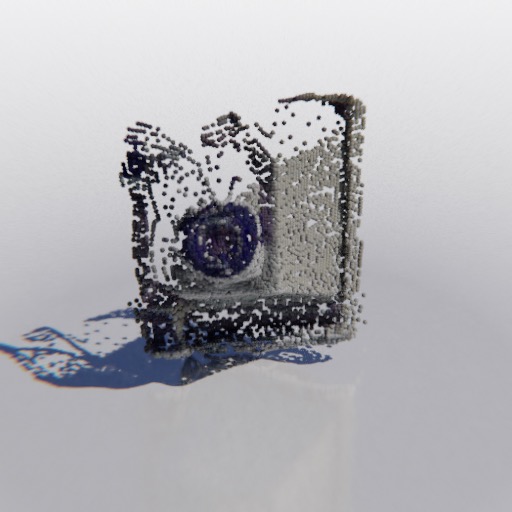}&
\myimg{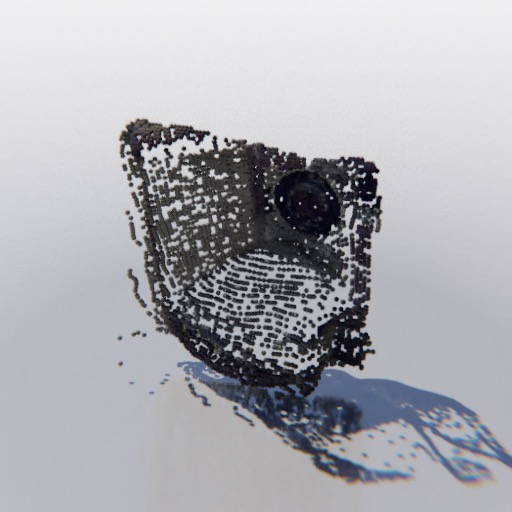}&
\myimg{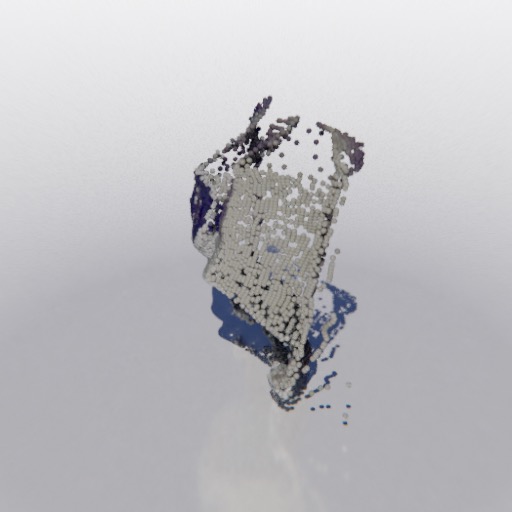}\\
&
\outimg{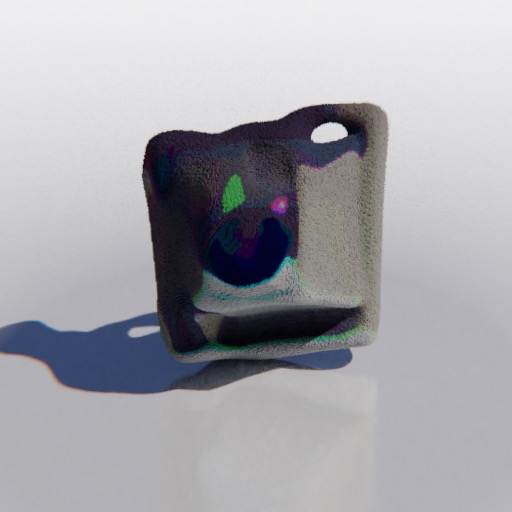}&
\myimg{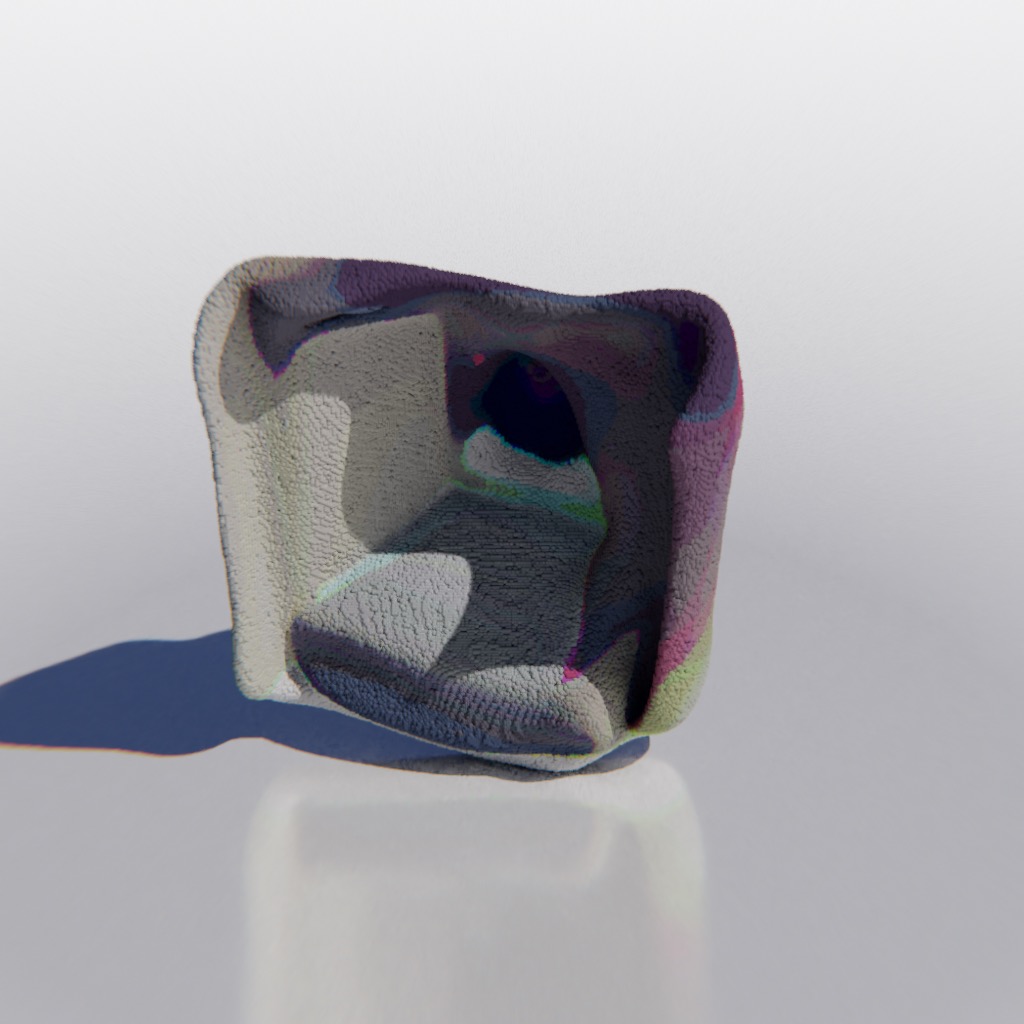}&
\myimg{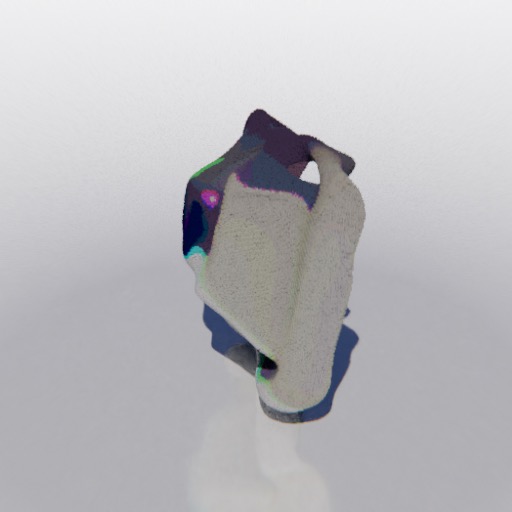}\\
\multirow{2}{*}[3mm]{\inputimg{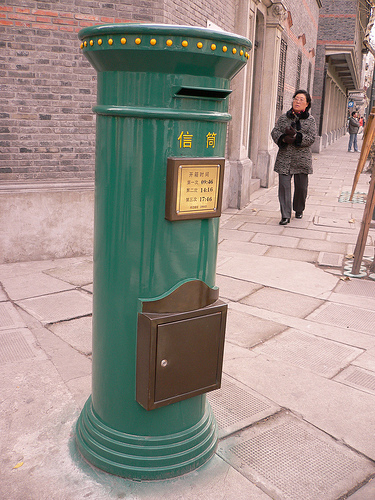}\hspace{-1.9mm}}&
\seenimg{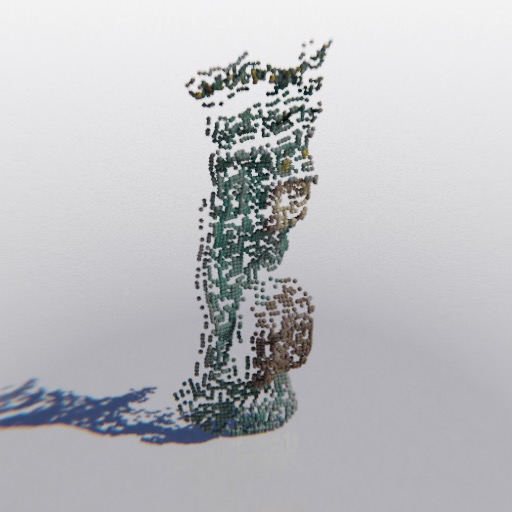}&
\myimg{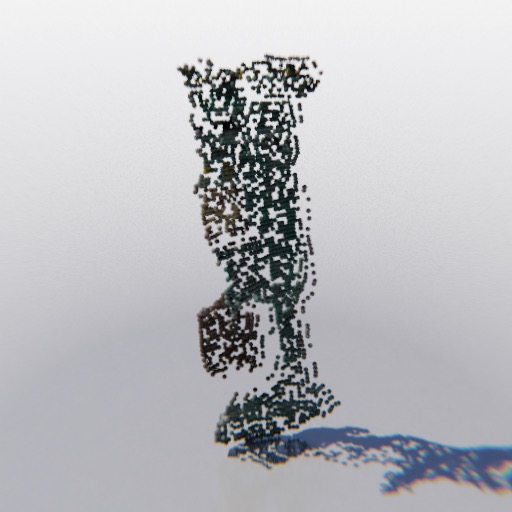}&
\myimg{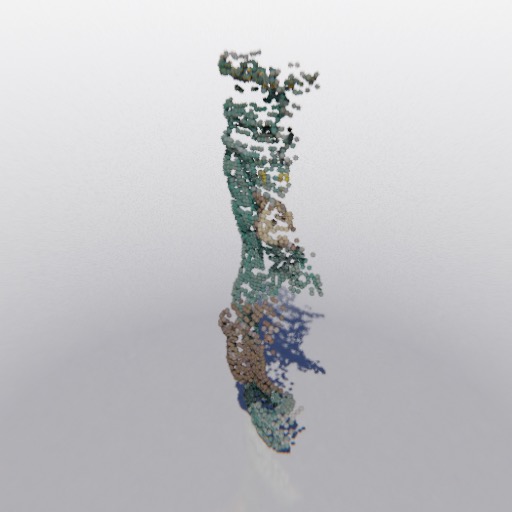}\\
&
\outimg{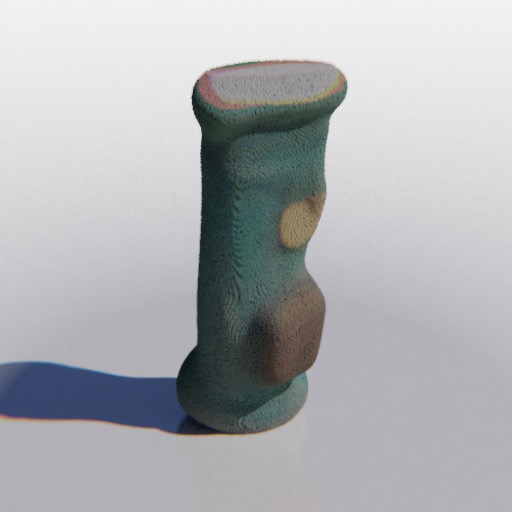}&
\myimg{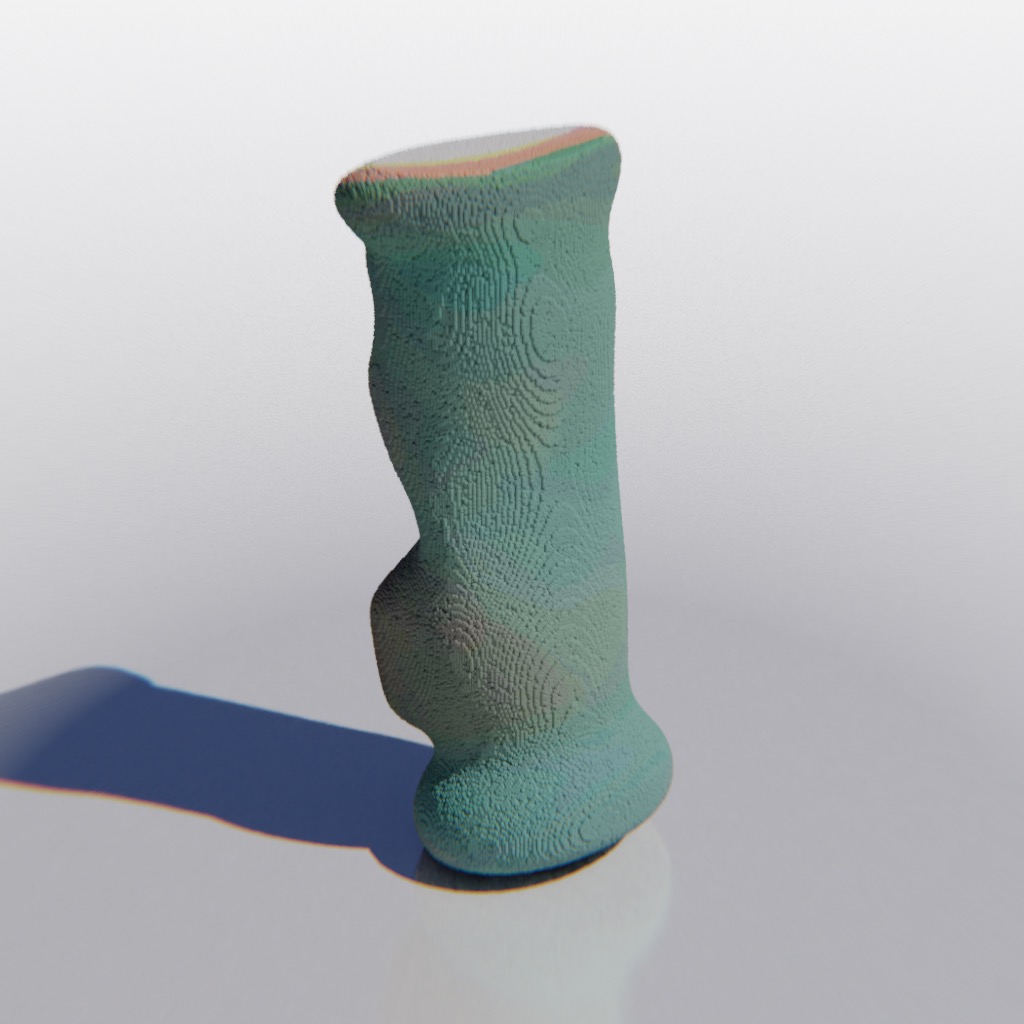}&
\myimg{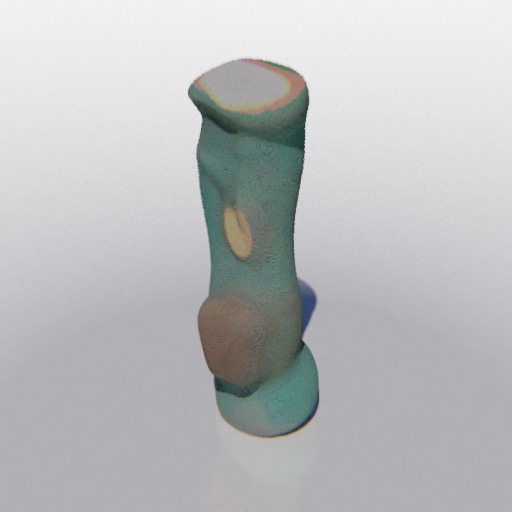}\\
\multirow{2}{*}[-2mm]{\inputimg{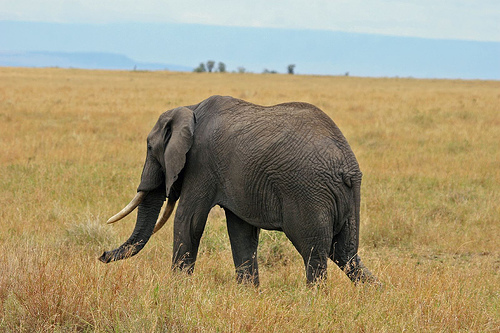}\hspace{-1.9mm}}&
\seenimg{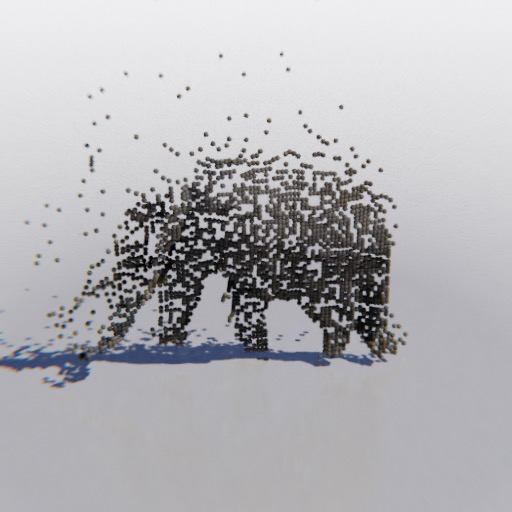}&
\myimg{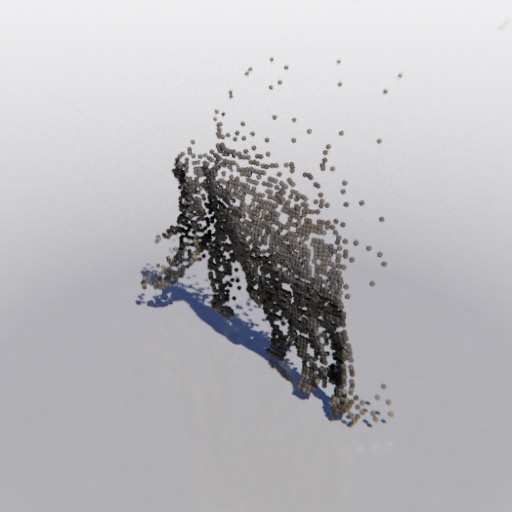}&
\myimg{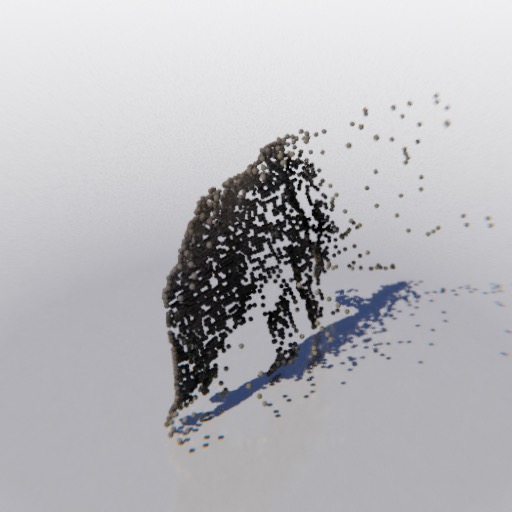}\\
&
\outimg{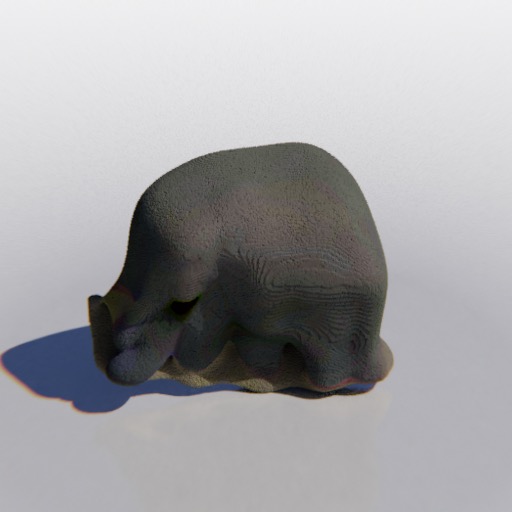}&
\myimg{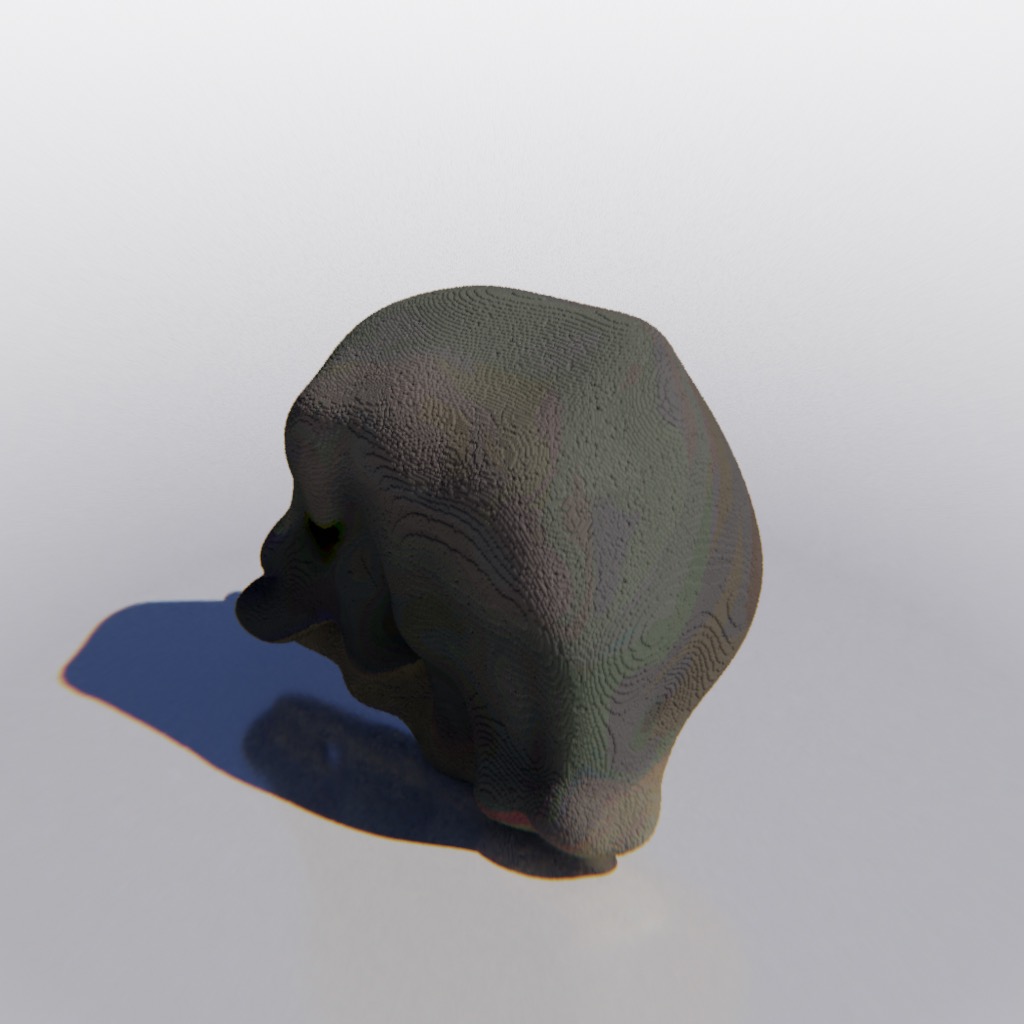}&
\myimg{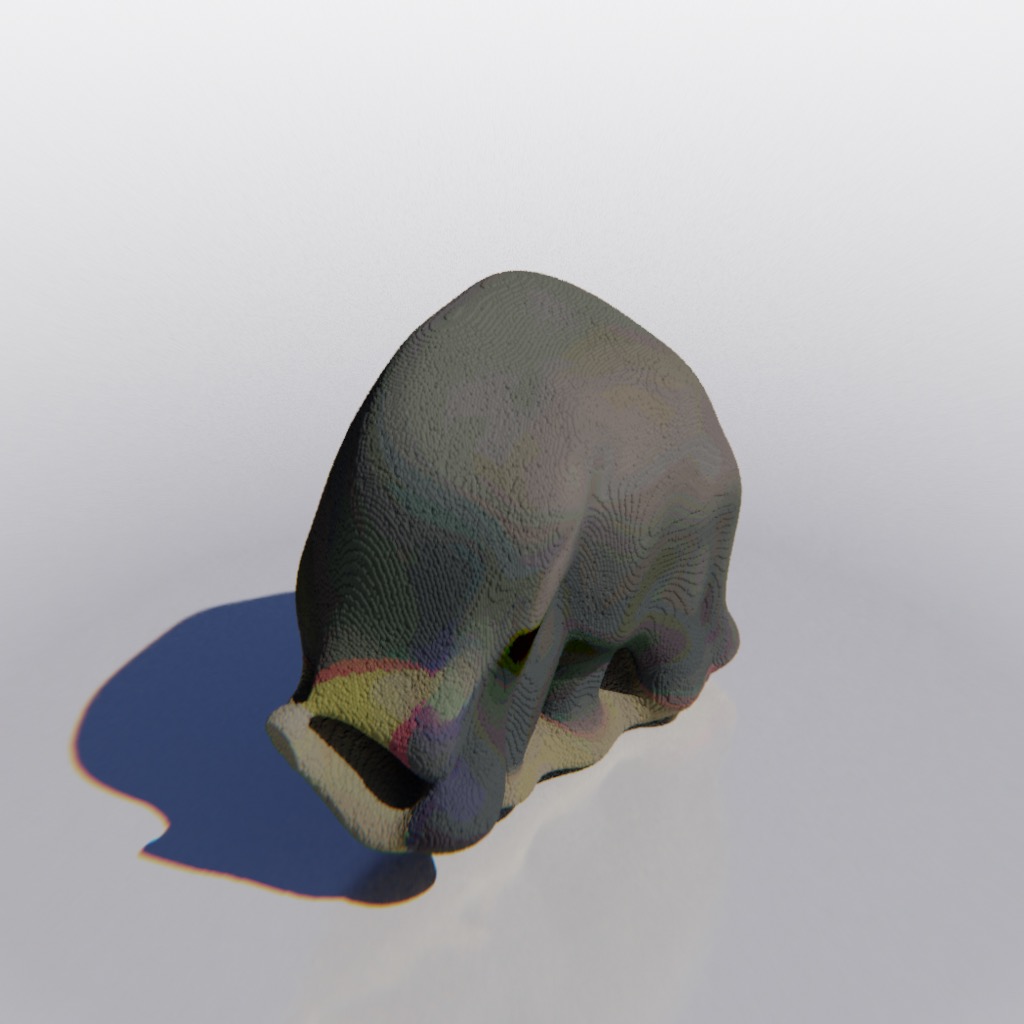}
\end{tabular}
}}\hspace{0mm}
\subfloat[DALL$\cdot$E 2\label{fig:dalle}]{%
\resizebox{0.333\linewidth}{!}{
\begin{tabular}{@{}x{45}x{45}x{45}x{45}@{}}
\multirow{2}{*}[-2mm]{\inputimg{figs/dalle2/An_armchair_in_the_shape_of_an_avocado.jpg}\hspace{-1.9mm}}&
\seenimg{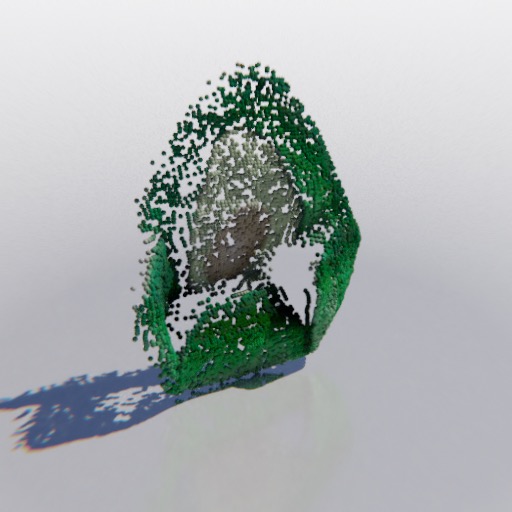}&
\myimg{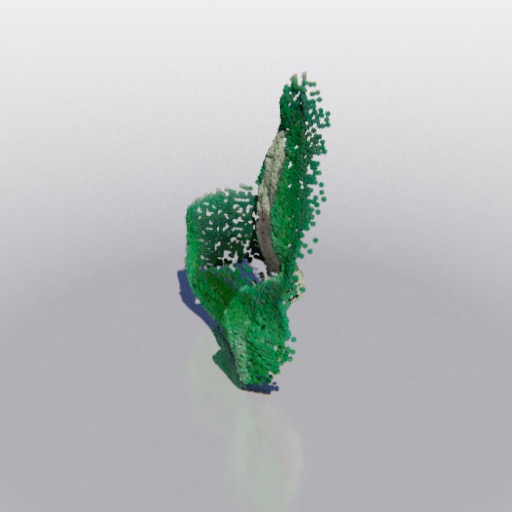}&
\myimg{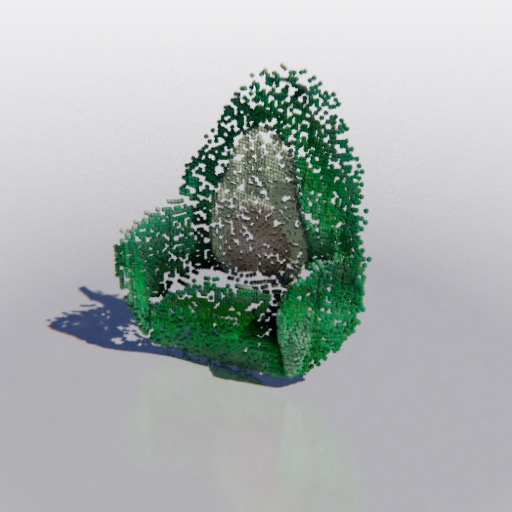}\\
&
\outimg{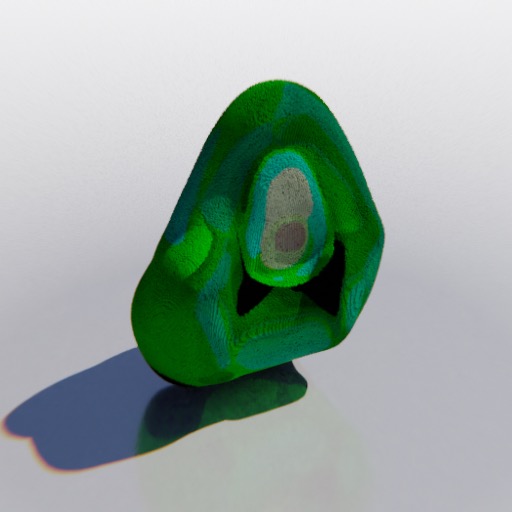}&
\myimg{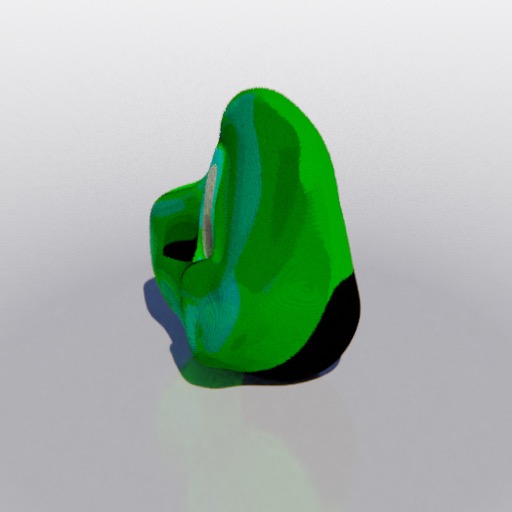}&
\myimg{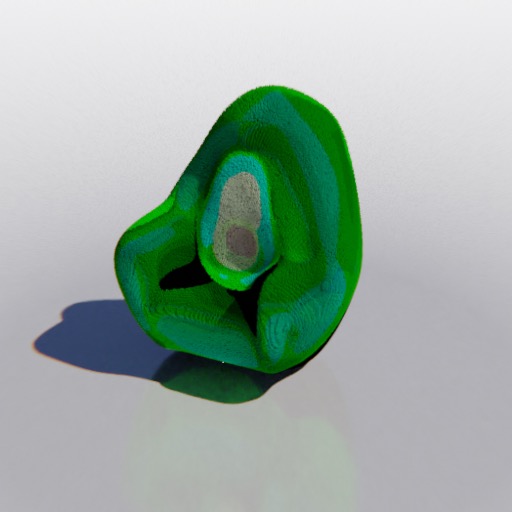}\\
\multirow{2}{*}[-2mm]{\inputimg{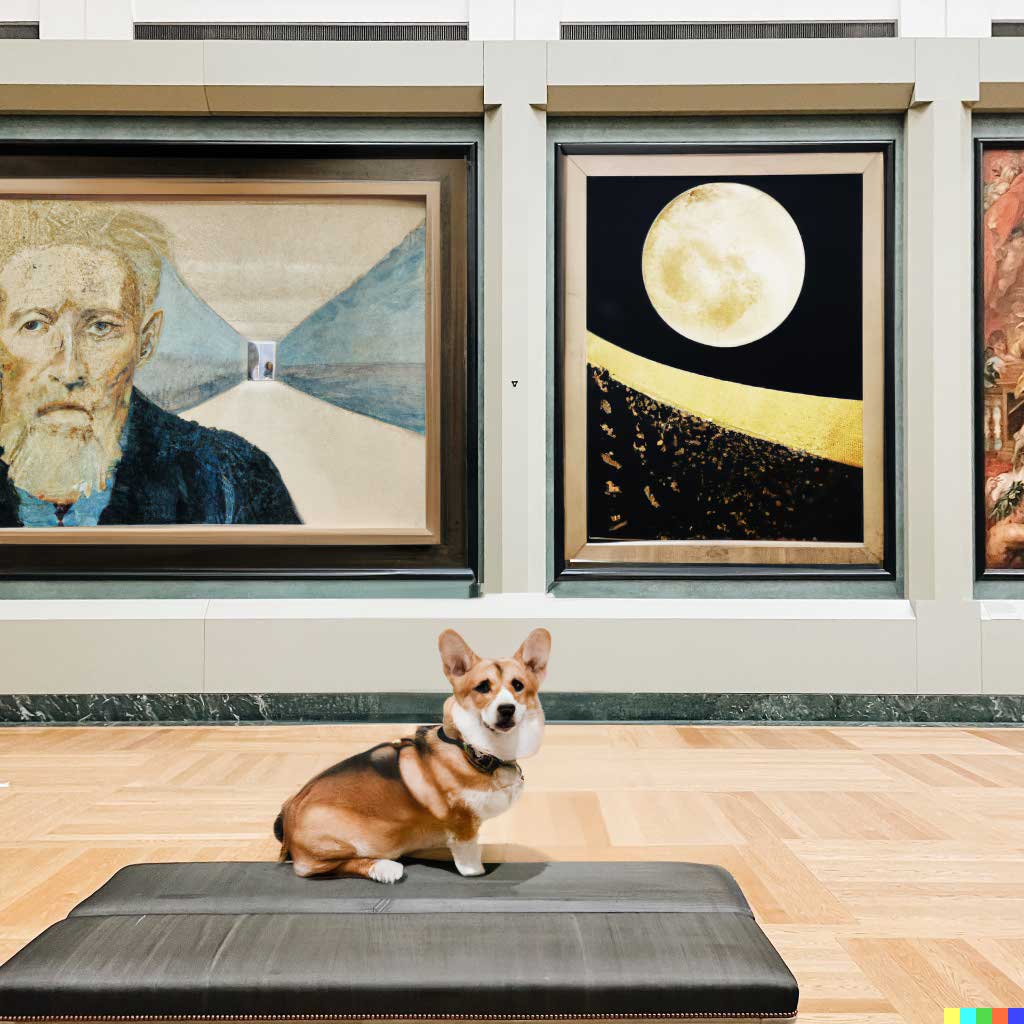}\hspace{-1.9mm}}&
\seenimg{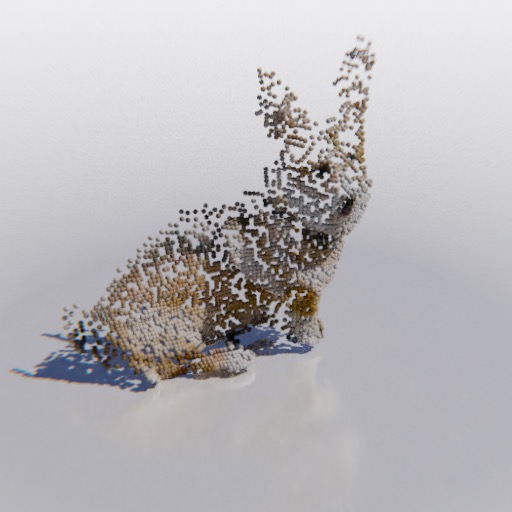}&
\myimg{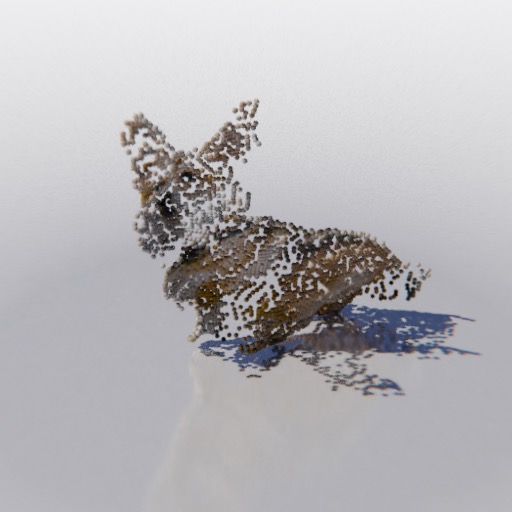}&
\myimg{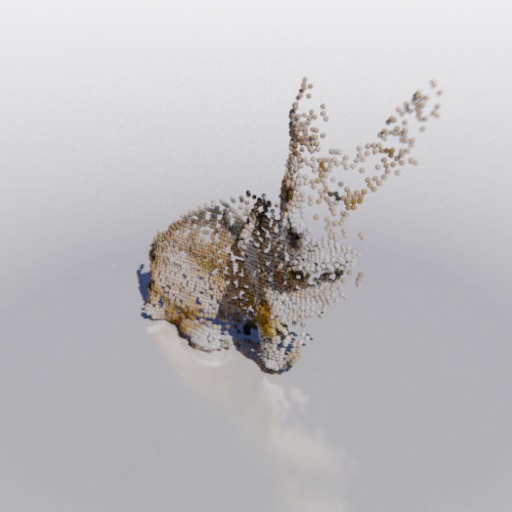}\\
&
\outimg{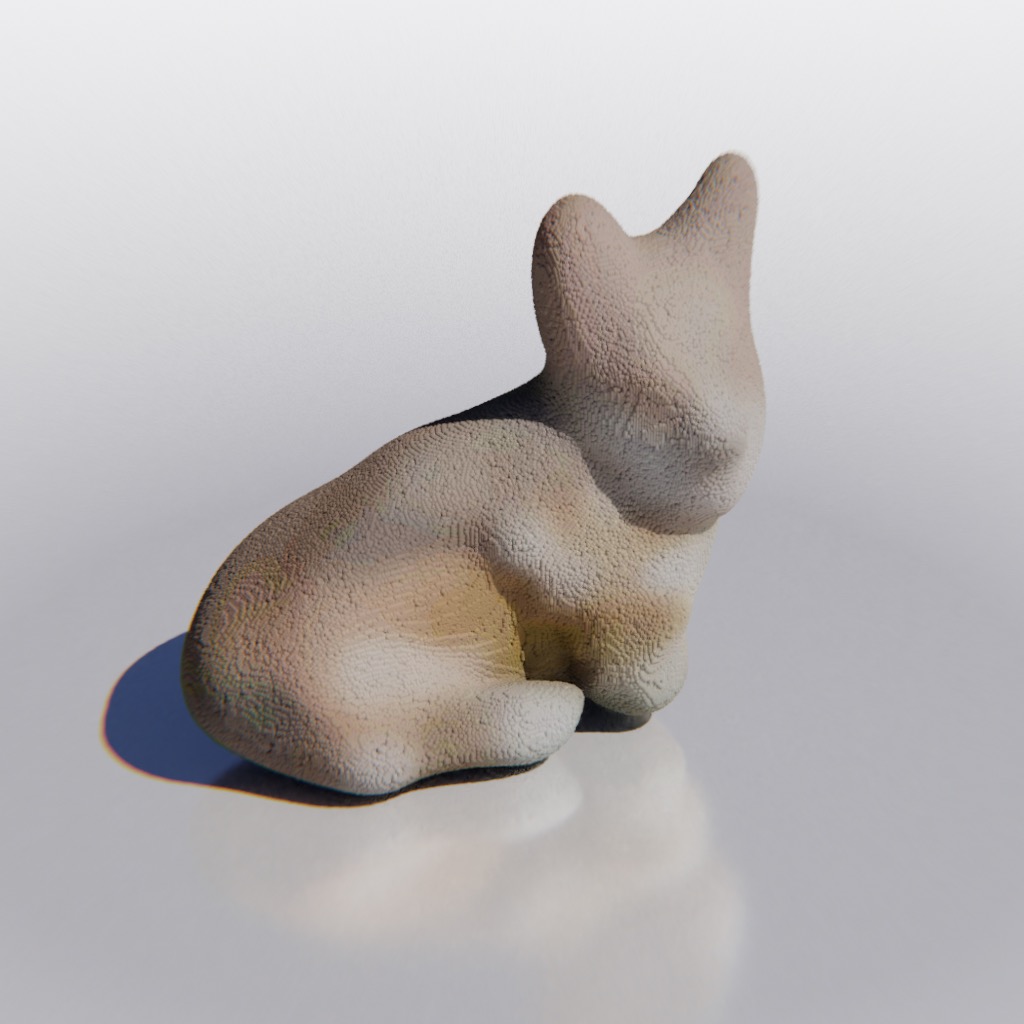}&
\myimg{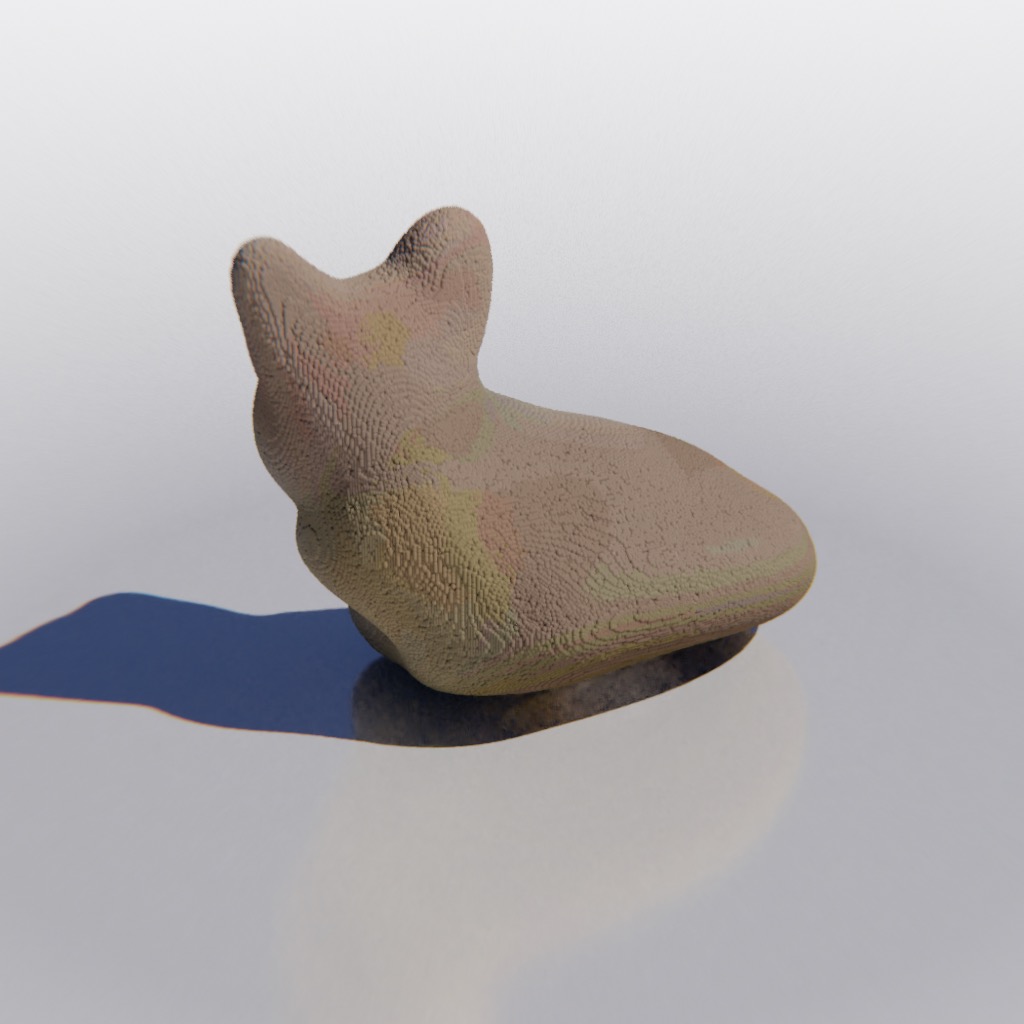}&
\myimg{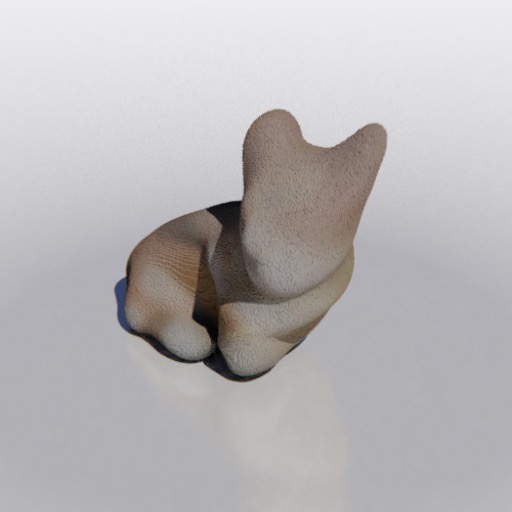}\\
\multirow{2}{*}[-2mm]{\inputimg{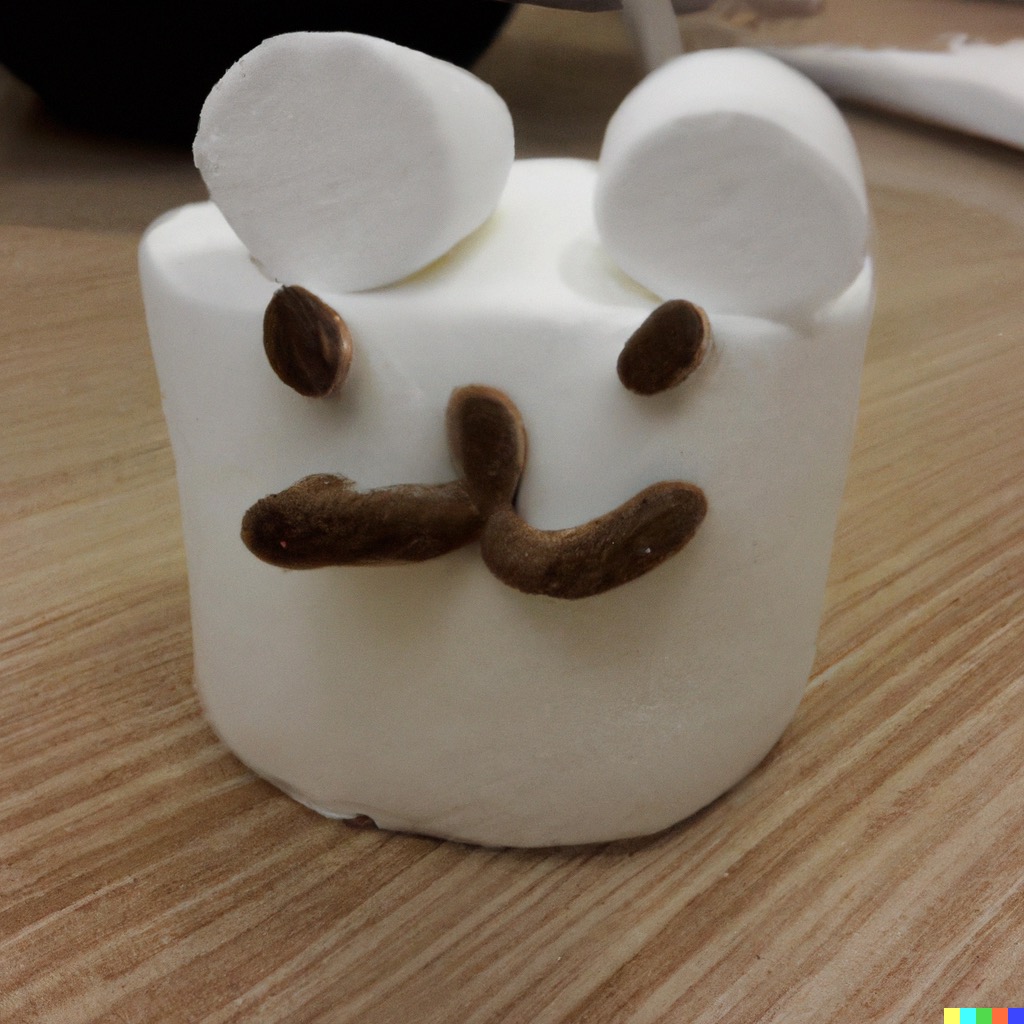}\hspace{-1.9mm}}&
\seenimg{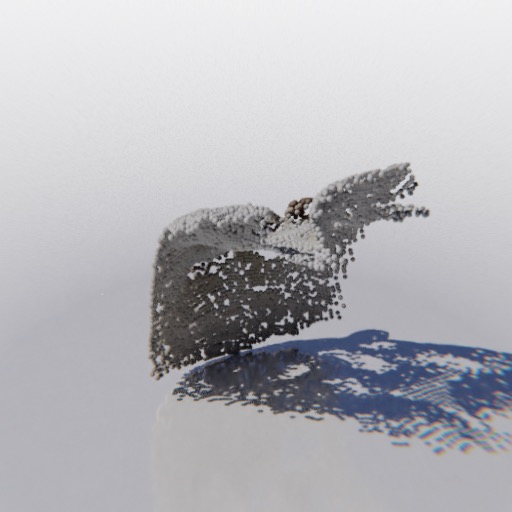}&
\myimg{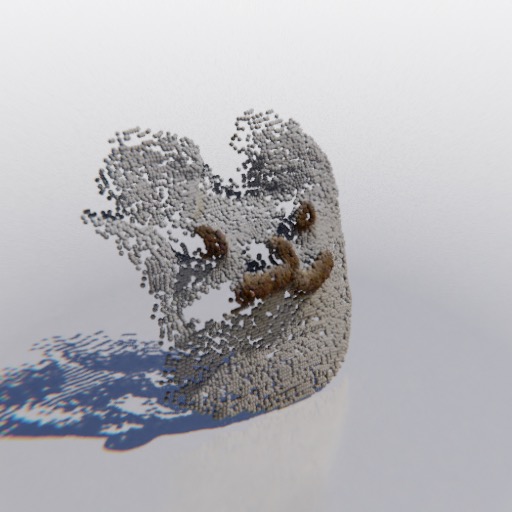}&
\myimg{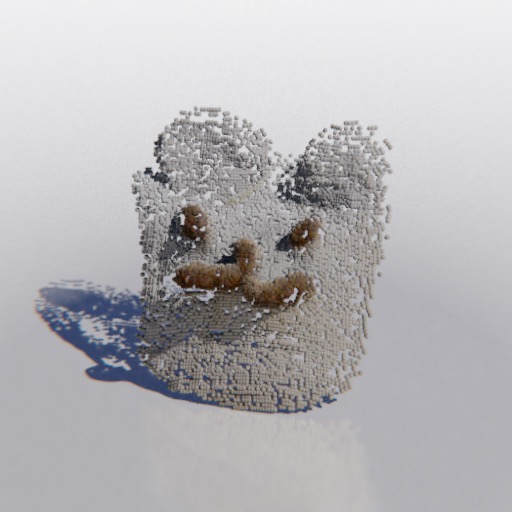}\\
&
\outimg{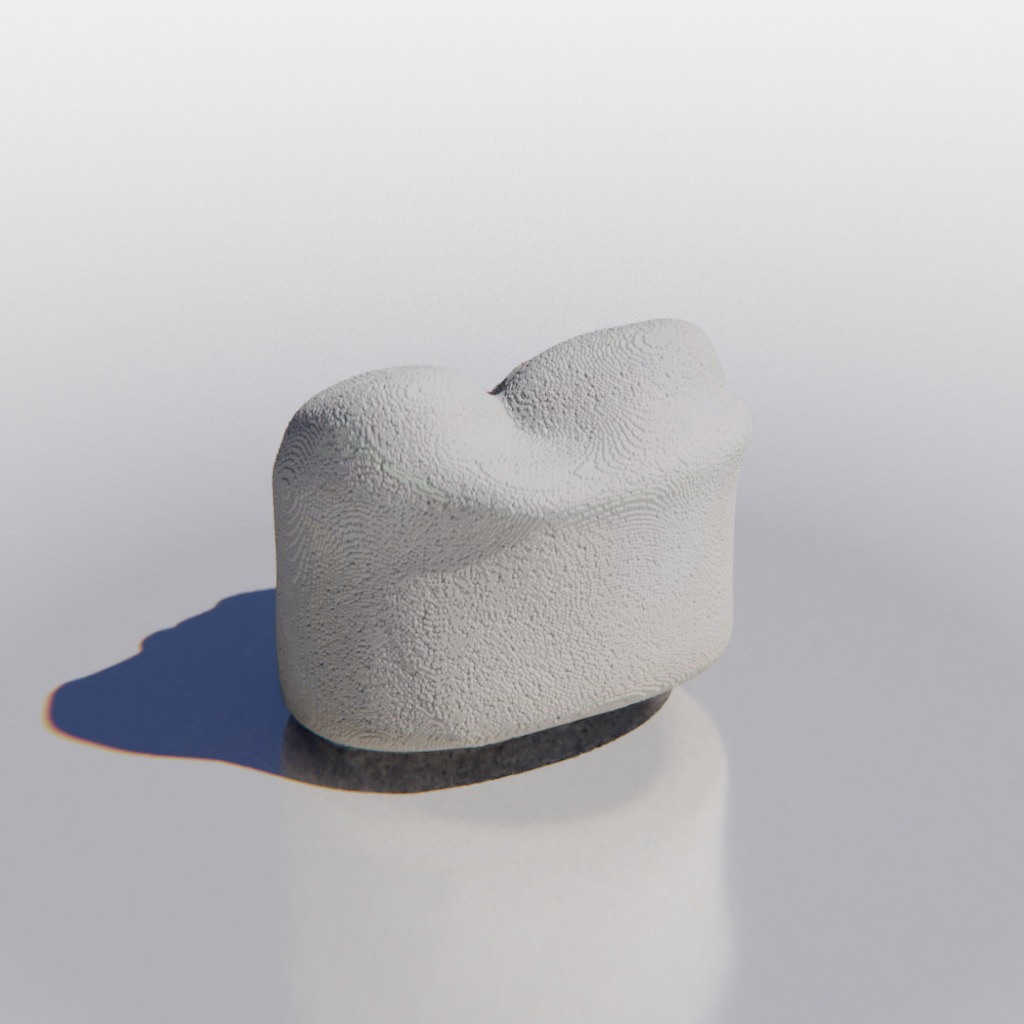}&
\myimg{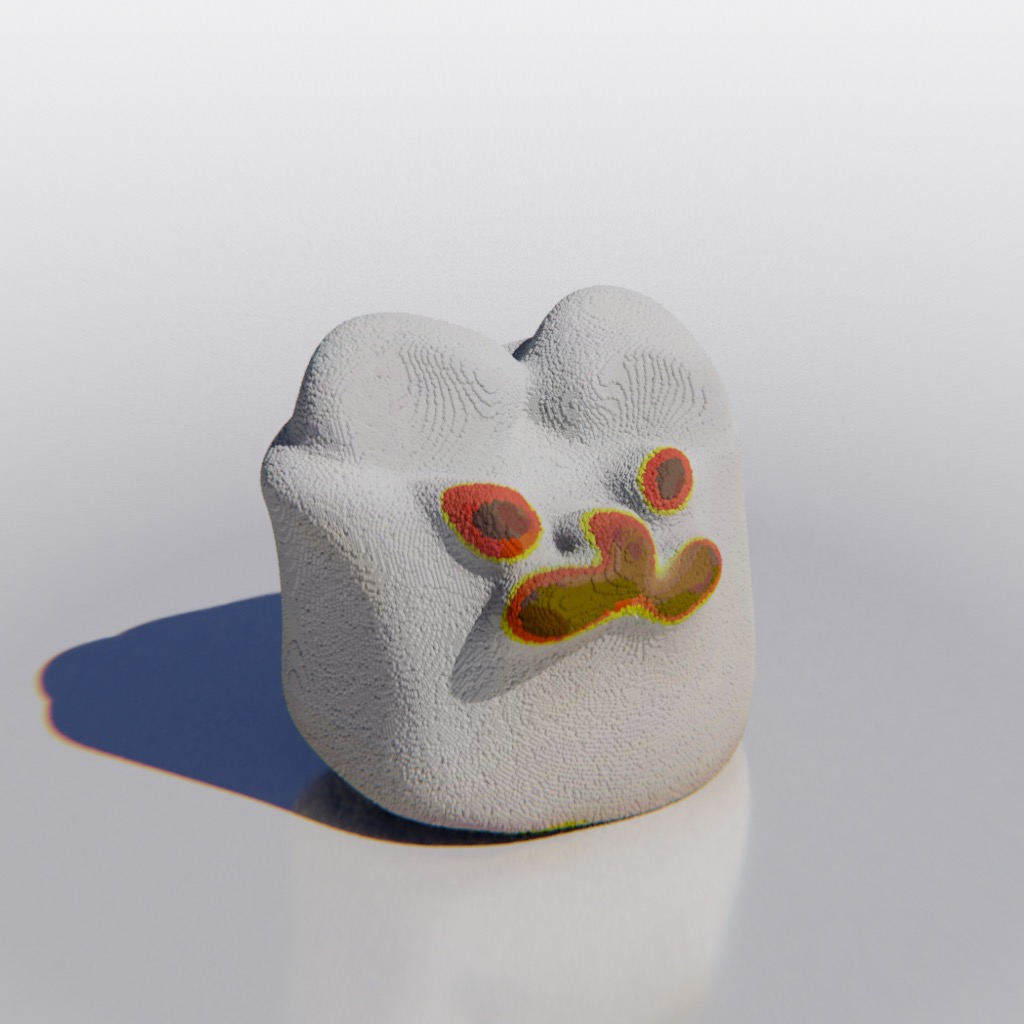}&
\myimg{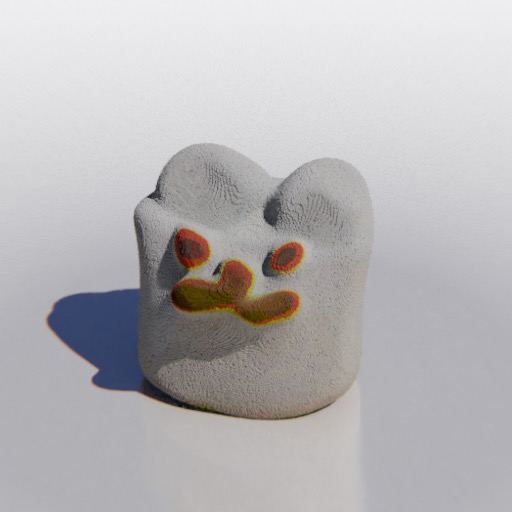}\\
\multirow{2}{*}[-2mm]{\inputimg{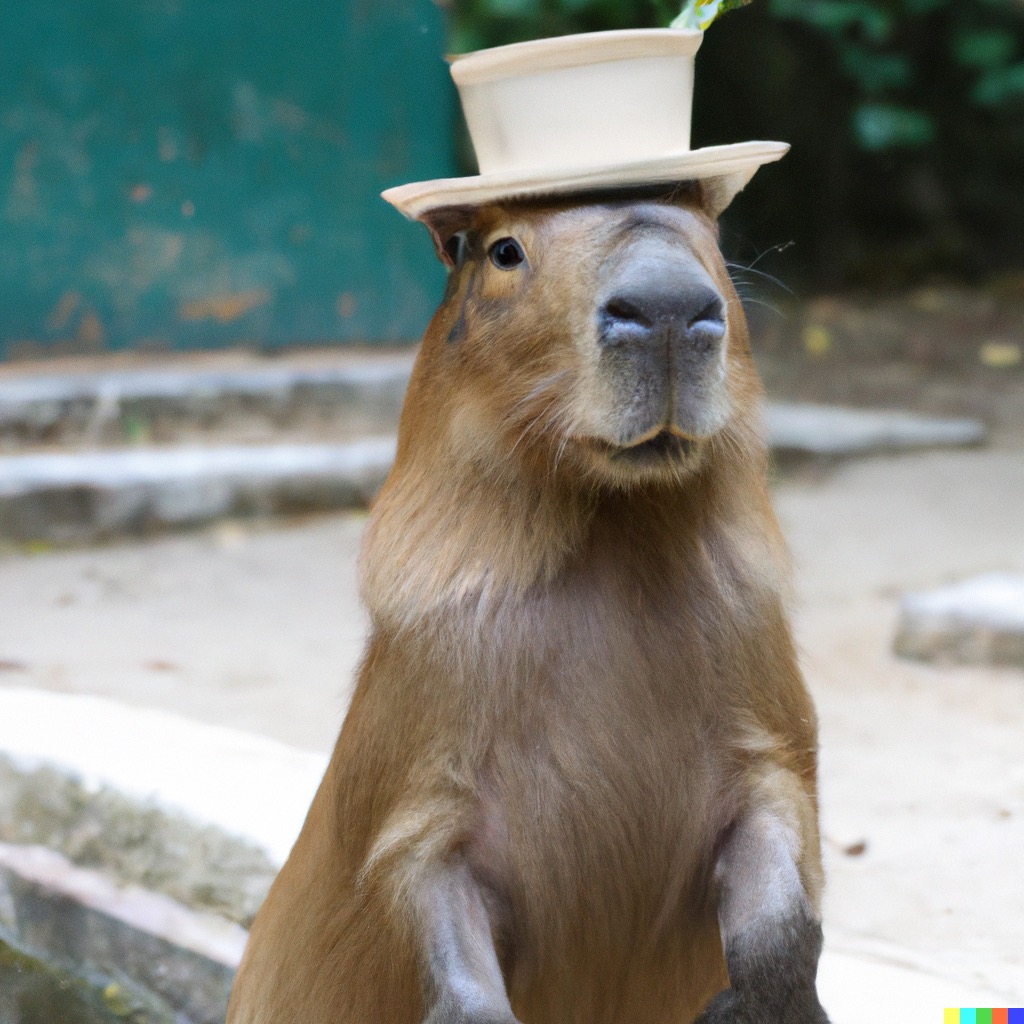}\hspace{-1.9mm}}&
\seenimg{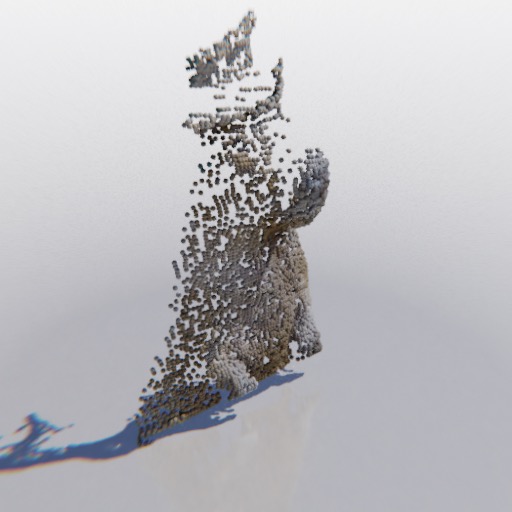}&
\myimg{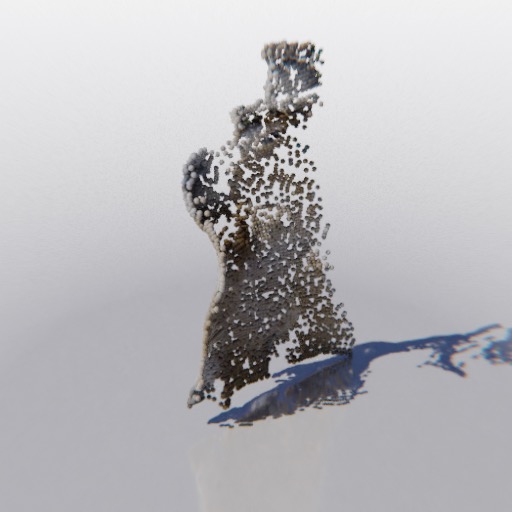}&
\myimg{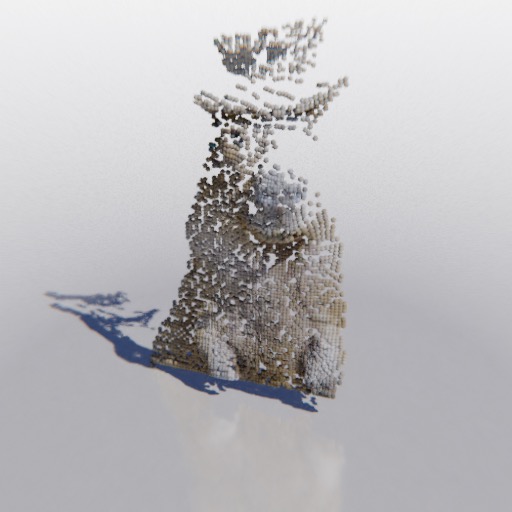}\\
&
\outimg{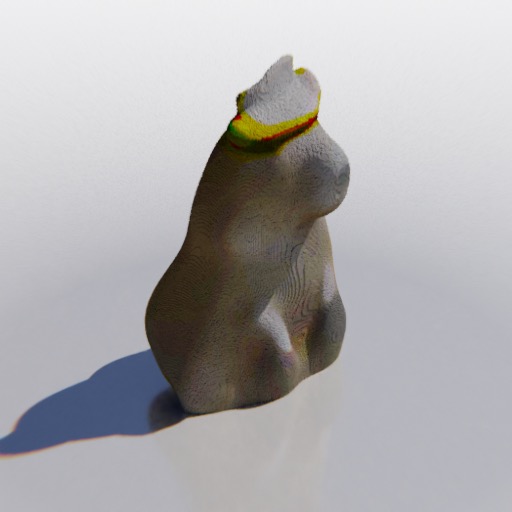}&
\myimg{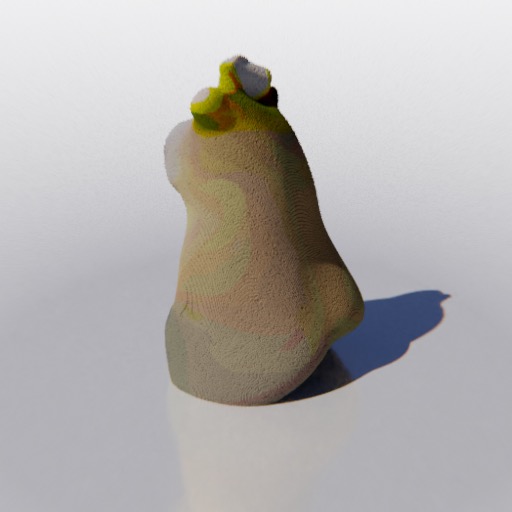}&
\myimg{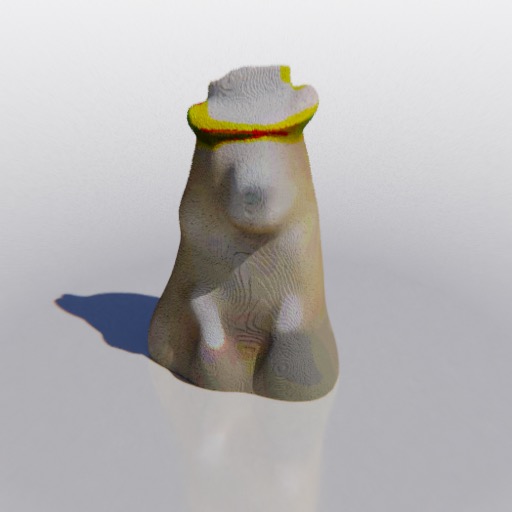}\\
\end{tabular}
}}
\vspace{-3mm}
\caption{\textbf{Zero-Shot Generalization.}
We test \method, trained on CO3D-v2~\cite{reizenstein21co3d}, on three challenging settings:
(a) iPhone captures with LiDAR sensor of everyday objects,
(b) Web images (from ImageNet) of in-the-wild objects with depth estimated by an off-the-shelf model~\cite{ranftl2021vision},
(c) AI-generated images (by DALL$\cdot$E 2) of imaginary objects with depth estimated by~\cite{ranftl2021vision}.
These examples are challenging as they demonstrate variance in object types (\eg, novel, imaginary objects), image styles (\eg, digital arts, natural), depth systems (\eg, depth sensor, off-the-shelf predictors), and visual context (\eg, safari, street scene). See \href{https://mcc3d.github.io/}{project page} for animations. }
\label{fig:zeroshot}
\vspace{-5mm}
\end{figure*}

\subsection{Scaling Behavior Analysis}
\label{sec:exp:scale}
\method's strength is that it only requires \emph{points} for training and does not rely on any shape priors.
As a result, \method can train on a large number of examples.
We analyze our model's performance as a function of data size.
\figref{scaling} shows that scaling the training data leads to steady performance improvements.
Furthermore, if we increase the number of categories, and thus the visual diversity of our training data, the improvements are even larger.
This suggests two things.
First, building category-agnostic scaleable models like \method is a promising direction towards general-purpose 3D reconstruction.
Second, expanding the datasets, and especially the set of categories, is promising.

\subsection{Zero-Shot Generalization In-the-Wild}
In \secref{co3dv2qual}, we show generalization to novel categories from the CO3D dataset.
Now, we turn to in-the-wild settings and show \method reconstructions on ImageNet~\cite{deng2009imagenet}, iPhone captures, and AI-generated images~\cite{ramesh2022hierarchical}.

\mypar{iPhone Captures.}
This is arguably the most popular in-the-wild setting --- our personal use of an off-the-shelf smart phone for capturing everyday objects.
Specifically, we use iPhones and their depth sensor to take RGB-D images on a diverse set of objects in two of the coauthors' homes (using a 12 and 14 Pro iPhone).
This is a challenging setting due to the domain shift from the training data and the difference in the depth estimation pipeline (COLMAP in CO3D \vs sensor from iPhone).
\figref{iphone} shows ours results.
Examples such as the vacuum or the VR headset in \figref{teaser:b} stand out as they deviate from our training set.
\figref{iphone} demonstrates \method's ability to learn general shape priors, instead of memorizing the training set.

\mypar{ImageNet.}
We turn to ImageNet~\cite{deng2009imagenet}, which contains highly diverse Internet photos, ranging from bears and elephants in their natural habitat to Japanese mailboxes, drastically different than the staged CO3D objects.
For depth, we use an off-the-shelf model from Ranftl~\etal~\cite{ranftl2021vision}, which differs from CO3D's COLMAP output.
\figref{imagenet} shows results on ImageNet images of diverse objects.

\mypar{AI-generated Images.}
We test \method on DALL$\cdot$E 2 which generates images of imaginary objects.
\figref{dalle} shows \method reconstructions including the Internet-famous avocado chair and a cat-shaped marshmallow with a mustache!

\begin{table}[t]
\tablestyle{1.0pt}{1.05}
\resizebox{1\linewidth}{!}{
\begin{tabular}{@{\extracolsep{4pt}}lx{44}x{27}x{20}x{20}x{20}x{20}@{}}
&&&\multicolumn{2}{c}{seen categ.} & \multicolumn{2}{c}{unseen categ.}\\
\cline{4-5}\cline{6-7}
& depth sup.~\cite{deng2022depth}&depth in&Abs & MSE & Abs & MSE\\
\shline
NeRF-WCE~\cite{henzler2021unsupervised}&&&8.43&175.5&10.1&156.4\\
NeRF-WCE~\cite{henzler2021unsupervised}&\checkmark&& 7.38 &92.2& 9.15&139.9\\
NeRF-WCE~\cite{henzler2021unsupervised}&&\checkmark& 7.46 &156.3& 8.30 & 119.4\\
NeRF-WCE~\cite{henzler2021unsupervised}&\checkmark&\checkmark& 2.75 &78.4 & 2.79&30.5\\
NerFormer~\cite{reizenstein21co3d}&&& 2.02&70.4&2.00&20.6\\
NerFormer~\cite{reizenstein21co3d}&\checkmark&& 2.19&72.8&2.18&23.5\\
NerFormer~\cite{reizenstein21co3d}&&\checkmark& 2.20&72.1&2.17&22.5\\
NerFormer~\cite{reizenstein21co3d}&\checkmark&\checkmark& 2.34&80.7&2.28&24.1\\
\textbf{MCC} &\checkmark&\checkmark& \textbf{1.46} & \textbf{38.8} & \textbf{1.17} & \textbf{13.6}
\end{tabular}}
\vspace{-3mm}
\caption{\textbf{Comparison to the State-of-the-Art on CO3D-v2~\cite{reizenstein21co3d}.}
For a fair comparison with \method, we extend baselines~\cite{reizenstein21co3d,henzler2021unsupervised} with depth supervision~\cite{deng2022depth} or using depth as input.
MCC outperforms prior state of the art on CO3D-v2 for shape reconstruction.}
\label{tab:co3dnerf}
\vspace{-5mm}
\end{table}

\begin{figure}[t]
\small
\tablestyle{1.0pt}{1.05}
\resizebox{1\linewidth}{!}{
\subfloat{%
\begin{tabular}{@{}x{35}x{35}x{35}x{35}@{}}%
Seen & NerFormer & {MCC} & GT\\
\frame{\includegraphics[width=\linewidth]{figs/co3d_compare/apple_black_18_seen_rgb.jpg}}&
\frame{\includegraphics[width=\linewidth]{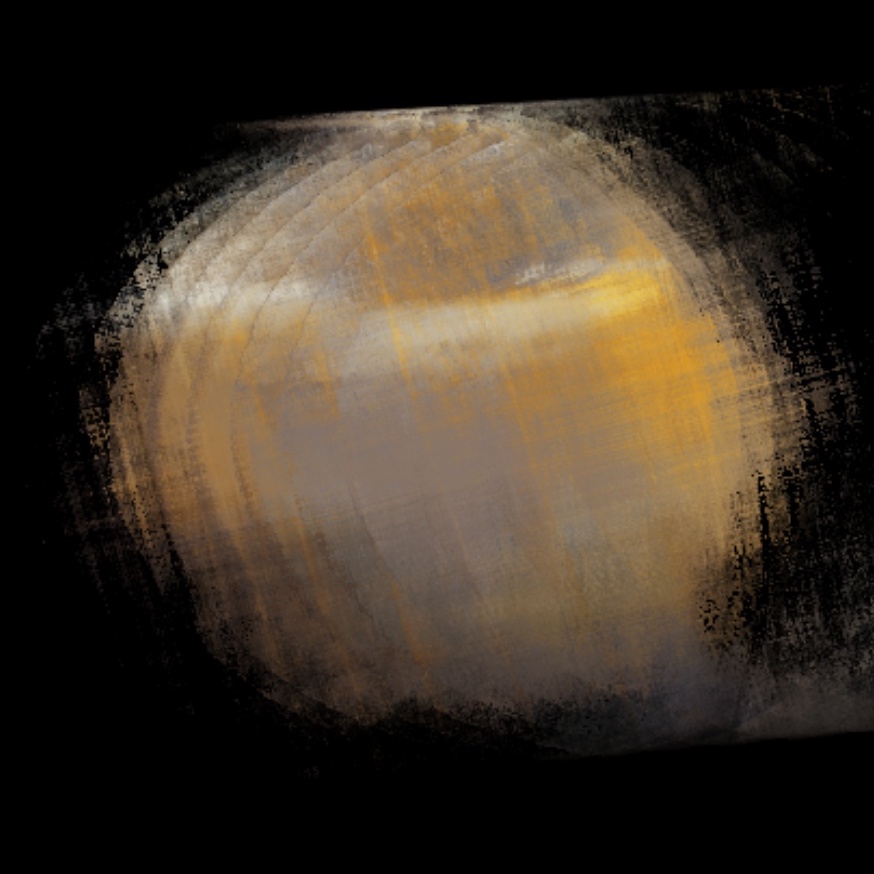}}&
\frame{\includegraphics[width=\linewidth]{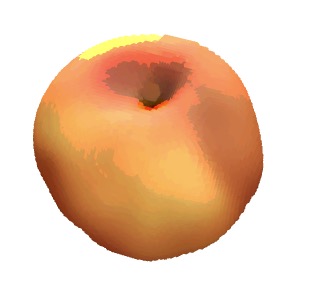}}&%
\frame{\includegraphics[width=\linewidth]{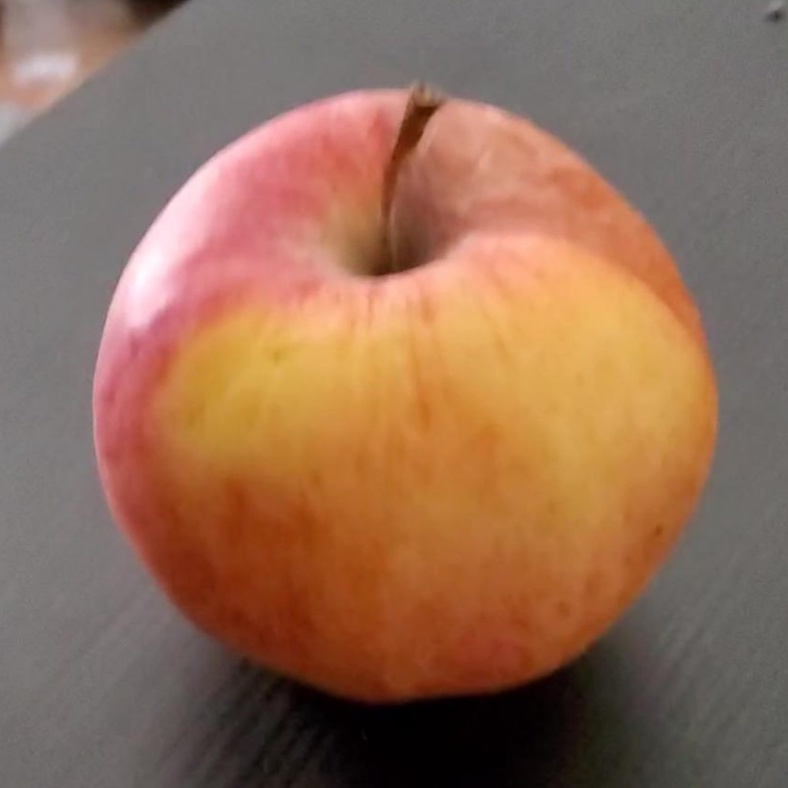}}
\end{tabular}
}\hfill
\subfloat{%
\begin{tabular}{@{}x{35}x{35}x{35}x{35}@{}}%
Seen & NerFormer & {MCC} & GT\\
\frame{\includegraphics[width=\linewidth]{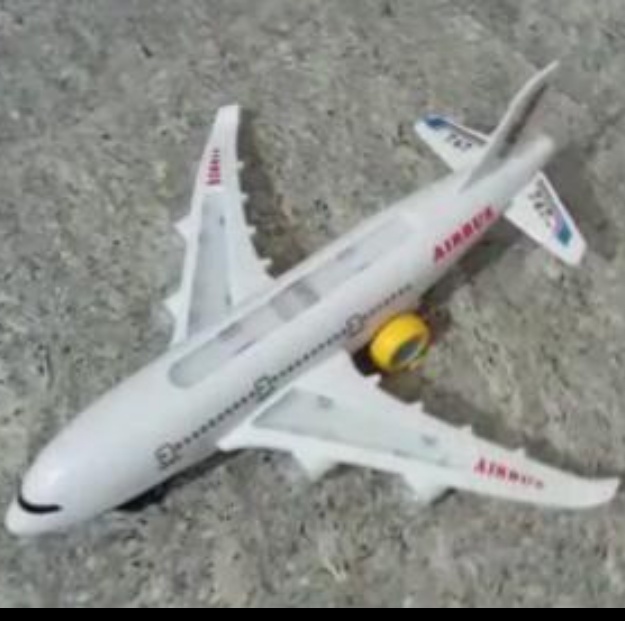}}&
\frame{\includegraphics[width=\linewidth]{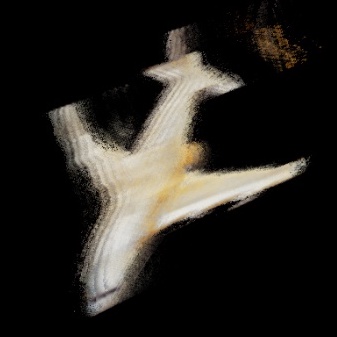}}&
\frame{\includegraphics[width=\linewidth]{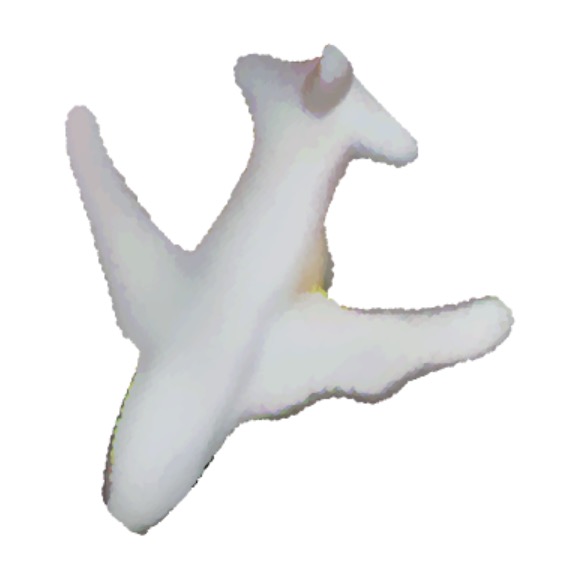}}&%
\frame{\includegraphics[width=\linewidth]{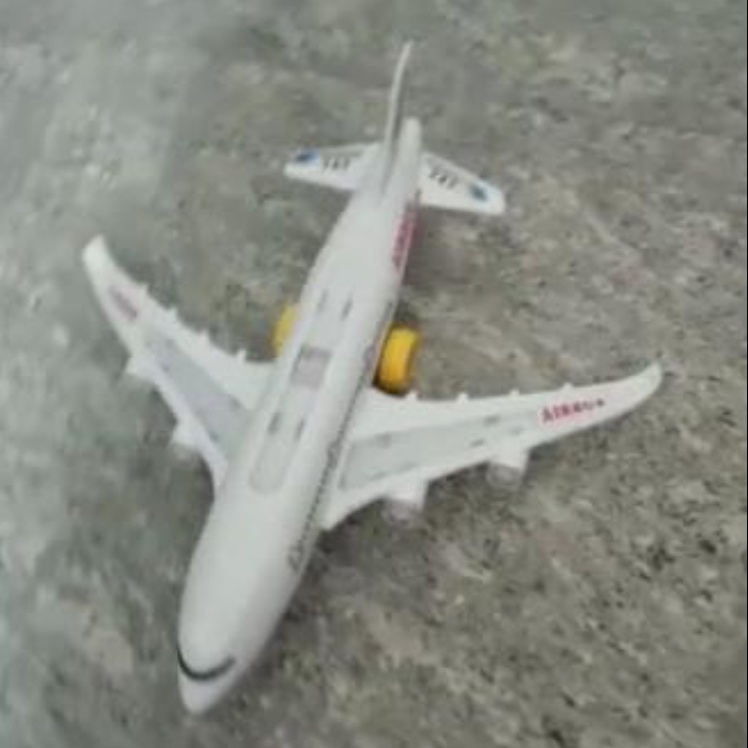}}
\end{tabular}
}}
\resizebox{1\linewidth}{!}{
\subfloat{%
\begin{tabular}{@{}x{35}x{35}x{35}x{35}@{}}%
\frame{\includegraphics[width=\linewidth]{figs/co3d_compare/suitcase_black_33_seen_rgb.jpg}}&
\frame{\includegraphics[width=\linewidth]{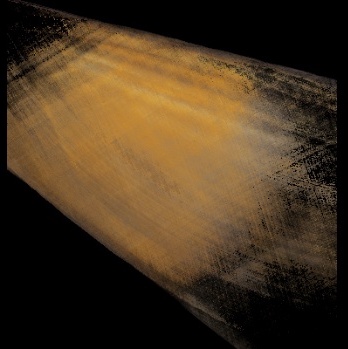}}&
\frame{\includegraphics[width=\linewidth]{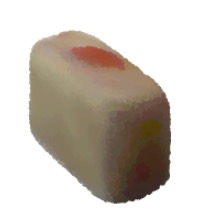}}&%
\frame{\includegraphics[width=\linewidth]{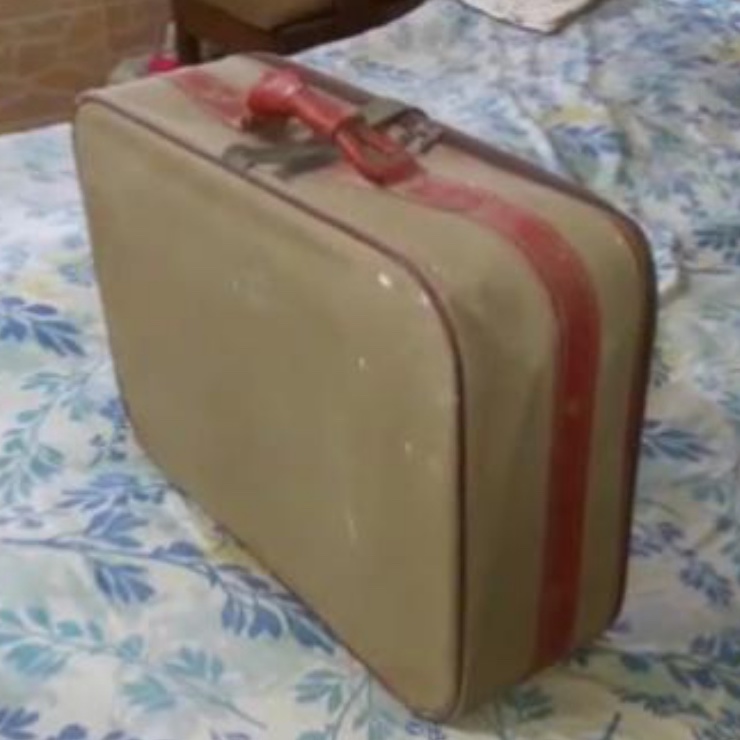}}
\end{tabular}
}%
\subfloat{%
\begin{tabular}{@{}x{35}x{35}x{35}x{35}@{}}%
\frame{\includegraphics[width=\linewidth]{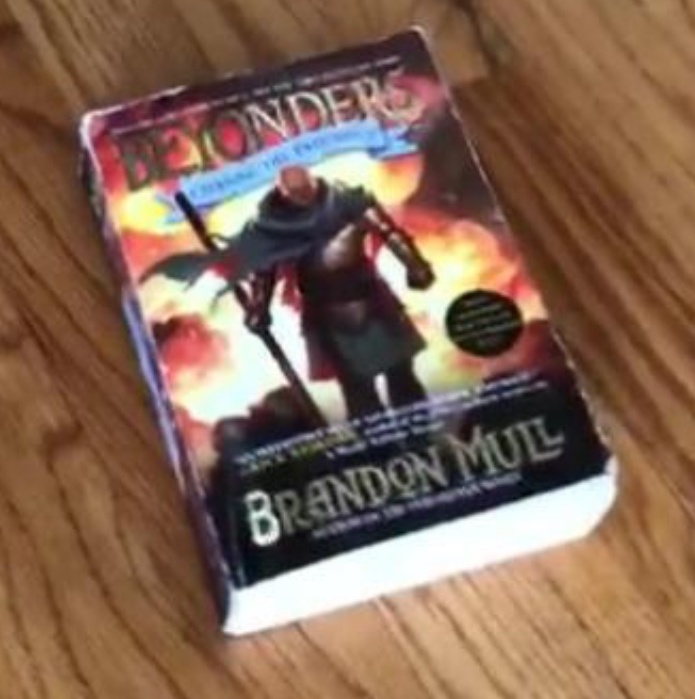}}&
\frame{\includegraphics[width=\linewidth]{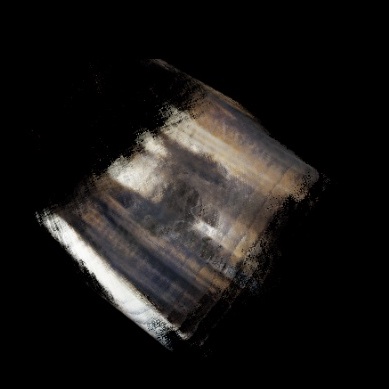}}&
\frame{\includegraphics[width=\linewidth]{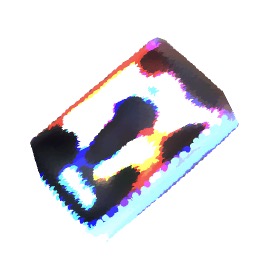}}&%
\frame{\includegraphics[width=\linewidth]{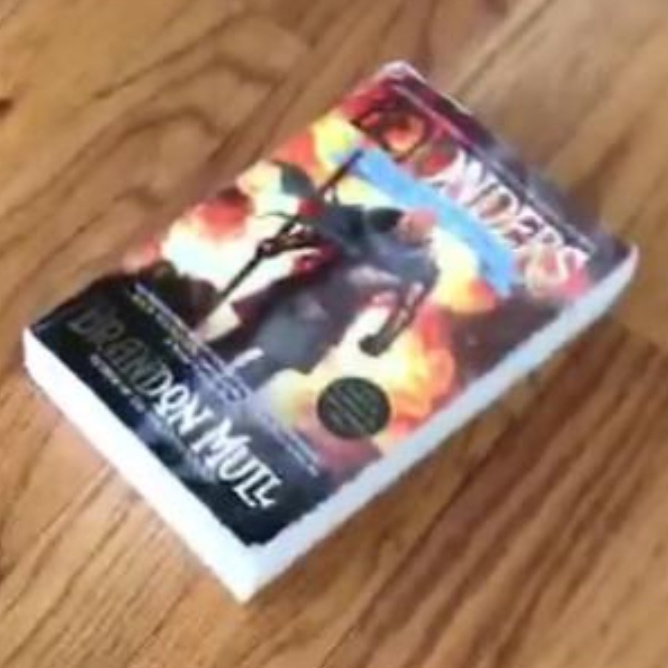}}
\end{tabular}
}}%
\vspace{-3mm}
\caption{\textbf{Qualitative Comparison between \method and NerFormer~\cite{reizenstein21co3d}.}
NerFormer captures texture but struggles with geometry; \method reconstructs shapes more accurately.}
\label{fig:co3dnerf}
\vspace{-6mm}
\end{figure}

\begin{figure*}[t]
\vspace{-1mm}
\tablestyle{1.0pt}{1.05}
\hspace{-1.5mm}\subfloat[Reconstruction of Hypersim Held-Out Scenes\label{fig:hypersim}]{%
\resizebox{0.5\linewidth}{!}{
\begin{tabular}{@{}x{70}x{70}@{}}
\multicolumn{2}{c}{\makecell{Input image\\ \frame{\includegraphics[height=1.5cm]{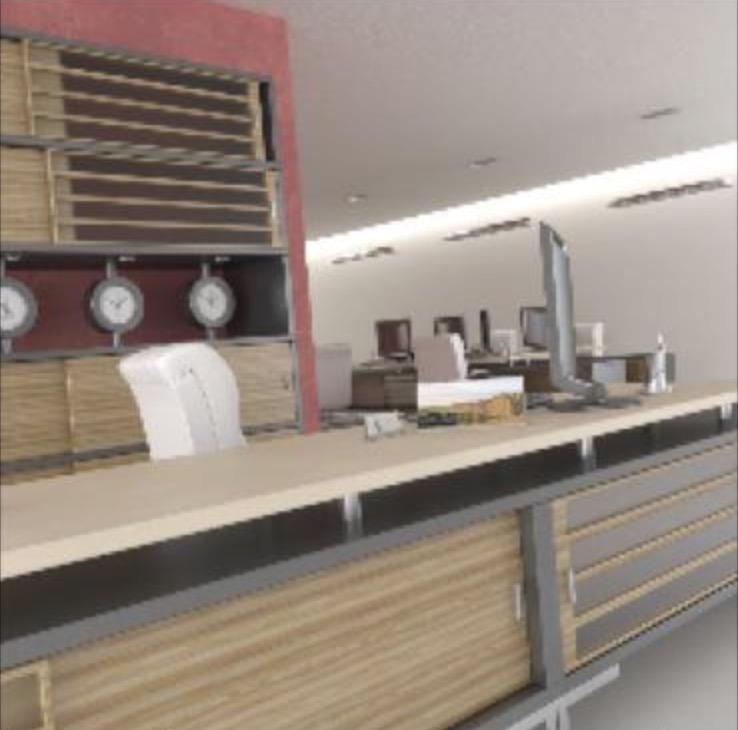}}}}\vspace{-2mm}\\
\makecell{Seen\vspace{-3mm}}&\makecell{Output\vspace{-3mm}}\\
\sceneimg{1.5cm}{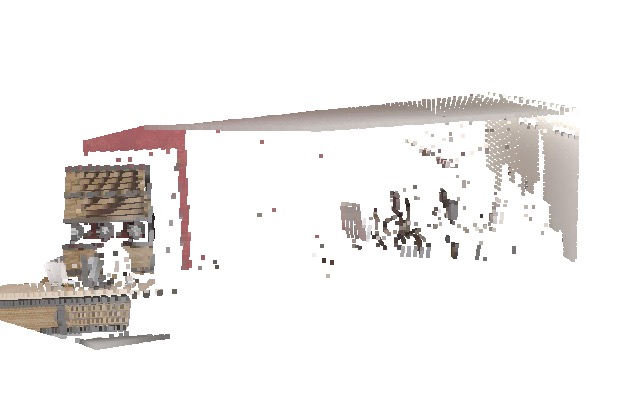}&
\sceneimg{1.5cm}{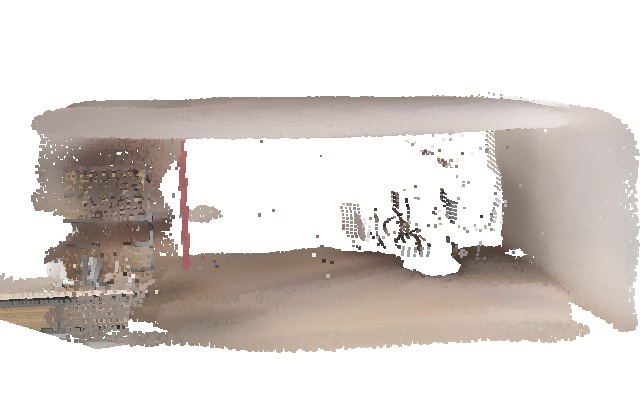}\\
\sceneimg{1.5cm}{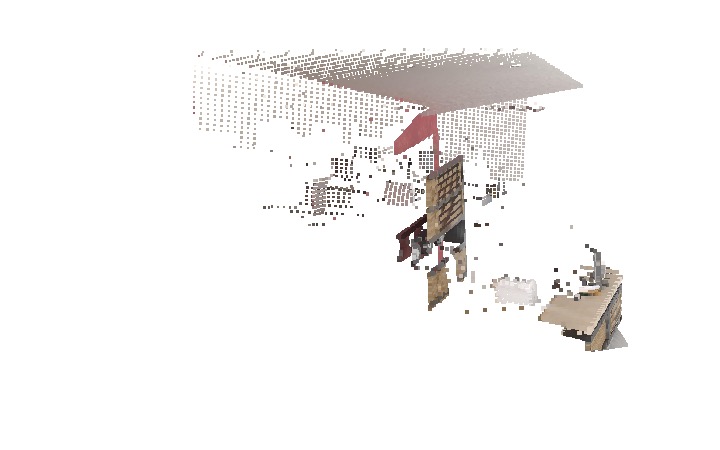}&
\sceneimg{1.5cm}{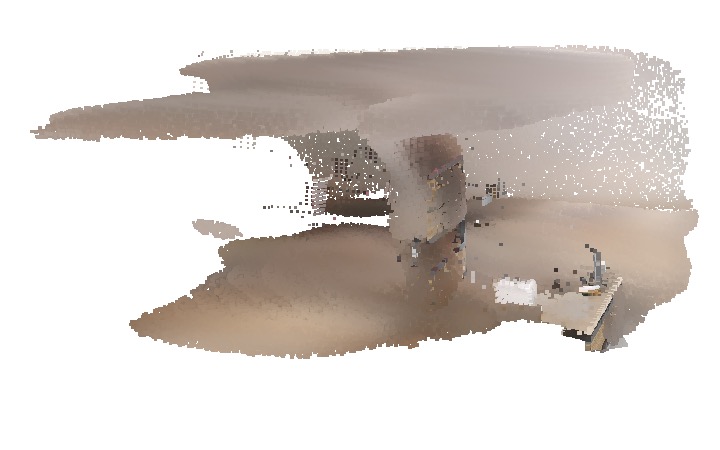}\\
\sceneimg{1.5cm}{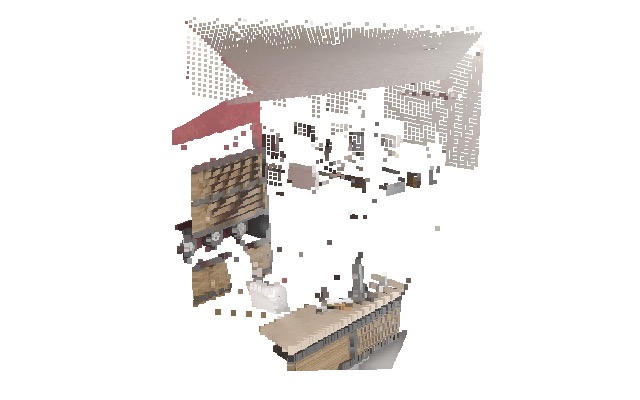}&
\sceneimg{1.5cm}{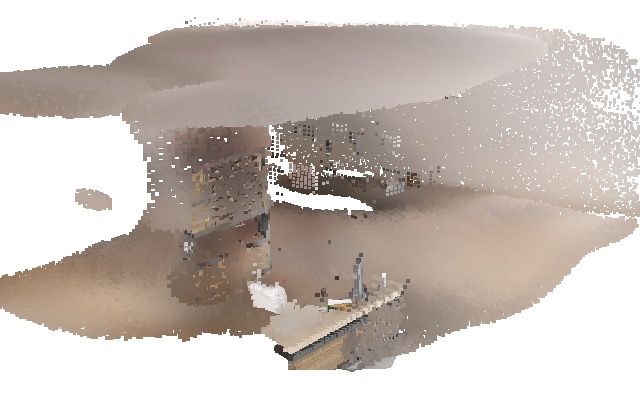}\vspace{-2mm}
\end{tabular}
\begin{tabular}{@{}x{70}x{70}@{}}
\multicolumn{2}{c}{\makecell{Input image\\ \frame{\includegraphics[height=1.5cm]{figs/hypersim_qual/rank7_i0_img.jpg}}}}\vspace{-2mm}\\
\makecell{Seen\vspace{-3mm}}&\makecell{Output\vspace{-3mm}}\\
\sceneimg{1.5cm}{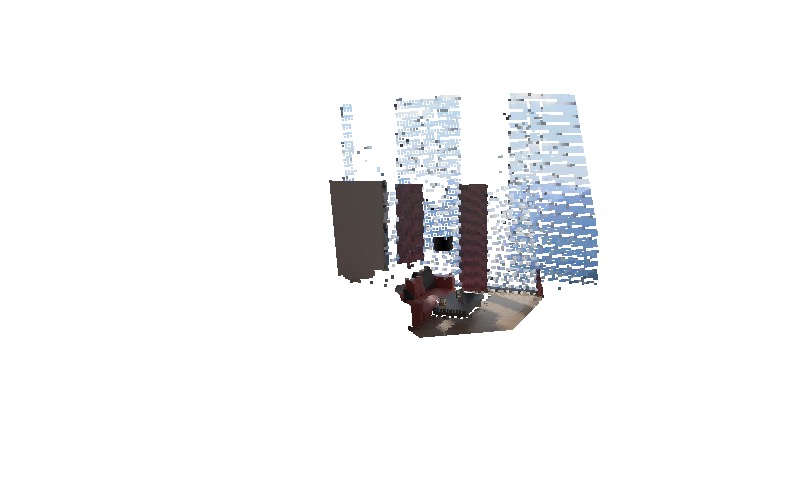}&
\sceneimg{1.5cm}{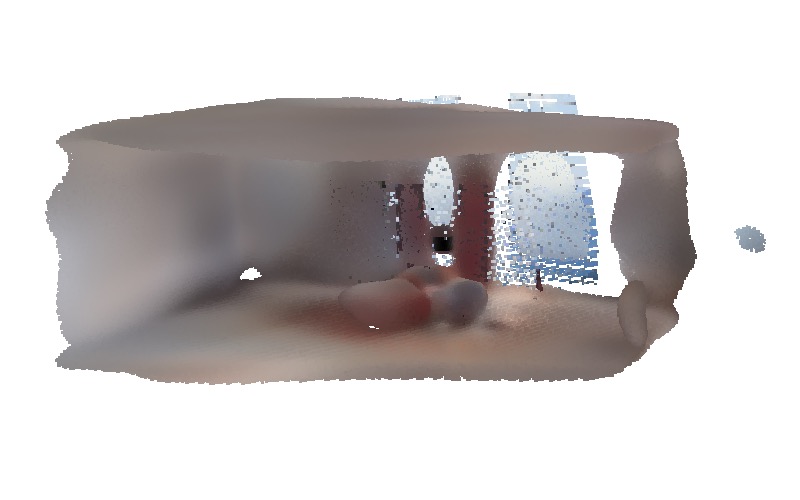}\\
\sceneimg{1.5cm}{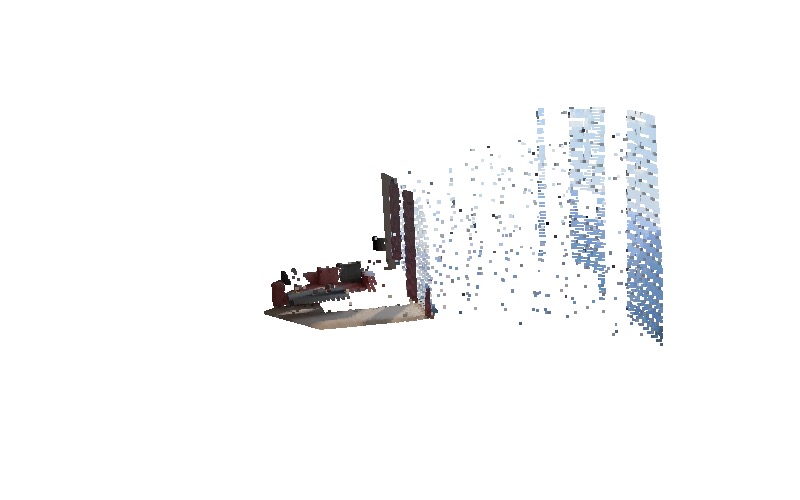}&
\sceneimg{1.5cm}{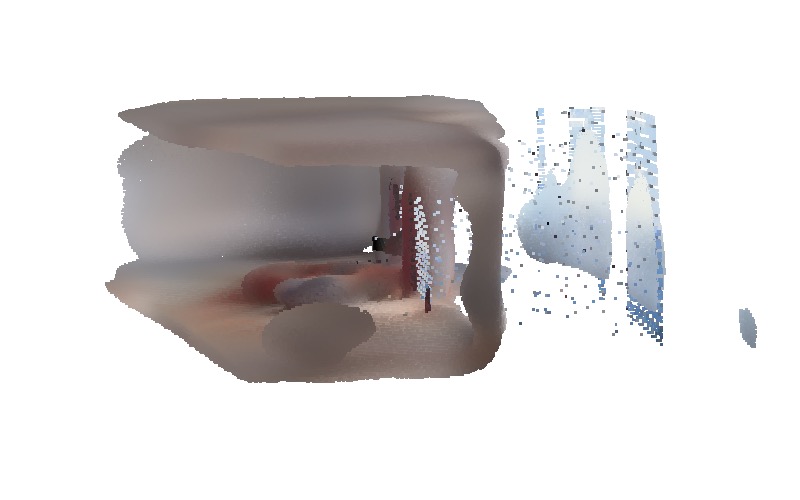}\\
\sceneimg{1.5cm}{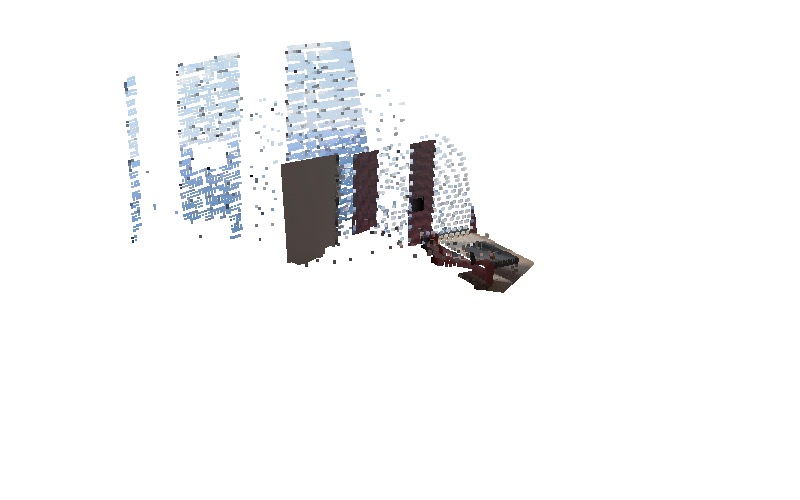}&
\sceneimg{1.5cm}{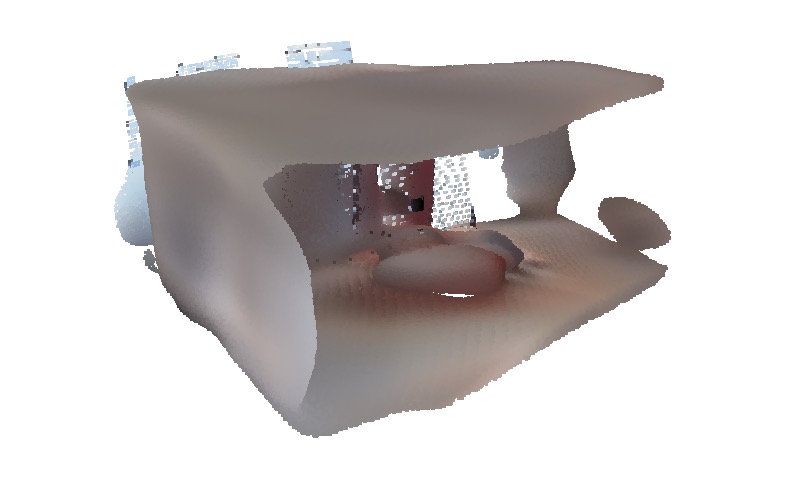}\vspace{-2mm}
\end{tabular}
}
}\hfill
\subfloat[\emph{Zero-Shot} Generalization to the Taskonomy Dataset\label{fig:taskonomy}]{%
\resizebox{0.5\linewidth}{!}{
\begin{tabular}{@{}x{70}x{70}@{}}
\multicolumn{2}{c}{\makecell{Input image\\ \frame{\includegraphics[height=1.5cm]{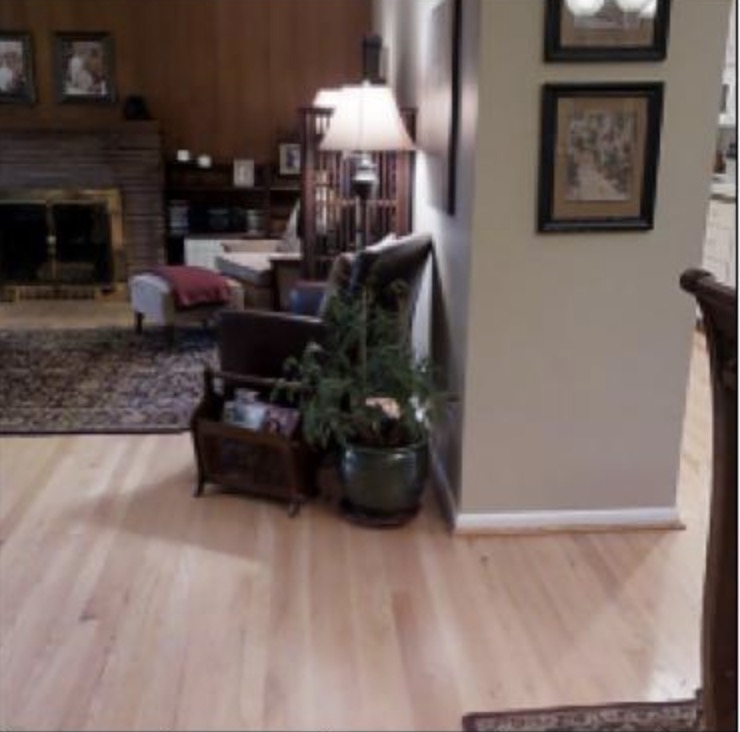}}}}\vspace{-2mm}\\
\makecell{Seen\vspace{-3mm}}&\makecell{Output\vspace{-3mm}}\\
\sceneimg{1.5cm}{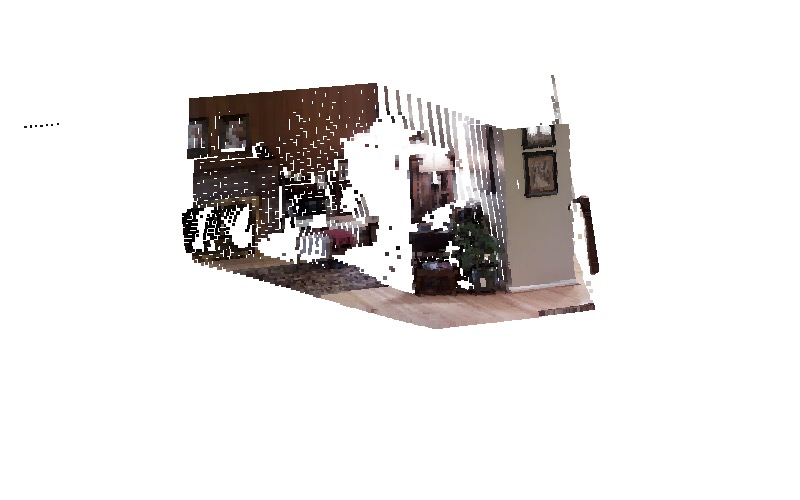}&
\sceneimg{1.5cm}{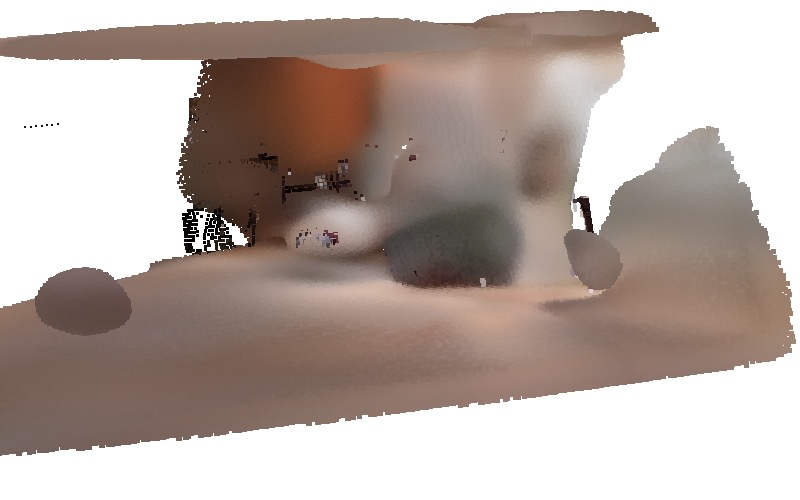}\\
\sceneimg{1.5cm}{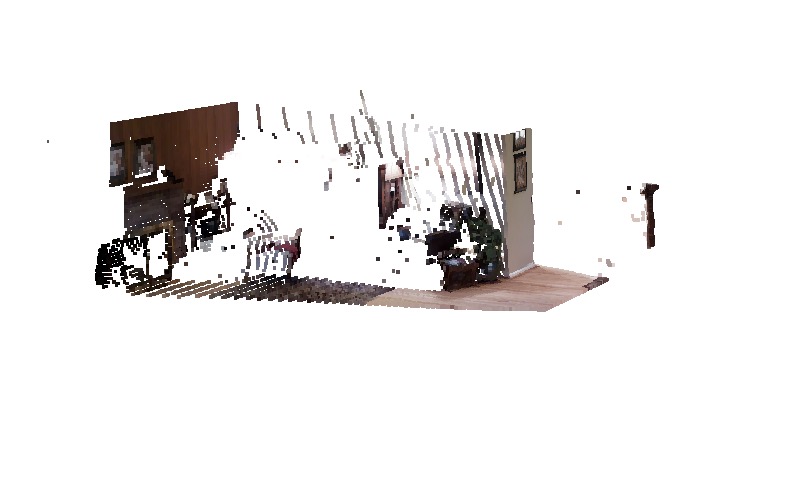}&
\sceneimg{1.5cm}{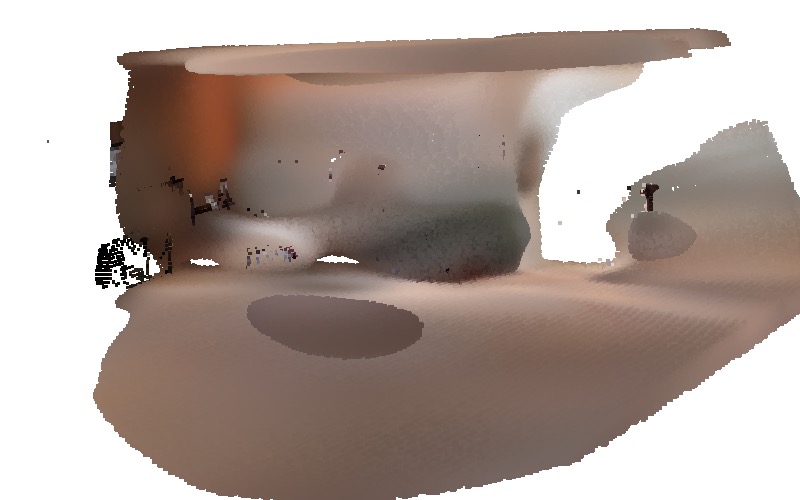}\\
\sceneimg{1.5cm}{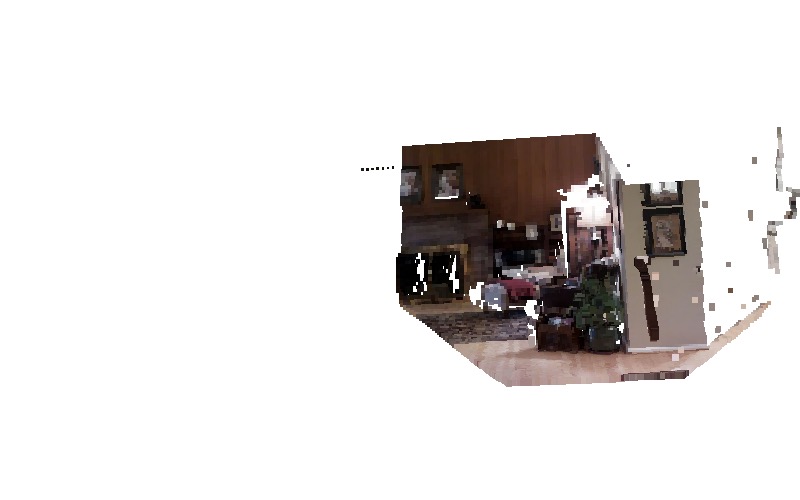}&
\sceneimg{1.5cm}{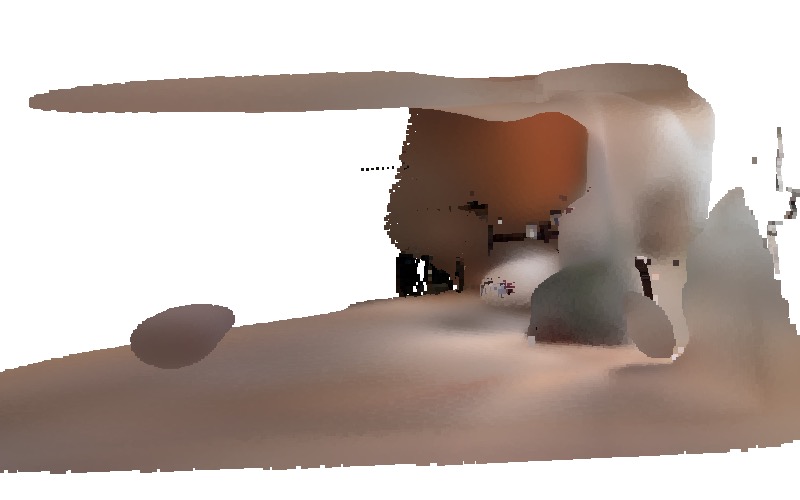}\vspace{-2mm}
\end{tabular}
\begin{tabular}{@{}x{70}x{70}@{}}
\multicolumn{2}{c}{\makecell{Input image\\ \frame{\includegraphics[height=1.5cm]{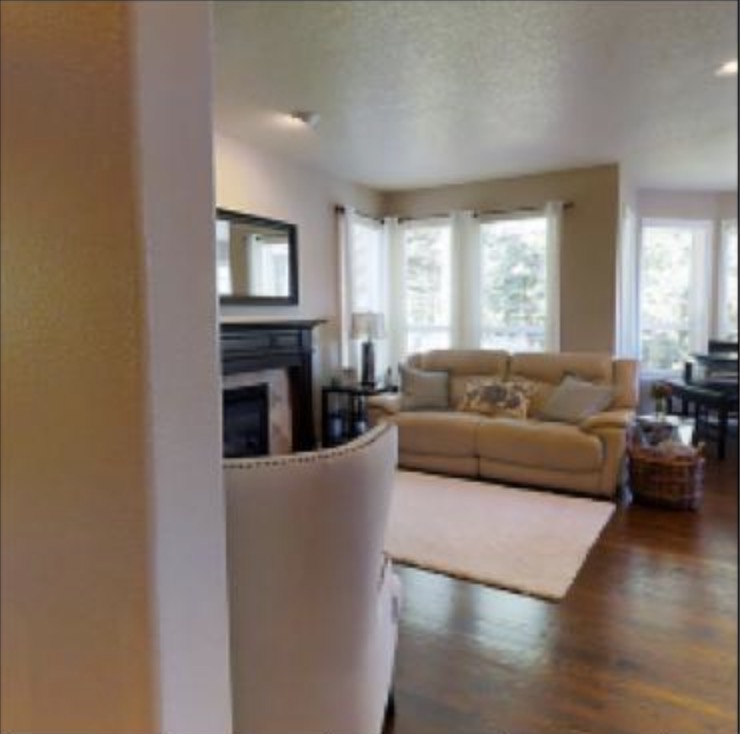}}}}\vspace{-2mm}\\
\makecell{Seen\vspace{-3mm}}&\makecell{Output\vspace{-3mm}}\\
\sceneimg{1.5cm}{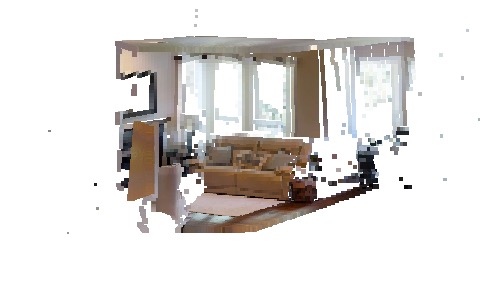}&
\sceneimg{1.5cm}{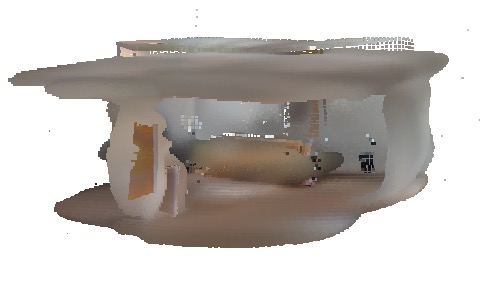}\\
\sceneimg{1.5cm}{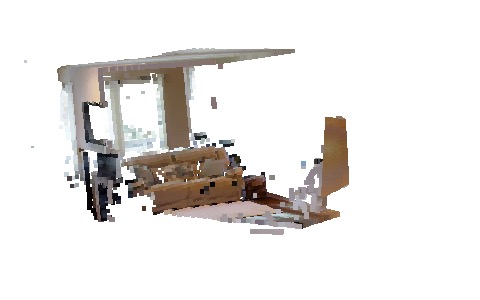}&
\sceneimg{1.5cm}{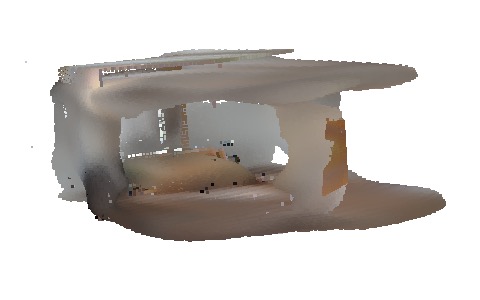}\\
\sceneimg{1.5cm}{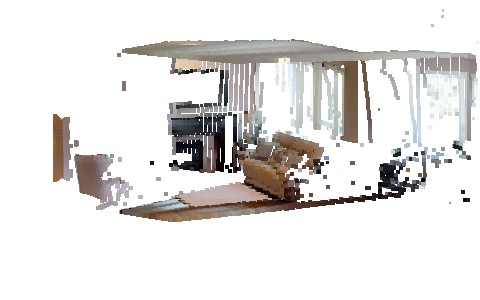}&
\sceneimg{1.5cm}{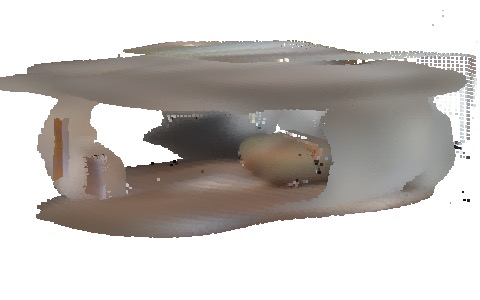}\vspace{-2mm}
\end{tabular}
}
}
\vspace{-3mm}
\caption{\textbf{Scene Reconstructions.} 
With a model trained on Hypersim, we show reconstructions on (a) held-out Hypersim scenes, and (b) novel scenes from Taskonomy.
From a single RGB-D image, \method reconstructs furniture, walls, floors, and ceilings, even outside the view frustum.
Capturing fine scene details is hard, but more data can help as our analysis in \secref{exp:scale} suggests.
See \href{https://mcc3d.github.io/}{project page} for animations.}
\vspace{-6mm}
\end{figure*}

\subsection{Comparison to Image-Conditioned NeRF}
\label{sec:nerf}
A recent successful line of work for 3D reconstruction extends NeRF~\cite{nerf} to cross-scene generalization from one or few views by conditioning on image embeddings~\cite{yu2021pixelnerf,henzler2021unsupervised,reizenstein21co3d,raj2021pixel}.
We compare to two recent best performing methods on CO3D from this family, NeRF-WCE~\cite{henzler2021unsupervised} and NerFormer~\cite{reizenstein21co3d}.
We evaluate for shape reconstruction using the official CO3D novel view depth metrics~\cite{reizenstein21co3d}: absolute (abs) and mean-squared error (MSE) on
the official CO3D challenge evaluation frames.
This puts \method at a disadvantage as it is not designed for synthesis via rendering.
Since \method uses RGB-D as input, we extend both methods, which originally use posed RGB views, to take depth as input or supervision.
For depth supervision we follow Deng~\etal~\cite{deng2022depth}, which shows strong results by supervising NeRF models with depth. 
To input depth, we fuse the XYZ input encoding, \ie $E^\mathrm{XYZ}(P)$, to the input image features.
\tabref{co3dnerf} shows that the baselines benefit from depth, as expected; \method outperforms them by a clear margin.
\figref{co3dnerf} qualitatively compares \method to the best baseline, NerFormer~\cite{reizenstein21co3d}.
NerFormer captures texture but struggles with geometry under the challenging single-view novel-category setting, thus rendering relatively blurry novel views.
Admittedly, these methods tend to work better with more (5-10) input views. 
\method predicts more accurate shapes from just a single view.

\section{Scene Reconstruction Experiments}
\label{sec:exp:scene}
\method naturally handles singles objects and scenes without modifications to its design.
So, now we turn to scenes.

\mypar{Task.}
We test 3D scene reconstruction from a single RGB-D image.
Formally, we aim to reconstruct everything in front of the camera ($z>0$ in camera coordinate system) up to a certain range.
Note that this includes areas outside the camera frustum, which increases the complexity of the task.

\mypar{Dataset.}
We experiment on the Hypersim dataset~\cite{roberts2021hypersim}, which contains complex, diverse scenes, such as warehouse, lofts, restaurants, church \etc, with over 77k images. 
We split the dataset into 365 scenes for training and 46 scenes for testing.
We use images along with the associated depth as ground truth for training.
Since 3D meshes are available, we use them for evaluation and report the metrics from \secref{exp:obj}.

\begin{table}[t]
\tablestyle{1.0pt}{1.05}
\begin{tabular}{@{}lx{40}x{40}x{40}@{}}
& Acc & Cmp & F1\\
\shline
DRDF~\cite{kulkarni2021drdf} & 54.4 & 1.0 & 2.0\\
DRDF (our arch)~\cite{kulkarni2021drdf}\quad\quad\quad\quad\quad & 54.2 & 1.4 & 2.7\\
\textbf{MCC} &\textbf{66.3} & \textbf{1.5} & \textbf{2.8}
\end{tabular}
\vspace{-3.5mm}
\caption{\textbf{Comparison to DRDF on Hypersim.}
MCC outperforms the state-of-the-art scene reconstruction approach, DRDF~\cite{kulkarni2021drdf}, extended to input depth, with both its original and our architecture.}
\label{tab:hypersim}
\vspace{-5mm}
\end{table}

\subsection{Hypersim Scene Reconstruction}
\mypar{Qualitative Results.}
\figref{hypersim} shows qualitative results on Hypersim~\cite{roberts2021hypersim}.
While \method only sees the scene within the view frustum, it is able to complete furniture, walls, floors, and ceilings.
For instance, in the left example, \method predicts the space behind the kitchen, including the floors, which are almost entirely occluded in the input view.
In the right example, \method predicts the wall on the left which is entirely outside of the view frustum.
Scene reconstruction from a single view is hard; while \method reconstructs the room geometry it fails to capture fine details in both shape and texture.
We expect more data to significantly improve performance, as suggested by our scaling analysis in \secref{exp:scale}.

\mypar{Quantitative Evaluation.}
We compare to recent state-of-the-art on scene reconstruction, DRDF~\cite{kulkarni2021drdf}, which we extend to take RGB-D inputs like \method.
\tabref{hypersim} shows that \method outperforms DRDF across all metrics.
We also extend DRDF to use \method's architecture but keeping its original loss and ray-based inference.
This variant performs better than the original DRDF but still worse than \method.

\subsection{Zero-Shot Generalization to Taskonomy}
Finally, we deploy \method, trained on Hypersim, on novel scenes from Taskonomy~\cite{zamir2018taskonomy}.
While photorealistic, Hypersim is synthetic, while Taskonomy is real.
So, we test both generalization to novel scenes but also the ``sim-to-real" transfer.
\figref{taskonomy} shows \method's reconstructions, which demonstrate that our model is able to reconstruct the room layout (floors, walls, ceilings) in this challenging setting.

\vspace{-0.8mm}
\section{Failure Cases}
\vspace{-0.8mm}
While \method has demonstrated promising results, we observe three error modes:
(1) Sensitivity to depth input.
\method can recover from noisy depth inputs.
But if depth is largely incorrect, it will fail to reconstruct accurate 3D geometry.
(2) Distribution shifts.
For targets far from the training distribution, we see errors in texture and geometry (\eg, Rubik's cubes).
(3) High-fidelity texture.
Detailed texture predictions from a single view are difficult and \method often omits details (\eg, text on volleyball in \figref{co3dqual}).

\vspace{-0.8mm}
\section{Conclusions}
\vspace{-0.8mm}
We present MCC, a general-purpose 3D reconstruction model that works for both objects and scenes.
We show generalization to challenging settings, including in-the-wild captures and AI-generated images of imagined objects.
Our results show that a simple point-based method coupled with category-agnostic large-scale training is effective.
We hope this is a step towards building a general vision system for 3D understanding.
Models and code are available \href{https://github.com/facebookresearch/MCC}{online}. 

From an ethics standpoint, as with all data-driven methods, MCC can potentially inherit the bias (if any) in data.
In this project, we solely train on inanimate objects and scenes to minimize the risk.
We do not foresee immediate negative repercussions with the model, but caution against future use without paying careful attention to the training dataset.

\appendix
\section{Appendix}

\subsection{Animations and Interactive Visualizations}
We provide 360-view animations and interactive 3D visualizations for all qualitative results, in Figures 4, 7 and 9, and more in our \href{https://mcc3d.github.io/}{project page}.
Our video animations are shown in the main window and interactive 3D visualizations are available by clicking on the \emph{3D icon}, per the instructions in the webpage.

\subsection{Architecture Specifications}
\tabref{archi} describes in detail the MCC architecture for the $E^\mathrm{RGB}$ and $E^\mathrm{XYZ}$ encoders and the decoder.

The $E^\mathrm{RGB}$ and $E^\mathrm{XYZ}$ encoders follow the ``ViT-Base" transformer architecture by Dosovitskiy~\etal~\cite{transformer,vit}.
The transformer architecture is composed of 12 layers of a 768-dimensional self-attention operator with 12 heads, referred to as multi-head attention (MHA), followed by a 3072-dimensional 2-layer MLP.
The input image size is 224$\times$224.
The RGB image $I$, input to the $E^\mathrm{RGB}$ encoder, is embedded via a single convolutional layer, of a $16\times16$-sized kernel and a $16\times16$ stride, to produce $N^{enc}=196$ tokens.
The (seen) points $P$, input to the $E^\mathrm{XYZ}$ encoder, are first linearly projected to a 768-dimensional representation and then embedded via a single transformer layer which operates on $16\times16$ non-overlapping patches as explained in Section 3.4 of the main paper and further described in \tabref{archi}, resulting also in $N^{enc}=196$ tokens.
The single transformer layer used for the patch embeddings defines a \texttt{[cls]} token whose output is the embedding for each patch, as is commonly used in~\cite{devlin2018bert,vit} and referred to as a readout token.
 
Our decoder follows the decoder design from MAE~\cite{he2022masked}.
It is composed of 8 layers of a 512-dimensional self-attention operator with 16 heads followed by a 2048-dimensional 2-layer MLP.
The input to the decoder is: (a) $N^q=550$ 3D point queries which are linearly projected to a 768-dimensional vector, and (b) input $R$ which concatenates the $N^{enc}$ output tokens from $E^\mathrm{RGB}$ and $E^\mathrm{XYZ}$ in the channel dimension and then linearly projects each to a 768-dimensional vector. This results in a $768\times (N^q+N^{enc}) = 768 \times 746$ input to the decoder.
Our decoder additional defines a global \texttt{[cls]} token whose role is to ``summarize" all inputs in the transformer and can be attended freely by other tokens.

LayerNorm~\cite{ba2016layer} is used in all self-attention and MLP layers following standard practice~\cite{transformer,vit,he2022masked}.

\begin{table}[t]
    \centering
    \scalebox{1.04}{
    \subfloat[\textbf{Encoder $E^\mathrm{RGB}$}]{%
        \tablestyle{5pt}{1.05}  
        \begin{tabular}{c|c|c}
            Stage & Operators & Output sizes \\
            \shline
            \multirow{1}{*}{Input $I$} & -   &  {\wcolor{3}}\x{\xycolor{224}}\x{\xycolor{224}}   \\
            \hline            
            \multirow{2}{*}{Patch embed} & \multicolumn{1}{c|}{Conv 16\x16, {768}} &    \multirow{2}{*}{\wcolor{768}\x{\xycolor{196}}}    \\
            & (stride 16\x16)   \\
            \hline
            \multirow{2}{*}{Transformer layers}  & \blockatta{768}{{3072}}{12} & \multirow{2}{*}{\wcolor{768}\x{\xycolor{196}}}  \\
            &  & \\
        \end{tabular}}}\vspace{5mm}

    \scalebox{1.04}{
    \subfloat[\textbf{Encoder $E^\mathrm{XYZ}$}]{%
        \tablestyle{5pt}{1.05}  
        \begin{tabular}{c|c|c}
            Stage & Operators & Output sizes \\
            \shline
            \multirow{1}{*}{Input $P$} & -   &  {\wcolor{3}}\x{\xycolor{224}}\x{\xycolor{224}}   \\
            \hline
            \multirow{1}{*}{Embed} & {Linear, 768} &   {\wcolor{768}}\x{\xycolor{224}}\x{\xycolor{224}}    \\
            \hline
            \multirow{3}{*}{Patch embed} & \blockattb{768}{{1536}}{1} & \multirow{3}{*}{\wcolor{768}\x{\xycolor{196}}}    \\
            &&\\
            &&\\
            & (on each 16\x16 patch)   \\
            \hline
            \multirow{2}{*}{Transformer layers}  & \blockatta{768}{{3072}}{12} & \multirow{2}{*}{\wcolor{768}\x{\xycolor{196}}}  \\
            &  & \\
        \end{tabular}}}\vspace{5mm}

    \scalebox{1.04}{
    \subfloat[\textbf{Fusion Module $f$}]{%
        \tablestyle{5pt}{1.05}  
        \begin{tabular}{c|c|c}
            Stage & Operators & Output sizes \\
            \shline
            \multirow{2}{*}{Input encodings} & -   &  {\wcolor{768}}\x{\xycolor{196}}   \\
            &&  {\wcolor{768}}\x{\xycolor{196}}\\
            \hline
            \multirow{1}{*}{Concat} & Concat &   {\wcolor{1536}}\x{\xycolor{196}}    \\
            \hline
            \multirow{1}{*}{Linear} & {Linear, 768} &   {\wcolor{768}}\x{\xycolor{196}}    \\
        \end{tabular}}}\vspace{5mm}
    \scalebox{1.04}{
    \subfloat[\textbf{Decoder $Dec$}]{%
        \tablestyle{5pt}{1.05}  
        \begin{tabular}{c|c|c}
            Stage & Operators & Output sizes \\
            \shline
            \multirow{1}{*}{Input queries} & -   &  {\wcolor{3}}\x{\xycolor{550}}   \\
            \hline
            
            \multirow{1}{*}{Embed} & {Linear, 768} &   {\wcolor{768}}\x{\xycolor{550}}    \\
            \hline
            \multirow{1}{*}{Concat with $R$} & Concat &   {\wcolor{768}}\x{\xycolor{746}}    \\
            \hline
            \multirow{2}{*}{Transformer layers}  & \blockatta{512}{{2048}}{8} & \multirow{2}{*}{\wcolor{768}\x{\xycolor{746}}}  \\
            &  & \\
        \end{tabular}}}\vspace{2mm}
\caption{\textbf{Architecture specification} for each part of the MCC model. MHA($c$): Multi-Head Attention~\cite{transformer} with $c$ channels. MLP($c'$): MultiLayer Perceptron with $c'$ channels. \texttt{[cls]} readout: feature readout with the \texttt{[cls]} token~\cite{devlin2018bert,vit}. Here, we use the default choice of $N^q=550$ queries. We omit the optional \texttt{[cls]} token in the outputs of the transformers for clarity.}
\label{tab:archi}
\end{table}

\subsection{Held-Out CO3D Categories}
In our experiments, we hold out 10 randomly selected categories which we use as our test set.
The 10 randomly selected held-out categories are:
\{\emph{apple,
ball,
baseballglove,
book,
bowl,
carrot,
cup,
handbag,
suitcase,
toyplane}\}.
They have 8,453 example videos in total.
Please see the original CO3D paper for more information about CO3D~\cite{reizenstein21co3d}.

\subsection{Additional Implementation Details for Scene Reconstruction Experiments}
Similar to the object reconstruction experiments, we train MCC on Hypersim~\cite{roberts2021hypersim} with Adam~\cite{kingma2014adam} for 100k iterations with an effective batch size of 512 using 32 GPUs, a base learning rate of 5$\times$10$^{-5}$ with a cosine schedule and a linear warm-up for the first 10\% of iterations.
Training takes $\app$1.6 days.
We normalize each scene to have zero-mean and unit-variance.
At inference time, we predict points up to 6.0 units (\ie,  6$\times$ standard deviation) away from the camera origin.
Since we aim at predicting the scene in front of the camera, we use the camera view coordinate system ($X$-axis points top/down, $Y$-axis points left/right, and $Z$-axis points out from the image plane).
We randomly scale augment images by $s \in \sbr{0.8, 1.2}$, as in the object reconstruction model, but do not perform rotation augmentation.
Other implementation details follow the CO3D experiments.

{\small
\bibliographystyle{ieee_fullname}
\bibliography{bib}
}

\end{document}